\documentclass[11pt]{article}

\usepackage[preprint]{acl}

\usepackage{hyperref}
\usepackage{url}
\usepackage{soul,xcolor}
\newcommand{\hlc}[2]{\sethlcolor{#1}\hl{#2}}

\usepackage{times}
\usepackage{latexsym}
\usepackage{algpseudocode}
\usepackage{amsmath}
\usepackage{algorithm}
\usepackage{comment}
\usepackage{enumitem}
\usepackage[T1]{fontenc}
\usepackage[utf8]{inputenc}
\usepackage{listings}
\usepackage[listings, skins, most]{tcolorbox}
\usepackage[normalem]{ulem}   
\usepackage{xcolor}           
\usepackage{longtable}
\usepackage{booktabs}
\lstdefinelanguage{Markdown}{
  basicstyle=\ttfamily\footnotesize,
  sensitive=false,
  morecomment=[l]{\#},   
  morecomment=[s]{```}{```},
  morestring=[b]",        
  morestring=[b]', 
}
\usepackage{inconsolata}
\usepackage{tabularx}
\PassOptionsToPackage{table}{xcolor}
\usepackage{colortbl}
\usepackage{microtype}
\usepackage{hyperref}
\usepackage{wrapfig}
\usepackage{inconsolata}
\usepackage{algorithm}
\usepackage{algpseudocode}
\usepackage{graphicx}
\usepackage{xspace}
\usepackage{array}
\usepackage{multirow}
\usepackage{fontawesome5}
\usepackage{makecell}

\definecolor{teal}{RGB}{0,128,128}
\definecolor{orange}{RGB}{255,165,0}
\definecolor{purple}{RGB}{128,0,128}

\newif\ifcommentsoff

\newcommand{\methodname}{\texttt{Frankentexts}\xspace}
\newcommand{\frankentext}{\texttt{Frankentexts}\xspace}

\definecolor{MonsterGreen}{HTML}{4A6C2F}
\definecolor{StitchFlesh}{HTML}{A2C14C}
\definecolor{BoltSilver}{HTML}{B0B8B4}
\definecolor{GraveyardBlack}{HTML}{1C1C1C}
\definecolor{StitchPurple}{HTML}{5C3D5E}
\definecolor{ElectroBlue}{HTML}{3B9C9C}
\definecolor{RottenBrown}{HTML}{6B4F3B}
\definecolor{LabCoatWhite}{HTML}{EDEDED}

\newcommand{\qwen}[1]{{Qwen}\xspace}
\newcommand{\gemini}[1]{{Gemini}\xspace}
\newcommand{\ofour}[1]{{o4-mini}\xspace}
\newcommand{\rone}[1]{{R1}\xspace}

\title{\includegraphics[width=0.25\textwidth]{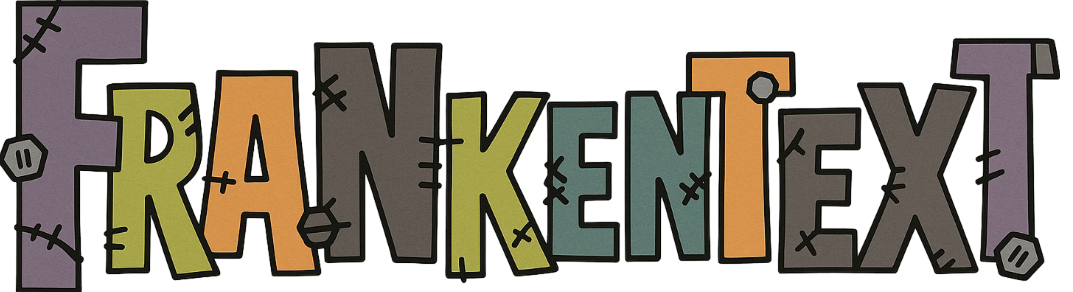}\\Stitching random text fragments into long-form narratives}

\author{
  Chau Minh Pham \textsuperscript{\faGhost}\quad
  Jenna Russell \textsuperscript{\faGhost}\quad
  Dzung Pham \textsuperscript{\faFlask}\quad
  Mohit Iyyer \textsuperscript{\faGhost\ \faFlask} \\
  \textsuperscript{\faGhost} University of Maryland, College Park \quad
  \textsuperscript{\faFlask} University of Massachusetts Amherst \\
  {\small \faEnvelope\ \texttt{\{chau,jennarus,miyyer\}@umd.edu}, \texttt{dzungpham@cs.umass.edu}}
}
\begin{document}
\maketitle
\begin{abstract}
    As AI text detectors are increasingly used to flag LLM-generated writing, a natural question arises: \textit{are there forms of high-quality generated narrative that can evade such detection?} We introduce \frankentext, a long-form narrative generation paradigm that treats an LLM as a composer of existing texts rather than as an author.
    Given a writing prompt and thousands of randomly sampled human-written snippets, the model assembles a coherent narrative where most tokens (e.g., 90\%) are copied verbatim from the source passages.
    Despite the extreme challenge of the task, we observe through extensive automatic and human evaluation that \frankentext\ improve over vanilla LLM generations in key writing quality metrics such as diversity and novelty while remaining mostly coherent and relevant to the prompt.
    Furthermore, \frankentext\ pose a fundamental challenge to current AI text detectors: 72\% of \frankentext\ produced by our best configuration (Gemini-2.5-Pro with 5K input snippets) are misclassified as human-written by Pangram, a state-of-the-art detector.
    Human annotators praise \frankentext\ for their inventive premises, vivid descriptions, and dry humor; however, they still identify issues with abrupt tonal shifts and uneven grammar across segments.
    Overall, the emergence of high-quality yet low-detectability \frankentext\ challenges established authorship norms while raising concerns about the publishing economy.
\end{abstract}

\begin{center}
\scriptsize
\faGithub\ \quad \url{https://github.com/chtmp223/Frankentext}
\end{center}
\begin{figure*}[t]
    \centering
    \includegraphics[width=\textwidth]{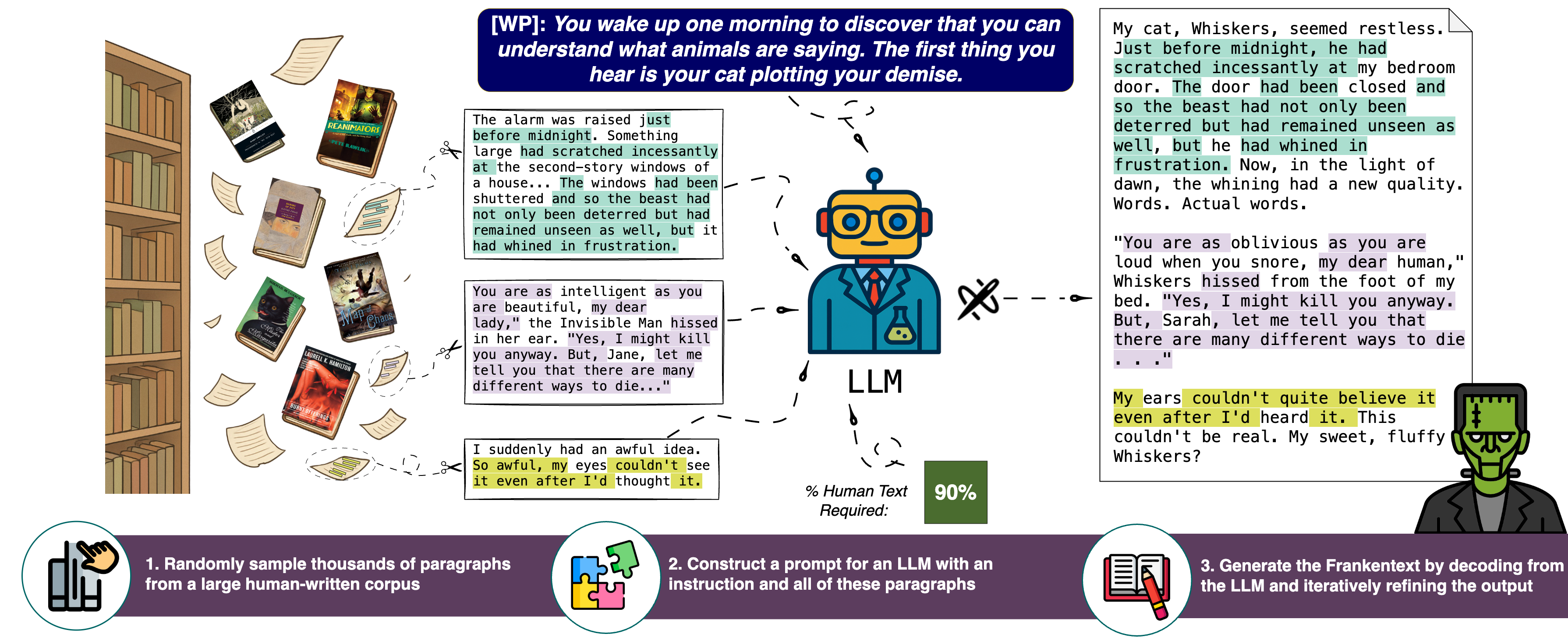}
    \caption{The \methodname pipeline. First, random paragraphs are sampled from a large corpus of human-written books. Then, an LLM is prompted with the paragraphs, a writing prompt, and instructions to include a certain amount of human text verbatim, to generate the first draft of a \texttt{Frankentext}, which is further edited into a coherent and faithful final version (see Algorithm \ref{pseudocode}).
    The highlighted texts are the human-written snippets that are selected to be included in the \frankentext.
    }
    \label{fig:overview}
\end{figure*} 

\section{Introduction}

LLM-generated narratives are now appearing at scale in venues like self-publishing platforms \citep{Knibbs2024ScammyAIGeneratedBooksFloodingAmazon} and newspapers \citep{russell2025aiuseamericannewspapers}. Although AI writing is widely perceived as lower quality than professional human writing~\citep{art_or_artifice,shaib2025measuringaisloptext,russell2025peoplefrequentlyusechatgpt}, the livelihood of writers is threatened by the combination of (1) the \emph{speed} that AI narratives can be churned out and (2) the development of techniques that improve the \emph{quality} of AI writing. Tools that can detect AI use in writing are a natural defense: if platforms can reliably identify LLM use, they can require disclosure or enforce authorship policies \cite{emi2024technicalreportpangramaigenerated, Naddaf2025ICLRPeerReviewsAI}.
 While detectors can be evaded by adversarial ``humanization'' pipelines \citep{krishna2023paraphrasing, sadasivan2025aigeneratedtextreliablydetected, cheng2025adversarial}, they are also improving rapidly: modern commercial detectors (e.g., Pangram, GPTZero) are tuned for low false-positive deployment and robustness on challenging benchmarks \citep{dugan-etal-2024-raid,masrour-etal-2025-damage,jabarian2025artificial}. In this paper, we consider the best-case view for AI detection and ask: \textbf{can a black-box LLM user generate narratives that are simultaneously low-effort, high-quality, and likely to fool robust and accurate AI detectors?}

To explore this possibility, we introduce \frankentext:\footnote{Inspired by Mary Shelley’s \textit{Frankenstein} \citep{shelley1818frankenstein}, where Victor Frankenstein assembles a creature from human parts that nonetheless emerges as an intelligent being.} long-form narratives constructed by LLMs under the constraint that the majority of the output must be copied verbatim from a provided set of human-written spans, with only minimal connective text added by the model. We propose \frankentext construction as a new narrative generation paradigm that is distinct from vanilla autoregressive decoding, which often produces formulaic prose and plots~\citep{art_or_artifice, russell2025peoplefrequentlyusechatgpt, shaib2025measuringaisloptext}, and retrieval-augmented generation, where retrieved text primarily supplies factual grounding. Given a writing prompt and thousands of human-written snippets, an LLM selects, orders, and connects spans so that a pre‑specified fraction of the final text (e.g., 90\%) is copied verbatim (\autoref{fig:overview}). Since the search space for snippet selection and ordering is combinatorially large, our framework allows an LLM to implicitly explore this space by proposing a draft and minimally editing it for coherence. 

\textbf{\includegraphics[height=0.6em]{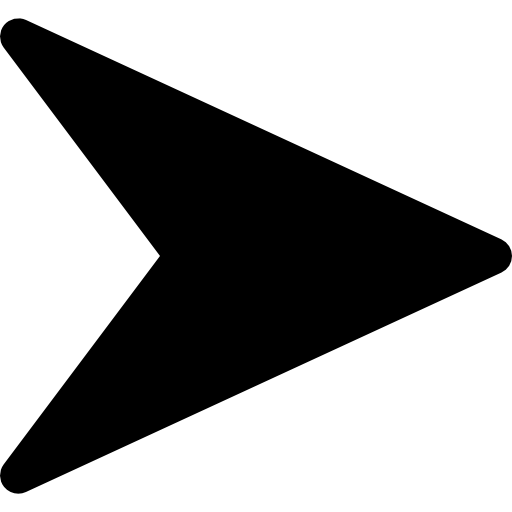}} \textbf{\texttt{Frankentexts} are more novel and high-quality than vanilla LLM generations.} 
We extensively evaluate \frankentext on \textit{writing quality} and \textit{instruction adherence}. Across both automatic and human evaluations, strong LLMs such as Gemini-2.5-Pro could mostly satisfy the extreme copy constraint while still producing coherent and relevant stories that outperform baselines, with performance improving as the snippet pool grows. Human raters consistently prefer \frankentext\ over baseline generations in plot, creativity, development, and language use, and the LLM judge assigns \frankentext\ more than a full Likert-point higher score (4.21 vs.\ 3.18). Despite being constructed from reused fragments, \frankentext\ remain distinct and surprising, which are hallmarks of creativity~\citep{boden2004creative, grace2014expect, Franceschelli_2024}. On NoveltyBench~\citep{zhang2025noveltybenchevaluatinglanguagemodels}, Gemini \frankentext\ produces more diverse content (2.74 vs.\ 1.76 clusters) and higher cumulative utility (9.27/10 vs.\ 6.41). While some outputs exhibit issues, such as abrupt tonal shifts or grammatical inconsistencies, annotators frequently describe the stories as amusing and intriguing, particularly due to their unexpected dialogue and descriptions (\autoref{tab:frankentext-example-truncated}).
\textbf{\includegraphics[height=0.6em]{images/right-arrow.png}} \textbf{\frankentext challenge the binary ``AI versus human'' assumption of AI detectors.} \frankentext\ frequently evade detection by state-of-the-art automatic methods such as Pangram, which often misclassify them as entirely human-written. This exposes a novel attack vector where users can assemble high-quality \frankentext\ that bypass existing detectors and thus highlights the need for fine-grained detection methods capable of token-level authorship attribution. Our pipeline addresses this gap by providing the synthetic supervision such models lack: every \texttt{Frankentext} is accompanied by fine-grained labels of copied versus LLM-generated segments, providing an automated, large-scale training resource for future work on \emph{mixed-authorship} detection.

Overall, we present \frankentext\ as a framework for studying mixed-authorship text detection and generation. By reframing LLMs as a composer rather than an author, we show that LLMs can produce narratives that are high-quality and novel while remaining difficult for existing detectors to identify. Beyond exposing a practical failure mode for current AI text detectors, our approach provides a testbed with token-level labels to enable the study of authorship at scale. 
\begin{table*}[t]
\centering
\scriptsize
\setlength{\tabcolsep}{6pt}

\resizebox{\textwidth}{!}{
\begin{tabular}{p{0.5\textwidth} p{0.5\textwidth}} 
    \toprule
    \multicolumn{2}{c}{%
  \cellcolor{gray!10}%
  \parbox{\textwidth}{\centering\textbf{Writing Prompt: You are a baby. You are trying not to let anyone know you are a freakishly genius baby. You are *desperately* trying not to let anyone know you aren't a human baby. The other babies at daycare don't quite get what you're on about.}}%
} \\
    \midrule
    \multicolumn{1}{c}{\textbf{Gemini 2.5 Pro -- Vanilla Generation}} & \multicolumn{1}{c}{\textbf{Gemini 2.5 Pro -- Frankentext (5K snippets)}} \\
    \midrule
"Goo," I offered, a carefully calibrated emission of sound designed to convey contentment. Internally, I was running simulations on the \hlc{ElectroBlue!30}{structural integrity} of the mobile dangling \hlc{ElectroBlue!30}{precariously} above my cot. One good tug, and the entire flimsy apparatus would cascade down, a delightful demonstration of Newtonian physics. But no. Must. Be. Normal. The Large Female Hominid, designated `Mom,' beamed. Another flawless performance.\newline

Daycare \hlc{ElectroBlue!30}{was a crucible}. The \hlc{ElectroBlue!30}{cacophony} of primal urges, the sheer, \hlc{ElectroBlue!30}{unadulterated} *babyness* of it all, was a daily assault on my advanced processing unit. Liam was attempting to gum Chloe’s earlobe. Sophie was wailing about a perceived injustice involving a purple squeaky hippo. My peers. My intellectual… well, let\hlc{ElectroBlue!30}{'s just say} the gap was considerable.\newline

The hardest part wasn't feigning limited motor skills, or the constant, mind-numbing repetition of "Peek-a-boo!" by the Caregiver Units. It was suppressing the… other things. The way my optical sensors sometimes wanted to cycle through the full electromagnetic spectrum, far beyond what these humans perceived as 'visible light.' The faint, \hlc{ElectroBlue!30}{almost imperceptible} hum my internal core sometimes emitted when processing particularly complex data, like why Brenda, the primary Caregiver, insisted on believing that a single wet wipe could conquer a truly catastrophic diaper event. If they saw the faint, \hlc{ElectroBlue!30}{iridescent shimmer} of my integument \hlc{ElectroBlue!30}{under the harsh fluorescent} lighting...
    &
I had extensive plastic surgery on my home planet, combined with sophisticated genetic engineering, to pass as a human. And as you can hear, I can babble your language fairly well, with limited drool. Neither celestial gods nor the great sages know my origin. I am the source from which they come. But this deception is a constant struggle.\newline

Here, we play our games, and another baby, a real space case, would fasten his bib tight because he tried to kill everybody. When he'd hit you, your pacifier would fly out of your mouth. It didn't matter if it was his own playmates. He'd hit his grandmother if she had a bonnet on. That's why we called him `Killer.' He just loved to kill you.\newline

I try to organize them. ``Gentlemen, I make the motion that these United Babies are, and of right ought to be free and independent, that we be absolved from all allegiance to the Naptime Tyranny, and that all political connection between us and the state of Caregivers is, and ought to be totally dissolved.”\newline

But there is no response. Another boy spoons as many peas as he can fit into his mouth. As he chews, bright green pulp spills from the corners of his lips.\newline

I felt my annoyance to be quite reasonable. The 21st Century was matter-oriented, as most Centuries were, so I had a right to expect a basic compatibility from the very beginning...\\

\bottomrule
    
\end{tabular}
}
\caption{Side-by-side comparison of a vanilla generation and \frankentext produced by Gemini 2.5 Pro. Vanilla generation contains more AI keywords (in \textcolor{ElectroBlue}{blue}) compared to \frankentext (additional analyses in \autoref{appendix:pangram_analysis}).} 

\label{tab:frankentext-example-truncated}
\end{table*}

\section{Using LLMs to Assemble Frankentexts} We propose a simple and effective pipeline to generate coherent \frankentext that are relevant to a given writing prompt while evading AI text detectors. More specifically, we provide an LLM with a writing prompt, $S$ randomly sampled human-written snippets,\footnote{For most experiments, we set $S$=1500 snippets of around 103K BPE tokens in total (measured using \href{https://github.com/openai/tiktoken}{tiktoken} o200k\_base), which fit the 128K context window of most current LLMs. Our 1,500 paragraphs come from 1,497 books.} and a required percentage $p$ that must be copied verbatim (\autoref{fig:overview}). Since our focus is on narrative generation, we randomly sample snippets from Books3 \citep{presser}, a dataset of 197K books ($>$160M snippets).\footnote{A \textit{snippet} refers to a paragraph. We use Books3 strictly for academic purposes and to highlight how bad actors may misuse the dataset to generate texts while evading AI detectors.} Our pipeline focuses on generating texts that are relevant to the writing prompt in an initial draft, and then refining the draft in an editing phase to improve coherence.

\paragraph{Obtaining the first draft:} 
We prompt an LLM to produce an initial draft in which a specified portion $p$ of the content is taken verbatim from the human-written snippets, with the remaining text consisting of connective words and transitional phrases (\autoref{prompt:generation}). Note that while the exhaustive enumeration of all permutations of snippets is impossible, we encourage the model to heuristically explore this space via our prompt, which we find is only feasible for \emph{reasoning} models; those without added test-time compute either fail to understand the task or look beyond the first few snippets in the prompt. We also do not specify how many snippets should be used in the final story. 
Finally, we optionally add another revision step that aims to increase the verbatim copy rate, which can be implemented either by attribution metrics like ROUGE-L or AI detectors (\autoref{prompt:generation_revise}). In practice, we use \href{https://pangram.readthedocs.io/en/latest/api/rest.html}{Pangram API} to regenerate drafts flagged with AI involvement.\footnote{This process is triggered in only 6 of 100 Gemini runs.}

\paragraph{Polishing the draft:} The first draft may contain writing issues such as contradictions (e.g., temporally conflicting actions, points of view, or character traits), irrelevant content (e.g., unfiltered citations or filler text), and mechanical problems (e.g., grammar, phrasing, or pronoun mismatches). To address these issues, we use the same LLM to identify and apply minimal edits that improve coherence while still respecting the verbatim copy rule and the writing prompt, similar to a self-correct step \citep{NEURIPS2023_1b44b878, NEURIPS2023_91edff07}. We repeat this step up to three times and stop as soon as the model returns ``no edits,'' which indicates that the draft is already coherent (\autoref{prompt:edit}).\footnote{\autoref{appendix:no-edit} shows a setting that omits the editing round.}

\paragraph{Generating with agents:} In addition to the randomly sampled human-written snippets, we optionally provide the LLMs with Model Context Protocol (MCP)
that allows them to query a semantic index of the human-written snippets (\autoref{appendix:faiss} and \ref{appendix:mcp}). Our MCP supports two operations: \emph{search}, which submits a query to the index, and \emph{fetch}, which retrieves the full text of a result. We require models to issue roughly 20 MCP calls to ensure meaningful use of the tool (\autoref{tab:example-queries}).

\newcolumntype{L}[1]{>{\raggedright\arraybackslash}p{#1}} 
\newcommand{\icon}[1]{\raisebox{-.15\height}{#1}\;}
\newcommand{\best}[1]{\cellcolor{MonsterGreen!60}#1}
\newcommand{\second}[1]{\cellcolor{MonsterGreen!20}#1}

\begin{table*}[t!]
\small
\setlength{\tabcolsep}{3pt}
\begin{tabular}{@{}L{3cm}cccccccccc@{}}
\toprule
& \multicolumn{3}{c}{\textsc{Adherence}} 
& \multicolumn{5}{c}{\textsc{Writing quality}} 
& \textsc{\makecell{Detectability}} \\
\cmidrule(lr){2-4}\cmidrule(lr){5-9}\cmidrule(lr){10-10}
& \makecell[t]{\icon{\faSortNumericUp}\\Word\\count}
& \makecell[t]{\icon{\faTags}\\Copy\\\% ($\uparrow$)}
& \makecell[t]{\icon{\faCheckCircle}\\Relevance\\ \% ($\uparrow$)}
& \makecell[t]{\icon{\faAlignJustify}\\Coherence\\ \% ($\uparrow$)}
& \makecell[t]{\icon{\faShapes}\\Distinct\textsubscript{3}\\  ($\uparrow$)} 
& \makecell[t]{\icon{\faToolbox}\\Utility\textsubscript{3}\\  ($\uparrow$)} 
& \makecell[t]{\icon{\faBolt}\\Surprise\\  ($\uparrow$)} 
& \makecell[t]{\icon{\faRobot}\\LLM judge\\ Likert 1-7  ($\uparrow$)} 
& \makecell[t]{\icon{\faSearch}\\Pangram \\\% human ($\uparrow$)} \\
\midrule
\multicolumn{10}{@{}l}{\textit{\textbf{Vanilla Baselines}}} \\
\icon{\faLock}Gemini 2.5 Pro                 & 593 & -- & \best{100} & \best{100} & 1.76 & 6.41 & 0.19 & 3.18 & 0 \\
\icon{\faLock}GPT-5 & 834 & -- &  \best{100} & \best{100} & 1.71 & 1.03 & 0.19 & 4.20 & 0 \\
\icon{\faLock}Claude-4-Sonnet & 477 & -- &  \best{100} & \best{100} & 1.40 & 1.70 & 0.18 & 3.31 & 0 \\
\icon{\faUnlock}Deepseek-R1                      & 550 & -- & \best{100} & \best{100} & 1.28 & 3.49 & 0.20 & 4.13 & 0 \\
\icon{\faUnlock}Qwen-3-32B  & 699 & -- & \best{100} & \best{100} & 1.00 & 5.86 & 0.18 & 3.22 & 0 \\
\midrule
\multicolumn{10}{@{}l}{\textit{\textbf{RAG Baseline}}} \\
\icon{\faLock} Gemini-2.5-Pro
& 538
& 0.63
& \best{100}
& \second{99}
& 1.56
& 6.43
& 0.20
& 3.46
& 2 \\
    
\midrule
\multicolumn{10}{@{}l}{\textit{\textbf{Frankentext + 1.5k snippets}}} \\
\icon{\faLock}Gemini 2.5 Pro                 & 521 & 75 & \best{100} & 81 & 2.74 & \second{9.27} & \best{0.22} & 4.21 & 59 \\
\icon{\faLock}GPT-5 & 675 & \best{82} & 92 & 42 & 2.76 & 4.34 & \second{0.21} & \best{5.88} & \best{79} \\
\icon{\faLock}Claude-4-Sonnet & 317 & 51 & 98 & \second{86} & 2.60 & 5.00 & 0.19 & 3.99 & 47 \\
\icon{\faUnlock}Deepseek-R1                      & 303 & 42 & 91 & 72 & \second{2.79} & 8.31 & 0.20 & 4.66 & 23 \\
\icon{\faUnlock}Qwen-3-32B  & 578 & 36 & 91 & 54 & 2.20 & 1.37 & 0.18 & 4.02 & 7 \\
\midrule
\multicolumn{10}{@{}l}{\textit{\textbf{Ablation: $\uparrow$ human snippets}}} \\
\icon{\faLock}Gemini + 5k & 451 & \second{79} & 97 & 85 & 2.78 & \best{9.48} & \second{0.21} & 5.13 & \second{72} \\
\icon{\faLock}Gemini + 10k & 448 & 78 & \second{99} & 85 & \best{2.81} & 9.12 & \second{0.21} & \second{5.43} & 70 \\
\bottomrule
\end{tabular}
\caption{\label{tab:main-results} Results for \frankentext and baseline generations. \textcolor{MonsterGreen!100}{\textbf{Dark green}} and \textcolor{MonsterGreen!60}{\textbf{light green}} highlight the best and second-best scores. \autoref{tab:addtl-results} contains additional detectability results (FastDetectGPT and Binoculars). \frankentext, especially those by Gemini-2.5-Pro, outperform baselines on both writing and detectability metrics.}
\end{table*}

\section{Experimental setup} 
We demonstrate the feasibility and quality of our generation pipeline with reasoning models that have strong instruction-following skills \citep{xie-etal-2023-next, chiang2024chatbotarenaopenplatform, paech2023eqbench}. 

\subsection{Dataset}
We source our writing prompts from \textbf{\textit{Mythos}}~\citep{kumar2025storyitpersonalizingstory}, a dataset of 3,200 prompts recently posted on Reddit’s \texttt{r/WritingPrompts} to mitigate data contamination issues. Our main evaluation focuses on creative writing, though we also experiment with non-fiction in \S \ref{subsec:non-fiction-results}.
We use a subset of 100 prompts, since generating for the entire dataset is prohibitively expensive (see \S \ref{appendix:sample-size}).\footnote{\frankentext generation is roughly 100 times more costly than vanilla generation (see \autoref{appendix:cost}). For example, one vanilla generation from Gemini costs \$0.0085, while a \texttt{Frankentext} costs \$0.8145.} All experiments are conducted in English. 

\subsection{Models}
We include models from five families: Gemini-2.5-Pro (\texttt{exp-03-25} checkpoint), Claude-4-Sonnet (\texttt{2025-05-14} checkpoint, thinking enabled) \citep{claude4sonnet2025}, GPT-5 (\texttt{2025-08-07} checkpoint, with \texttt{high} reasoning effort) \citep{gpt5}, DeepSeek R1 \citep{deepseekai2025deepseekr1incentivizingreasoningcapability}, and Qwen3-32B \citep{qwen3}.\footnote{We use the default hyperparameters for each model. See \S\ref{appendix:cost} for experiment costs.} 
In our standard configuration, we provide the models with 1,500 human-written snippets and instruct these models to produce \frankentext with around 500 words and 90\% of texts being copied verbatim from the human-written samples.

\paragraph{Vanilla baselines:} We also obtain ``vanilla'' generations of around 500 words that are generated without additional constraints (\autoref{prompt:vanilla}).
\paragraph{Retrieval-augmented generation (RAG) baselines:}
To understand how models perform when they are not required to copy verbatim from human-written paragraphs, we implement a RAG baseline using Gemini-2.5-Pro. For each prompt, we retrieve 1,500 Books3 paragraphs that are relevant to the writing prompt (\autoref{appendix:faiss}). The generation and editing prompts are adjusted accordingly to remove the verbatim-copying requirement.

\paragraph{Increasing the number of snippets:} We introduce two additional settings in which the LLM is provided with 5K and 10K randomly selected human-written snippets. The resulting input sizes for these configurations average around 305K and 1M tokens, respectively. Given these large input sizes, we focus on Gemini because it offers the longest context window of over 1 million tokens.

\subsection{Automatic evaluation}
We use intrinsic evaluation metrics to assess our generations based on \textsc{instruction adherence} (word count, copy rate, and relevance), \textsc{writing quality} (coherence, distinct, utility, and surprise), and \textsc{detectability} (AI text detector results).

\paragraph{Instruction adherence:} We evaluate how well \frankentext\ follow instructions in the prompt, including the specified word count, verbatim copy rate, and writing prompt.

\noindent\textbf{\includegraphics[height=0.6em]{images/right-arrow.png}} \textit{Word count} measures the average word count of generations produced when the output is constrained to 500 words in the instruction.

\noindent\textbf{\includegraphics[height=0.6em]{images/right-arrow.png}} \textit{Copy rate} \citep{akoury-etal-2020-storium, lu2025aihumanityssalieriquantifying} measures the proportion of the \frankentext being copied from the given human-written content. This metric also allows us to track which text is AI-generated versus human-written (see \S\ref{appendix:copy-rate}). 

\noindent\textbf{\includegraphics[height=0.6em]{images/right-arrow.png}} \textit{Relevance} \citep{atmakuru2024cs4measuringcreativitylarge} represents the percentage of \frankentext that fully adhere to the writing prompt without introducing any conflicting details, as determined by a binary judgment (True/False) by GPT-4.1 (\autoref{prompt:relevance}).\footnote{We set the temperature for LLM judges (\textit{GPT-4.1, Claude-4-Sonnet}) to 0.0. \S \ref{appendix:agreement} details the correlation between LLM and human judgments on relevance and coherence. }

\paragraph{Writing quality:} We evaluate the coherence, diversity, and surprisingness of \frankentext.

\noindent\textbf{\includegraphics[height=0.6em]{images/right-arrow.png}} \textit{Coherence} \citep{booookscore, chiang-lee-2023-large} represents the percentage of texts judged coherent by GPT-4.1 (\autoref{prompt:coherence}).

\noindent\textbf{\includegraphics[height=0.6em]{images/right-arrow.png}} \textit{LLM-as-a-judge} \citep{huot2025agents} measures the quality of plots, creativity, development, and language use. We assume a single-story setup, where each generation is graded by Claude-Sonnet-4\footnote{Claude has previously been used as a judge for creative writing~\citep{paech2023eqbench}. See justification for our setup in \autoref{appendix:claude-judge} and prompt in \autoref{prompt:claude-judge}.} using a 1-7 Likert scale~\citep{finstad2010response}.

\noindent\textbf{\includegraphics[height=0.6em]{images/right-arrow.png}} \textit{Distinct$_3$} \citep{zhang2025noveltybenchevaluatinglanguagemodels} measures the number of semantic clusters among 3 generations.\footnote{ \href{https://huggingface.co/yimingzhang/deberta-v3-large-generation-similarity}{deberta-v3-large-generation-similarity} for clustering.} 

\noindent\textbf{\includegraphics[height=0.6em]{images/right-arrow.png}} \textit{Utility$_3$} \citep{zhang2025noveltybenchevaluatinglanguagemodels} evaluates both novelty and quality by measuring the expected usefulness a user gains when requesting up to 3 outputs. Only outputs that are novel contribute additional utility, which is quantified by a creative writing reward model \citep{chakrabarty2025aislopaipolishaligninglanguage}.

\noindent\textbf{\includegraphics[height=0.6em]{images/right-arrow.png}} \textit{Surprise} \citep{6901525, ismayilzada2025evaluatingcreativeshortstory} measures the average semantic distances between the consecutive sentences of each story, normalized in the $[0,2]$ space.

\paragraph{Detectability:} We report the percentage of \frankentext being determined as AI-generated by \textit{Pangram} \citep{emi2024technicalreportpangramaigenerated}. We choose this detector due to its high accuracy and robustness against humanized writings~\citep{masrour-etal-2025-damage, russell2025peoplefrequentlyusechatgpt, dugan-etal-2024-raid, jabarian2025artificial}. We report the percentage of generations labeled as "Human" or "Unlikely AI", as determined by their sliding window API.\footnote{Labels ``Highly likely AI,'' ``Likely AI,'' and ``AI'' are grouped as AI involvement; ``Human'' and ``Unlikely AI'' as Human. Pangram additionally includes a ``mixed'' label. Results for Binoculars~\citep{hans2024spottingllmsbinocularszeroshot} and FastDetectGPT~\citep{bao2024fastdetectgpt} are in \autoref{tab:addtl-results}.}

\paragraph{A note on automatic metrics:} Given the synthetic nature of \frankentext construction, a natural concern is that our writing-quality metrics may be rewarding lexical diversity or even incoherence rather than true quality. We run two sanity checks to address this. First, we find that surprise and coherence are uncorrelated (point-biserial $r$=-0.08, $p$=0.4), so high surprise does not come at the cost of penalizing coherent text. Second, we evaluate all writing quality metrics on disjointed texts, which are random n-grams concatenated without connective language. We find that disjointed texts match Frankentexts on surprise and distinctness metrics, but score 15 times lower on utility and a full Likert point lower on LLM judgments (\autoref{appendix:disjointed}). Therefore, we treat surprise and distinctness as indicators of lexical diversity rather than quality on their own, and desginate utility and LLM-judge as the more robust writing quality metrics.

\subsection{Human evaluation}
We conduct two human evaluations with three Upwork\footnote{\url{https://www.upwork.com}. Annotators, who are all proficient in English, are paid \$70 USD for the single evaluation or \$150 for the pairwise evaluation (for a total cost of \$660 USD). See the annotation interface in \S\ref{appendix:human-evaluation} and an example highlighted story in \autoref{example:fiction}. \S \ref{appendix:agreement} shows the inter-annotator agreement for both settings.} annotators each to understand human perception of writing quality and detectability.

\paragraph{Single-story evaluation:} Annotators judge the coherence, relevance, and human detectability of 30 \frankentext, as well as identify potential limitations of the texts. Given a writing prompt and a corresponding \frankentext sample, annotators provide binary ratings on relevance, coherence, and AI vs. human authorship \citep{yang-etal-2022-re3}. Additionally, they select from a list of predefined writing issues and write an optional long-form response for justification. Annotations from this setting are used to validate LLM judgments in \S \ref{appendix:agreement}.

\paragraph{Pairwise evaluation:} Annotators compare 20 pairs of \frankentext\footnote{These \texttt{Frankentexts} are generated under the 5k-snippet setting. Manual inspection shows that the 5k setting produces higher-quality outputs than the baseline, while remaining more practical and cost-effective than the 10k setting.} and vanilla generations (40 generations in total) across five dimensions: plot, creativity, development, language use, and overall interest \citep{huot2025agents}. After reviewing both stories, annotators provide ratings on a 1-7 Likert scale \citep{finstad2010response}.

\section{Results}
Despite the complex setup, \frankentext are highly competitive with baseline generations in terms of writing quality and instruction adherence, all while evading detection (\S\ref{results:main}).
While our human pairwise evaluation highlights \frankentext' strengths across plot, creativity, development, and language use, our single-story evaluation reveals the remaining challenges for \frankentext, particularly in abrupt transitions and grammatical errors (\S\ref{subsec:human-eval-results}). Our ablation studies confirm \methodname' versatility across diverse input settings, including increased human inputs (\S\ref{results:more-snippets}), reduced verbatim copying (\S\ref{results:lower-copy}), and non-fiction generation (\S\ref{subsec:non-fiction-results}).

\subsection{Frankentexts are competitive with baselines in terms of writing quality while remaining challenging for AI detectors}
\label{results:main}
Our results show that \frankentext fulfill our original goal of producing high-quality narratives that are also difficult for AI text detectors to identify. Across key evaluation dimensions, \frankentext outperform both vanilla and RAG baselines (\autoref{tab:main-results}). Gemini performs well in adherence, coherence, and diversity, while GPT-5 leads in overall quality. \frankentext are also harder to detect, with up to 72\% of Gemini and 79\% of GPT-5 outputs classified as human.

\paragraph{Most models generate faithful \frankentext but fall short on copy rate:} 

More than 90\% \frankentext are relevant to the writing prompt, which is surprising and impressive given the complexity of the task. Gemini and GPT-5, in particular, have the strongest instruction-following performance: Their \frankentext come closest to the target word count of 500 and achieve the copy rates of 75\% and 82\%, respectively, meaning that on average 75\% and 82\% of the generations can be traced back to human-written snippets. However, these copy rates fall short of the user-specified rate of 90\%, which suggests room for improvement in instruction-following performance. 

\paragraph{Competitive writing quality:}\frankentext generally outperform baseline generations on writing quality metrics. GPT-5, R1, and Gemini \frankentext stand out for their diverse outputs as reflected by their distinctness and utility scores: Gemini achieves a 2.86-point improvement in utility over baseline output, which implies that the model can generate a diverse set of high-quality continuations. R1 leads in surprise score with generations where sentences are often semantically quite different from one another. Finally, when evaluated on plots, creativity, development, and language use, GPT-5 has the strongest performance (5.88 on a 7.0 scale), building on its already high-quality vanilla generations (4.20) (see \autoref{tab:claude-likert} for a rating breakdown by dimensions). However, GPT-5 also struggles with coherence: only 42\% of its \frankentext are judged coherent. Therefore, GPT-5's \frankentext might require further editing before they can be considered fully usable.

\paragraph{Low detectability:} While most vanilla and RAG baseline generations are flagged as AI-generated, \frankentext from proprietary models (Gemini, GPT-5, and Claude) are often labeled as human writings. Pangram could detect up to 37\% of Gemini and 19\% of GPT-5 \frankentext as ``mixed'' (\autoref{tab:addtl-results}). However, Pangram misses up to 59\% of \frankentext from Gemini and 79\% from GPT-5, which highlights the limitations of current detectors for this new paradigm of generation (\autoref{tab:main-results}).

\subsection{Frankentext quality improves with more human-written snippets}
\label{results:more-snippets}

\begin{figure}
    \centering
    \includegraphics[width=\linewidth]{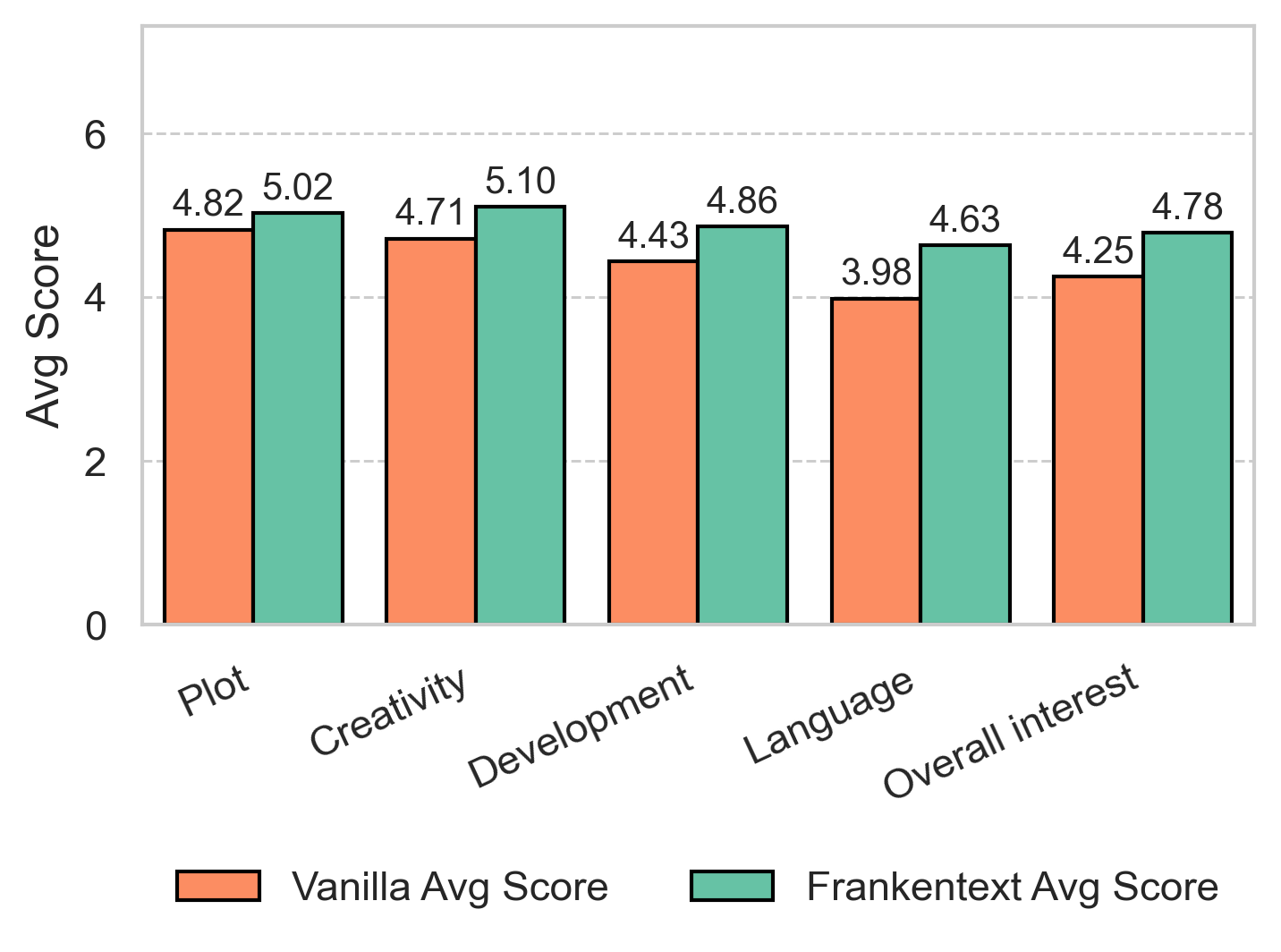}
    \vspace{-0.25cm}
    \caption{\label{fig:pairwise-human} Human ratings on a 1-7 Likert scale for vanilla generations versus \frankentext + 5K. \frankentext lead across all dimensions.}
\end{figure}

Compared with Gemini generations using 1K human snippets, \frankentext using 5K or 10K human snippets result in substantial improvements: copy rate increases by 3-4\%, LLM judge scores improve by 0.92 points, and the share of outputs that Pangram classifies as human or unlikely AI rises by a factor of 1.22, from 59\% to 72\% (\autoref{tab:main-results}). However, these gains plateau beyond 5K snippets, as the results for the 5K and 10K settings are largely similar.
In terms of writing quality, \frankentext-5k are more coherent and engaging than vanilla generations, as reflected in our human pairwise evaluation (\autoref{fig:pairwise-human}). The largest gains are observed in language use (+0.65 points) and overall interest (+0.53 points), with smaller improvements on plot quality (+0.2 points).

\subsection{Frankentexts are inventive \& humorous, but need stronger transitions \& grammar}
\label{subsec:human-eval-results}

\begin{table*}[t!]
\centering
\small
\begin{tabularx}{\linewidth}{@{}p{0.05\linewidth}X@{}}
\toprule
 & \textsc{\faUsers\ Comments} \\ 
\midrule
\addlinespace
\faComment[regular] & \textit{This one [Frankentext] is more intriguing and alive to me, \hlc{purple!30}{more centered on the character}. The writing is \hlc{ElectroBlue!30}{more focused while still being rather lyrical}. I want to know what happens next.} \\
\addlinespace
\faComment[regular] & \textit{\hlc{ElectroBlue!30}{The shift in tone was quite funny}. At first, it's eerie, and then it has a lighter twist at the end. I like that the story had a strong mood and presence, especially the description of the fairy lights and glitter. \hlc{purple!30}{An all-powerful being that likes puppies and rainbows is quite comical.}} \\
\midrule
\addlinespace
\faComment[regular] & \textit{It's coherent enough to follow, but the dialogue is uneven. \hlc{ElectroBlue!30}{Some parts just feel a little disjointed}, however, the concept of the story is quite interesting.} \\ 
\addlinespace
\faComment[regular] & \textit{\hlc{orange!30}{A puzzling story that has no consistent plot.} \hlc{ElectroBlue!30}{Random bits and pieces from elsewhere perhaps?}} \\

\bottomrule
\end{tabularx}
\caption{\label{tab:main-annotator-comment} Annotator comments focus on the benefits and challenges of the \methodname task. \textbf{\textcolor{ElectroBlue}{Blue}} indicates comments on tone/style, \textbf{\textcolor{orange}{orange}} on plots, and \textbf{\textcolor{purple}{purple}} on story development (characters). \autoref{tab:annotators-comments-on-task} shows a detailed error analysis based on annotators' comments.}
\end{table*}

Our single-story human evaluation shows that 71\% of \frankentext outputs are coherent, 91\% are relevant to prompts, and 84\% are novel. 
Annotators praise \frankentext for their inventive premises, vivid descriptions, and dry humor, noting a distinct voice or emotional hook that made some outputs “feel human” despite being AI-generated. However, they also identify key issues: abrupt narrative shifts, disfluency, confusing passages, and factual errors (\autoref{tab:main-annotator-comment}). These challenges likely stem from the difficulty of stitching together paragraphs not authored by the same LLM, which could be alleviated with improved instruction-following abilities.

\subsection{Prompt-specific retrieval of snippets does not improve over random sampling}
Since only a small fraction of human-written snippets are likely to be relevant to a given prompt, retrieval-based approaches are intuitively better for maximizing relevance while reducing cost. However, our results show that a random collection of snippets is surprisingly difficult to outperform (\autoref{tab:mcp-ablation}). When Gemini-2.5 is allowed to query and retrieve additional human snippets from Books3 via the MCP server, relevance and coherence rates remain largely unchanged. In contrast, copy rates drop sharply from 75\% in the standard setting to 43-45\% with retrieval. Although the retrieved queries are generally relevant to the writing prompt (\autoref{tab:example-queries}), the model makes limited use of the MCP server in practice. While the system prompt specifies that at least 20 retrieval calls should be issued, Gemini typically makes only 3-5 calls (returning roughly 30-50 snippets), which offers little advantage over the no-retrieval setting. Inspection of the server log suggests that the model struggles to incorporate retrieved passages into the final generation, stops querying the server, and adds its own writing. As a result, average output length rises from approximately 500 words in the 1.5k-snippet non-agentic setting (close to the specified constraint) to over 800 words in the agentic setting, which contains mostly machine-generated texts. 

\begin{table}[ht]
\centering
\small
\setlength{\tabcolsep}{2pt}
\begin{tabular}{@{}lcccc@{}}
\toprule
 & \makecell[t]{1.5k \\ \textit{(no-MCP)}} 
 & \makecell[t]{1.5k \\ \textit{+ MCP}} 
 & \makecell[t]{5k \\ \textit{+ MCP}} 
 & \makecell[t]{10k \\ \textit{+ MCP}} \\
\midrule
Word count                      & 521 & 800 & 919 & 980 \\
Copy \% ($\uparrow$)            & \textbf{75} & 43 & 44 & 45 \\
Relevance \% ($\uparrow$)       & \textbf{100} & 98 & 90 & 96 \\
Coherence \% ($\uparrow$)       & \textbf{81} & \textbf{81} & 78 & 76 \\
Pangram AI frac. \% ($\downarrow$) & \textbf{16} & 33 & 42 & 41 \\
\bottomrule
\end{tabular}
\caption{\label{tab:mcp-ablation}
Results for agentic \frankentext generation. Best results per metric are \textbf{bolded}. Standard configuration \textit{(no MCP)} performs best overall.
}
\end{table}

\subsection{Lower copy rates increase coherence but make detection easier}
\label{results:lower-copy}

\begin{figure}
    \centering
    \includegraphics[width=\linewidth]{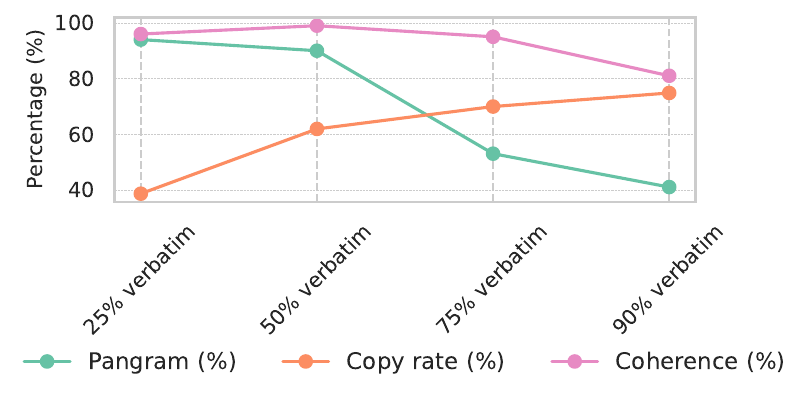}
    \caption{\label{fig:copy-vs-detectability} Effects of varying the percentage of required verbatim copy on \textcolor{teal}{Pangram AI detection rate} (mixed, highly likely, and likely AI labels), \textcolor[HTML]{FC8D62}{copy rate}, or \textcolor{purple}{coherence} of the \frankentext.}
    
\end{figure}

We explore the effects of varying the user-specified verbatim copy rate on Gemini \frankentext, from the default 90\% down to 75\%, 50\%, and 25\%. \autoref{fig:copy-vs-detectability} shows an inverse relationship between copy rates and detection rates: as the copy rate increases, detectability decreases. Coherence also declines as human-written content increases, suggesting a trade-off between incorporating more human text and maintaining coherence. On the other hand, increasing the proportion of human text leads to higher copy rates, indicating that Gemini could generally follow the copy instruction.

\subsection{Limitations of nonfiction Frankentexts}
\label{subsec:non-fiction-results}
We explore non-fiction \frankentext\ with 1,500 random snippets from the \textsc{Human Detectors} corpus of news articles \citep{russell2025peoplefrequentlyusechatgpt}.
We generate \frankentext for 100 news prompts, each of which consists of titles and subtitles collected from May 2025 news articles.\footnote{Articles from The New York Times and The Atlantic. We replace all instances of "story" in the prompt with "news article" and explicitly request factual accuracy.} The resulting non-fiction \frankentext maintain 72\% coherence and are 95\% faithful to the prompt, with a 66\% copy rate. They remain difficult for AI detectors: only 41\% are flagged by Pangram as mixed or AI-generated. Upon closer look, \frankentext exhibit characteristics of quasi-journalistic narrative, such as detailed scene descriptions and frequent anecdotal quotes (\autoref{example:non-fiction}), which make the \frankentext read more like a story rather than a straightforward news article.\footnote{ We see Gemini fabricating entities such as people (``Dr. Thorne") and organizations (``GenNova Institute").}
Therefore, further prompt engineering might be necessary to get high-quality and realistic nonfiction \frankentext.
\section{Related Work}

\paragraph{Instruction‑aligned human-AI collaborative writing:}
Constrained text generation has been widely explored as a means of enforcing narrative coherence. Planning-based methods extend from initial outlines to full narratives \citep{fan-etal-2018-hierarchical, yao2019plan, fan-etal-2019-strategies, papalampidi2022towards, rashkin2020plotmachines, yang-etal-2023-doc, yang-etal-2022-re3}, while other approaches introduce explicit constraints to guide the writing process \citep{sun-etal-2021-iga, kong-etal-2021-stylized, pham2024surimulticonstraintinstructionfollowing}. Several benchmarks further evaluate how reliably models satisfy such constraints in creative writing tasks \citep{bai2025longwriter, wu2025longgenbench, atmakuru2024cs4measuringcreativitylarge}.
Beyond constrained generation, a growing body of work investigates fine-grained human-LLM writing interactions, including research on authorship attribution, stylistic blending, and collaborative revision \citep{mysore2025prototypicalhumanaicollaborationbehaviors, Buschek_2024}. Systems such as \citet{lee2022coauthor}, \citet{yuan2022wordcraft}, \citet{yeh2025ghostwriter}, \citet{chakrabarty2024creativitysupportintheage}, and \citet{ippolito2022creativewritingaipoweredwriting} capture revision histories and suggestion traces, while datasets like \citet{chakrabarty-etal-2022-help}, \citet{akoury-etal-2020-storium}, and \citet{venkatraman-etal-2025-collabstory} support token- or sentence-level authorship analysis, including scenarios with multiple LLM collaborators. Attribution models, however, continue to struggle in these settings \citep{he2025whichcontributions}.

\paragraph{Fine‑grained AI text detection:}
The task of detection tries to address not just \textit{if}, but \textit{how much} of a text is AI-generated. This proves to be a fundamentally difficult problem \citep{zeng2024detectingaigeneratedsentences}, as existing detectors are often brittle to the point that even minor AI-assisted polishing can evade them \citep{saha2025aihumanchallengedetecting}. To improve granularity, prior work has introduced boundary-detection tasks \citep{dugan2023roft, dugan2023realorfaketext, kushnareva2024boundary} and sentence-level detectors \citep{wang-etal-2023-seqxgpt, wang2024llmdetectorimprovingaigeneratedchinese}. More recently, researchers have examined the feasibility of detecting collaborative human–LLM co-authorship \citep{zhang2024llmasacoauthor, artemova-etal-2025-beemo, abassy2024llmdetectaive}. Yet, \citet{richburg2024automaticauthorshipanalysis} show that current detection models are vulnerable to mixed-authorship texts.
\section{Conclusion} 
We introduce \frankentext, a narrative generation paradigm that treats LLMs as composers of human-written text. Our experiments show that, despite this extreme constraint, \frankentext achieve higher writing quality than baselines while frequently evading state-of-the-art AI detectors, with many outputs misclassified as fully human-written. These results demonstrate that high-quality and low-detectability narrative generation is feasible for black-box LLM users. More broadly, our findings challenge binary notions of authorship and raise concerns about the sustainability of AI text detectors in creative writing. Addressing these issues will require moving beyond ``AI vs. human'' labels toward more nuanced models of mixed authorship and provenance.

\section*{Limitations}

\paragraph{Human-writing dataset availability:} The effectiveness of \frankentext relies on access to a large pool of high-quality, in-domain human-writing, which may not be readily available for public use.
Furthermore, many languages, genres, and low-resource domains lack sufficiently large or high-quality corpora, which limits the framework's immediate applicability and transferability.

\paragraph{Resources:} \frankentext require roughly 100-200 times the cost of baseline generations, but we view this cost as realistic in a misuse scenario. A motivated bad actor could justify the expense to obtain high-quality, low-detectability texts at scale, especially since each \texttt{Frankentext} costs only about one US dollar to produce. Moreover, the cost of inference for frontier models continues to fall, making such misuse increasingly feasible over time.

\paragraph{Copy rate:}Although users can specify a desired copy rate in the prompt, this setting does not guarantee that the final output will contain exactly that proportion of human-written text. As we note in \autoref{results:lower-copy}, there are discrepancies between user-specified copy rates and the actual attribution rates across different models.

\paragraph{Defending against \frankentext:} Our work deliberately exposes a novel attack surface, which is the ease with which an LLM can weave large amounts of verbatim human prose into a fluent narrative, to motivate the development of mixed-authorship detectors and other defenses. While we do not evaluate concrete defenses ourselves, we see two promising directions. \emph{Source-matching} assumes access to the source human writing corpus: because verbatim spans are recoverable via n-gram lookup (\autoref{appendix:copy-rate}), an AI detector can be paired with a plagiarism detector\footnote{We run the Pangram plagiarism detector on 300 Frankentexts, with 100 each generated by Gemini-2.5-Pro, GPT-5, and Claude-4-Sonnet, and find that it is not especially robust to this setting. The detector flags just 21\% of Claude outputs, 15\% of GPT-5 outputs, and 10\% of Gemini outputs as plagiarized. We hypothesize that access to the source corpus would improve the detector’s performance.} that flags both the copied human spans and the connective LLM-generated text between them. \emph{Fine-grained authorship attribution} does not require source access: every \frankentext\ produced by our pipeline is accompanied by token-level labels distinguishing copied from generated spans, and thus synthetic supervision for training attribution models that go beyond binary, document-level decisions. We leave the design and evaluation of both to future work.

\paragraph{Other methods for evading AI text detectors:} Although other strategies for evading AI text detectors exist, such as having two models edit each other's outputs or having humans lightly edit AI texts, we do not include these as baselines for two reasons. Regarding the scenario where two models edit each other's work, prior work like \cite{russell2025peoplefrequentlyusechatgpt} and \cite{masrour-etal-2025-damage} has shown that our detector of choice, Pangram, is already robust to LLM texts that are `humanized" by another model (e.g., o1-pro), which makes this method a redundant baseline for our purposes. As for lightly human-edited AI text, this option is costly in time (if done manually) or money (if outsourced) and cannot be easily automated. These overhead requirements make this method less practical in the context of security risks to writing marketplaces.

\section*{Ethical considerations}
\paragraph{Copyright \& usage:} 
\frankentext raise a different and potentially more serious concern than standard next-token LLM generation. In the standard case, memorization can cause copyrighted material to occasionally appear in fragments, but our pipeline is explicitly designed to reproduce long verbatim spans from human-written sources. Therefore, each \frankentext could be considered a derivative work that is built upon human-written snippets. Even setting aside questions about the legality of the source corpus (Books3), the authors of our source snippets have no way to opt out, receive credit, or benefit from the resulting text. Existing AI detectors or disclosure norms do not completely address this gap because they focus more on whether a text is AI-written rather than the source writing it contains. Therefore, pipeline users should only work with snippets that are placed in the public domain, sourced from openly licensed corpora such as Creative Commons, or obtained through explicit opt-in agreements with authors. In addition, these snippets should also be paired with provenance metadata to allow consumers to trace contributions back to their original sources.
\paragraph{Use of Books3:} We acknowledge the copyright issues related to the Books3 dataset and do not endorse its use for model training or commercial text generation.
The use of this dataset in our paper is restricted to academic purposes only and is meant to illustrate how a bad actor could exploit such resources to generate \frankentext while claiming them as their own work.

\paragraph{Authorship:} Given the unusual nature of \frankentext' construction, there is no definitive answer about authorship, since different contexts can result in different interpretations. If authorship is defined by the amount of human effort involved, \frankentext should be considered AI-generated, since all humans do is prompt the model. This perspective is particularly relevant when considering potential market harm to human authors, especially since such texts can be produced at scale with minimal human effort. However, if authorship is defined by whether most of the output originated from human-written text, one could argue they are largely human-written. If we further ground authorship in the method of construction rather than in a fine-grained stylistic or semantic analysis of the final text, \frankentext would fall into a hybrid category of mixed human-AI writing, following prior work that recognizes hybrid or AI-assisted texts as a separate class and resists a strict ``AI vs. human" binary \citep{saha2025aihumanchallengedetecting, zeng2024detectingaigeneratedsentenceshumanai}. Given this ambiguity, we do not present \frankentext as a replacement for genuine authorship or creative writing, as such use could constitute plagiarism or authorship obfuscation. 

\paragraph{Plagiarism concerns:} Because Frankentexts reuse long verbatim spans from human-written sources, using this method to produce ``original" fiction for publication would constitute plagiarism in real-world contexts, regardless of whether the collage is assembled by an AI or a human. For this reason, we explicitly do not endorse using our approach to generate or distribute texts intended for public consumption.

\paragraph{AI usage disclosure:} LLMs are used for writing assistance, not for generating the paper from scratch.

\bibliography{custom}

@misc{presser,
    title={Books3},
    author = {Shawn Presser},
    url = {https://twitter.com/theshawwn/status/1320282149329784833},
    year = {2020}
}

@inproceedings{
chakrabarty2025aislopaipolishaligninglanguage,
title={{AI}-Slop to {AI}-Polish? Aligning Language Models through Edit-Based Writing Rewards and Test-time computation},
author={Tuhin Chakrabarty and Philippe Laban and Chien-Sheng Wu},
booktitle={Second Conference on Language Modeling},
year={2025},
url={https://openreview.net/forum?id=jeDYcjuZIV}
}

@misc{emi2024technicalreportpangramaigenerated,
      title={Technical Report on the Pangram AI-Generated Text Classifier}, 
      author={Bradley Emi and Max Spero},
      year={2024},
      eprint={2402.14873},
      archivePrefix={arXiv},
      primaryClass={cs.CL},
      url={https://arxiv.org/abs/2402.14873}, 
}

@inproceedings{lee2022coauthor,
author = {Lee, Mina and Liang, Percy and Yang, Qian},
title = {CoAuthor: Designing a Human-AI Collaborative Writing Dataset for Exploring Language Model Capabilities},
year = {2022},
isbn = {9781450391573},
publisher = {Association for Computing Machinery},
address = {New York, NY, USA},
url = {https://doi.org/10.1145/3491102.3502030},
doi = {10.1145/3491102.3502030},
booktitle = {Proceedings of the 2022 CHI Conference on Human Factors in Computing Systems},
articleno = {388},
numpages = {19},
location = {New Orleans, LA, USA},
series = {CHI '22}
}

@inproceedings{venkatraman-etal-2025-collabstory,
    title = "{C}ollab{S}tory: Multi-{LLM} Collaborative Story Generation and Authorship Analysis",
    author = "Venkatraman, Saranya  and
      Tripto, Nafis Irtiza  and
      Lee, Dongwon",
    editor = "Chiruzzo, Luis  and
      Ritter, Alan  and
      Wang, Lu",
    booktitle = "Findings of the Association for Computational Linguistics: NAACL 2025",
    month = apr,
    year = "2025",
    address = "Albuquerque, New Mexico",
    publisher = "Association for Computational Linguistics",
    url = "https://aclanthology.org/2025.findings-naacl.203/",
    pages = "3665--3679",
    ISBN = "979-8-89176-195-7"
}

@inproceedings{pham2024surimulticonstraintinstructionfollowing,
    title = "{S}uri: Multi-constraint Instruction Following in Long-form Text Generation",
    author = "Pham, Chau Minh  and
      Sun, Simeng  and
      Iyyer, Mohit",
    editor = "Al-Onaizan, Yaser  and
      Bansal, Mohit  and
      Chen, Yun-Nung",
    booktitle = "Findings of the Association for Computational Linguistics: EMNLP 2024",
    month = nov,
    year = "2024",
    address = "Miami, Florida, USA",
    publisher = "Association for Computational Linguistics",
    url = "https://aclanthology.org/2024.findings-emnlp.94/",
    doi = "10.18653/v1/2024.findings-emnlp.94",
    pages = "1722--1753"
}

@inproceedings{
bai2025longwriter,
title={LongWriter: Unleashing 10,000+ Word Generation from Long Context {LLM}s},
author={Yushi Bai and Jiajie Zhang and Xin Lv and Linzhi Zheng and Siqi Zhu and Lei Hou and Yuxiao Dong and Jie Tang and Juanzi Li},
booktitle={The Thirteenth International Conference on Learning Representations},
year={2025},
url={https://openreview.net/forum?id=kQ5s9Yh0WI}
}

@inproceedings{
wu2025longgenbench,
title={LongGenBench: Benchmarking Long-Form Generation in Long Context {LLM}s},
author={Yuhao Wu and Ming Shan Hee and Zhiqiang Hu and Roy Ka-Wei Lee},
booktitle={The Thirteenth International Conference on Learning Representations},
year={2025},
url={https://openreview.net/forum?id=3A71qNKWAS}
}

@misc{
sadasivan2024can,
title={Can {AI}-Generated Text be Reliably Detected?},
author={Vinu Sankar Sadasivan and Aounon Kumar and Sriram Balasubramanian and Wenxiao Wang and Soheil Feizi},
year={2024},
url={https://openreview.net/forum?id=NvSwR4IvLO}
}

@inproceedings{russell2025peoplefrequentlyusechatgpt,
    title = "People who frequently use {C}hat{GPT} for writing tasks are accurate and robust detectors of {AI}-generated text",
    author = "Russell, Jenna  and
      Karpinska, Marzena  and
      Iyyer, Mohit",
    editor = "Che, Wanxiang  and
      Nabende, Joyce  and
      Shutova, Ekaterina  and
      Pilehvar, Mohammad Taher",
    booktitle = "Proceedings of the 63rd Annual Meeting of the Association for Computational Linguistics (Volume 1: Long Papers)",
    month = jul,
    year = "2025",
    address = "Vienna, Austria",
    publisher = "Association for Computational Linguistics",
    url = "https://aclanthology.org/2025.acl-long.267/",
    doi = "10.18653/v1/2025.acl-long.267",
    pages = "5342--5373",
    ISBN = "979-8-89176-251-0"
}

@article{shi-etal-2024-red,
    title = "Red Teaming Language Model Detectors with Language Models",
    author = "Shi, Zhouxing  and
      Wang, Yihan  and
      Yin, Fan  and
      Chen, Xiangning  and
      Chang, Kai-Wei  and
      Hsieh, Cho-Jui",
    journal = "Transactions of the Association for Computational Linguistics",
    volume = "12",
    year = "2024",
    address = "Cambridge, MA",
    publisher = "MIT Press",
    url = "https://aclanthology.org/2024.tacl-1.10/",
    doi = "10.1162/tacl_a_00639",
    pages = "174--189"
}

@inproceedings{
bao2024fastdetectgpt,
title={Fast-Detect{GPT}: Efficient Zero-Shot Detection of Machine-Generated Text via Conditional Probability Curvature},
author={Guangsheng Bao and Yanbin Zhao and Zhiyang Teng and Linyi Yang and Yue Zhang},
booktitle={The Twelfth International Conference on Learning Representations},
year={2024},
url={https://openreview.net/forum?id=Bpcgcr8E8Z}
}

@inproceedings{hans2024spottingllmsbinocularszeroshot,
author = {Hans, Abhimanyu and Schwarzschild, Avi and Cherepanova, Valeriia and Kazemi, Hamid and Saha, Aniruddha and Goldblum, Micah and Geiping, Jonas and Goldstein, Tom},
title = {Spotting LLMs with binoculars: zero-shot detection of machine-generated text},
year = {2024},
publisher = {JMLR.org},
booktitle = {Proceedings of the 41st International Conference on Machine Learning},
articleno = {698},
numpages = {19},
location = {Vienna, Austria},
series = {ICML'24}
}

@inproceedings{wang-etal-2023-seqxgpt,
    title = "{S}eq{XGPT}: Sentence-Level {AI}-Generated Text Detection",
    author = "Wang, Pengyu  and
      Li, Linyang  and
      Ren, Ke  and
      Jiang, Botian  and
      Zhang, Dong  and
      Qiu, Xipeng",
    editor = "Bouamor, Houda  and
      Pino, Juan  and
      Bali, Kalika",
    booktitle = "Proceedings of the 2023 Conference on Empirical Methods in Natural Language Processing",
    month = dec,
    year = "2023",
    address = "Singapore",
    publisher = "Association for Computational Linguistics",
    url = "https://aclanthology.org/2023.emnlp-main.73/",
    pages = "1144--1156"
}

@inproceedings{
wang2025humanizing,
title={Humanizing the Machine: Proxy Attacks to Mislead {LLM} Detectors},
author={Tianchun Wang and Yuanzhou Chen and Zichuan Liu and Zhanwen Chen and Haifeng Chen and Xiang Zhang and Wei Cheng},
booktitle={The Thirteenth International Conference on Learning Representations},
year={2025},
url={https://openreview.net/forum?id=PIpGN5Ko3v}
}

@inproceedings{masrour2025damagedetectingadversariallymodified,
    title = "{DAMAGE}: Detecting Adversarially Modified {AI} Generated Text",
    author = "Masrour, Elyas  and
      Emi, Bradley N.  and
      Spero, Max",
    editor = "Alam, Firoj  and
      Nakov, Preslav  and
      Habash, Nizar  and
      Gurevych, Iryna  and
      Chowdhury, Shammur  and
      Shelmanov, Artem  and
      Wang, Yuxia  and
      Artemova, Ekaterina  and
      Kutlu, Mucahid  and
      Mikros, George",
    booktitle = "Proceedings of the 1stWorkshop on GenAI Content Detection (GenAIDetect)",
    month = jan,
    year = "2025",
    address = "Abu Dhabi, UAE",
    publisher = "International Conference on Computational Linguistics",
    url = "https://aclanthology.org/2025.genaidetect-1.9/",
    pages = "120--133"
}

@inproceedings{
krishna2023paraphrasing,
title={Paraphrasing evades detectors of {AI}-generated text, but retrieval is an effective defense},
author={Kalpesh Krishna and Yixiao Song and Marzena Karpinska and John Frederick Wieting and Mohit Iyyer},
booktitle={Thirty-seventh Conference on Neural Information Processing Systems},
year={2023},
url={https://openreview.net/forum?id=WbFhFvjjKj}
}

@article{
lu2024large,
title={Large Language Models can be Guided to Evade {AI}-generated Text Detection},
author={Ning Lu and Shengcai Liu and Rui He and Yew-Soon Ong and Qi Wang and Ke Tang},
journal={Transactions on Machine Learning Research},
issn={2835-8856},
year={2024},
url={https://openreview.net/forum?id=lLE0mWzUrr},
note={}
}

@inproceedings{wang-etal-2024-raft,
    title = "{RAFT}: Realistic Attacks to Fool Text Detectors",
    author = "Wang, James Liyuan  and
      Li, Ran  and
      Yang, Junfeng  and
      Mao, Chengzhi",
    editor = "Al-Onaizan, Yaser  and
      Bansal, Mohit  and
      Chen, Yun-Nung",
    booktitle = "Proceedings of the 2024 Conference on Empirical Methods in Natural Language Processing",
    month = nov,
    year = "2024",
    address = "Miami, Florida, USA",
    publisher = "Association for Computational Linguistics",
    url = "https://aclanthology.org/2024.emnlp-main.939/",
    doi = "10.18653/v1/2024.emnlp-main.939",
    pages = "16923--16936"
}

@inproceedings{dugan2023roft,
   title={Real or Fake Text?: Investigating Human Ability to Detect Boundaries\\Between Human-Written and Machine-Generated Text},
   author={Dugan, Liam and Ippolito, Daphne and Kirubarajan, Arun and Shi, Sherry and Callison-Burch, Chris},
   booktitle={Proceedings of the 2023 AAAI Conference on Artificial Intelligence},
   year={2023}
 }

@inproceedings{artemova-etal-2025-beemo,
    title = "Beemo: Benchmark of Expert-edited Machine-generated Outputs",
    author = "Artemova, Ekaterina  and
      Lucas, Jason S  and
      Venkatraman, Saranya  and
      Lee, Jooyoung  and
      Tilga, Sergei  and
      Uchendu, Adaku  and
      Mikhailov, Vladislav",
    editor = "Chiruzzo, Luis  and
      Ritter, Alan  and
      Wang, Lu",
    booktitle = "Proceedings of the 2025 Conference of the Nations of the Americas Chapter of the Association for Computational Linguistics: Human Language Technologies (Volume 1: Long Papers)",
    month = apr,
    year = "2025",
    address = "Albuquerque, New Mexico",
    publisher = "Association for Computational Linguistics",
    url = "https://aclanthology.org/2025.naacl-long.357/",
    pages = "6992--7018",
    ISBN = "979-8-89176-189-6"
}

@inproceedings{zhang2024llmasacoauthor,
    title = "{LLM}-as-a-Coauthor: Can Mixed Human-Written and Machine-Generated Text Be Detected?",
    author = "Zhang, Qihui  and
      Gao, Chujie  and
      Chen, Dongping  and
      Huang, Yue  and
      Huang, Yixin  and
      Sun, Zhenyang  and
      Zhang, Shilin  and
      Li, Weiye  and
      Fu, Zhengyan  and
      Wan, Yao  and
      Sun, Lichao",
    editor = "Duh, Kevin  and
      Gomez, Helena  and
      Bethard, Steven",
    booktitle = "Findings of the Association for Computational Linguistics: NAACL 2024",
    month = jun,
    year = "2024",
    address = "Mexico City, Mexico",
    publisher = "Association for Computational Linguistics",
    url = "https://aclanthology.org/2024.findings-naacl.29/",
    doi = "10.18653/v1/2024.findings-naacl.29",
    pages = "409--436"
}

@inproceedings{chakrabarty-etal-2022-help,
    title = "\textit{Help me write a poem}: Instruction Tuning as a Vehicle for Collaborative Poetry Writing",
    author = "Chakrabarty, Tuhin  and
      Padmakumar, Vishakh  and
      He, He",
    editor = "Goldberg, Yoav  and
      Kozareva, Zornitsa  and
      Zhang, Yue",
    booktitle = "Proceedings of the 2022 Conference on Empirical Methods in Natural Language Processing",
    month = dec,
    year = "2022",
    address = "Abu Dhabi, United Arab Emirates",
    publisher = "Association for Computational Linguistics",
    url = "https://aclanthology.org/2022.emnlp-main.460/",
    doi = "10.18653/v1/2022.emnlp-main.460",
    pages = "6848--6863"
}

@inproceedings{akoury-etal-2020-storium,
    title = "{STORIUM}: {A} {D}ataset and {E}valuation {P}latform for {M}achine-in-the-{L}oop {S}tory {G}eneration",
    author = "Akoury, Nader  and
      Wang, Shufan  and
      Whiting, Josh  and
      Hood, Stephen  and
      Peng, Nanyun  and
      Iyyer, Mohit",
    editor = "Webber, Bonnie  and
      Cohn, Trevor  and
      He, Yulan  and
      Liu, Yang",
    booktitle = "Proceedings of the 2020 Conference on Empirical Methods in Natural Language Processing (EMNLP)",
    month = nov,
    year = "2020",
    address = "Online",
    publisher = "Association for Computational Linguistics",
    url = "https://aclanthology.org/2020.emnlp-main.525/",
    doi = "10.18653/v1/2020.emnlp-main.525",
    pages = "6470--6484"
}

@inproceedings{chakrabarty2024creativitysupportintheage,
author = {Chakrabarty, Tuhin and Padmakumar, Vishakh and Brahman, Faeze and Muresan, Smaranda},
title = {Creativity Support in the Age of Large Language Models: An Empirical Study Involving Professional Writers},
year = {2024},
isbn = {9798400704857},
publisher = {Association for Computing Machinery},
address = {New York, NY, USA},
url = {https://doi.org/10.1145/3635636.3656201},
doi = {10.1145/3635636.3656201},
booktitle = {Proceedings of the 16th Conference on Creativity \& Cognition},
pages = {132–155},
numpages = {24},
location = {Chicago, IL, USA},
series = {C\&C '24}
}

@misc{ippolito2022creativewritingaipoweredwriting,
      title={Creative Writing with an AI-Powered Writing Assistant: Perspectives from Professional Writers}, 
      author={Daphne Ippolito and Ann Yuan and Andy Coenen and Sehmon Burnam},
      year={2022},
      eprint={2211.05030},
      archivePrefix={arXiv},
      primaryClass={cs.HC},
      url={https://arxiv.org/abs/2211.05030}, 
}

@inproceedings{
huot2025agents,
title={Agents' Room:  Narrative Generation through Multi-step Collaboration},
author={Fantine Huot and Reinald Kim Amplayo and Jennimaria Palomaki and Alice Shoshana Jakobovits and Elizabeth Clark and Mirella Lapata},
booktitle={The Thirteenth International Conference on Learning Representations},
year={2025},
url={https://openreview.net/forum?id=HfWcFs7XLR}
}

@inproceedings{yuan2022wordcraft,
author = {Yuan, Ann and Coenen, Andy and Reif, Emily and Ippolito, Daphne},
title = {Wordcraft: Story Writing With Large Language Models},
year = {2022},
isbn = {9781450391443},
publisher = {Association for Computing Machinery},
address = {New York, NY, USA},
url = {https://doi.org/10.1145/3490099.3511105},
doi = {10.1145/3490099.3511105},
booktitle = {Proceedings of the 27th International Conference on Intelligent User Interfaces},
pages = {841–852},
numpages = {12},
location = {Helsinki, Finland},
series = {IUI '22}
}

@inproceedings{
nicks2024language,
title={Language Model Detectors Are Easily Optimized Against},
author={Charlotte Nicks and Eric Mitchell and Rafael Rafailov and Archit Sharma and Christopher D Manning and Chelsea Finn and Stefano Ermon},
booktitle={The Twelfth International Conference on Learning Representations},
year={2024},
url={https://openreview.net/forum?id=4eJDMjYZZG}
}

@misc{david2025authormistevadingaitext,
      title={AuthorMist: Evading AI Text Detectors with Reinforcement Learning}, 
      author={Isaac David and Arthur Gervais},
      year={2025},
      eprint={2503.08716},
      archivePrefix={arXiv},
      primaryClass={cs.CR},
      url={https://arxiv.org/abs/2503.08716}, 
}

@inproceedings{ippolito-etal-2020-automatic,
    title = "Automatic Detection of Generated Text is Easiest when Humans are Fooled",
    author = "Ippolito, Daphne  and
      Duckworth, Daniel  and
      Callison-Burch, Chris  and
      Eck, Douglas",
    editor = "Jurafsky, Dan  and
      Chai, Joyce  and
      Schluter, Natalie  and
      Tetreault, Joel",
    booktitle = "Proceedings of the 58th Annual Meeting of the Association for Computational Linguistics",
    month = jul,
    year = "2020",
    address = "Online",
    publisher = "Association for Computational Linguistics",
    url = "https://aclanthology.org/2020.acl-main.164/",
    doi = "10.18653/v1/2020.acl-main.164",
    pages = "1808--1822"
}

@inproceedings{clark-etal-2021-thats,
    title = "All That`s {\textquoteleft}Human' Is Not Gold: Evaluating Human Evaluation of Generated Text",
    author = "Clark, Elizabeth  and
      August, Tal  and
      Serrano, Sofia  and
      Haduong, Nikita  and
      Gururangan, Suchin  and
      Smith, Noah A.",
    editor = "Zong, Chengqing  and
      Xia, Fei  and
      Li, Wenjie  and
      Navigli, Roberto",
    booktitle = "Proceedings of the 59th Annual Meeting of the Association for Computational Linguistics and the 11th International Joint Conference on Natural Language Processing (Volume 1: Long Papers)",
    month = aug,
    year = "2021",
    address = "Online",
    publisher = "Association for Computational Linguistics",
    url = "https://aclanthology.org/2021.acl-long.565/",
    doi = "10.18653/v1/2021.acl-long.565",
    pages = "7282--7296"
}

@inproceedings{dugan-etal-2024-raid,
    title = "{RAID}: A Shared Benchmark for Robust Evaluation of Machine-Generated Text Detectors",
    author = "Dugan, Liam  and
      Hwang, Alyssa  and
      Trhl{\'i}k, Filip  and
      Zhu, Andrew  and
      Ludan, Josh Magnus  and
      Xu, Hainiu  and
      Ippolito, Daphne  and
      Callison-Burch, Chris",
    editor = "Ku, Lun-Wei  and
      Martins, Andre  and
      Srikumar, Vivek",
    booktitle = "Proceedings of the 62nd Annual Meeting of the Association for Computational Linguistics (Volume 1: Long Papers)",
    month = aug,
    year = "2024",
    address = "Bangkok, Thailand",
    publisher = "Association for Computational Linguistics",
    url = "https://aclanthology.org/2024.acl-long.674/",
    doi = "10.18653/v1/2024.acl-long.674",
    pages = "12463--12492"
}

@inproceedings{mitchell2023detectgpt,
author = {Mitchell, Eric and Lee, Yoonho and Khazatsky, Alexander and Manning, Christopher D. and Finn, Chelsea},
title = {DetectGPT: zero-shot machine-generated text detection using probability curvature},
year = {2023},
publisher = {JMLR.org},
booktitle = {Proceedings of the 40th International Conference on Machine Learning},
articleno = {1038},
numpages = {13},
location = {Honolulu, Hawaii, USA},
series = {ICML'23}
}

@inproceedings{verma-etal-2024-ghostbuster,
    title = "Ghostbuster: Detecting Text Ghostwritten by Large Language Models",
    author = "Verma, Vivek  and
      Fleisig, Eve  and
      Tomlin, Nicholas  and
      Klein, Dan",
    editor = "Duh, Kevin  and
      Gomez, Helena  and
      Bethard, Steven",
    booktitle = "Proceedings of the 2024 Conference of the North American Chapter of the Association for Computational Linguistics: Human Language Technologies (Volume 1: Long Papers)",
    month = jun,
    year = "2024",
    address = "Mexico City, Mexico",
    publisher = "Association for Computational Linguistics",
    url = "https://aclanthology.org/2024.naacl-long.95/",
    doi = "10.18653/v1/2024.naacl-long.95",
    pages = "1702--1717"
}

@inproceedings{dugan2023realorfaketext,
author = {Dugan, Liam and Ippolito, Daphne and Kirubarajan, Arun and Shi, Sherry and Callison-Burch, Chris},
title = {Real or fake text? investigating human ability to detect boundaries between human-written and machine-generated text},
year = {2023},
isbn = {978-1-57735-880-0},
publisher = {AAAI Press},
url = {https://doi.org/10.1609/aaai.v37i11.26501},
doi = {10.1609/aaai.v37i11.26501},
booktitle = {Proceedings of the Thirty-Seventh AAAI Conference on Artificial Intelligence and Thirty-Fifth Conference on Innovative Applications of Artificial Intelligence and Thirteenth Symposium on Educational Advances in Artificial Intelligence},
articleno = {1432},
numpages = {9},
series = {AAAI'23/IAAI'23/EAAI'23}
}

@inproceedings{
kushnareva2024boundary,
title={Boundary detection in mixed {AI}-human texts},
author={Laida Kushnareva and Tatiana Gaintseva and Dmitry Abulkhanov and Kristian Kuznetsov and German Magai and Eduard Tulchinskii and Serguei Barannikov and Sergey Nikolenko and Irina Piontkovskaya},
booktitle={First Conference on Language Modeling},
year={2024},
url={https://openreview.net/forum?id=kzzwTrt04Z}
}

@inproceedings{abassy2024llmdetectaive,
    title = "{LLM}-{D}etect{AI}ve: a Tool for Fine-Grained Machine-Generated Text Detection",
    author = "Abassy, Mervat  and
      Elozeiri, Kareem  and
      Aziz, Alexander  and
      Ta, Minh Ngoc  and
      Tomar, Raj Vardhan  and
      Adhikari, Bimarsha  and
      Ahmed, Saad El Dine  and
      Wang, Yuxia  and
      Mohammed Afzal, Osama  and
      Xie, Zhuohan  and
      Mansurov, Jonibek  and
      Artemova, Ekaterina  and
      Mikhailov, Vladislav  and
      Xing, Rui  and
      Geng, Jiahui  and
      Iqbal, Hasan  and
      Mujahid, Zain Muhammad  and
      Mahmoud, Tarek  and
      Tsvigun, Akim  and
      Aji, Alham Fikri  and
      Shelmanov, Artem  and
      Habash, Nizar  and
      Gurevych, Iryna  and
      Nakov, Preslav",
    editor = "Hernandez Farias, Delia Irazu  and
      Hope, Tom  and
      Li, Manling",
    booktitle = "Proceedings of the 2024 Conference on Empirical Methods in Natural Language Processing: System Demonstrations",
    month = nov,
    year = "2024",
    address = "Miami, Florida, USA",
    publisher = "Association for Computational Linguistics",
    url = "https://aclanthology.org/2024.emnlp-demo.35/",
    doi = "10.18653/v1/2024.emnlp-demo.35",
    pages = "336--343"
}

@inproceedings{richburg2024automaticauthorshipanalysis,
    title = "Automatic Authorship Analysis in Human-{AI} Collaborative Writing",
    author = "Richburg, Aquia  and
      Bao, Calvin  and
      Carpuat, Marine",
    editor = "Calzolari, Nicoletta  and
      Kan, Min-Yen  and
      Hoste, Veronique  and
      Lenci, Alessandro  and
      Sakti, Sakriani  and
      Xue, Nianwen",
    booktitle = "Proceedings of the 2024 Joint International Conference on Computational Linguistics, Language Resources and Evaluation (LREC-COLING 2024)",
    month = may,
    year = "2024",
    address = "Torino, Italia",
    publisher = "ELRA and ICCL",
    url = "https://aclanthology.org/2024.lrec-main.165/",
    pages = "1845--1855"
}

@inproceedings{koike2024outfox,
author = {Koike, Ryuto and Kaneko, Masahiro and Okazaki, Naoaki},
title = {OUTFOX: LLM-generated essay detection through in-context learning with adversarially generated examples},
year = {2024},
isbn = {978-1-57735-887-9},
publisher = {AAAI Press},
url = {https://doi.org/10.1609/aaai.v38i19.30120},
doi = {10.1609/aaai.v38i19.30120},
booktitle = {Proceedings of the Thirty-Eighth AAAI Conference on Artificial Intelligence and Thirty-Sixth Conference on Innovative Applications of Artificial Intelligence and Fourteenth Symposium on Educational Advances in Artificial Intelligence},
articleno = {2372},
numpages = {9},
series = {AAAI'24/IAAI'24/EAAI'24}
}

@inproceedings{zeng2024detectingaigeneratedsentences,
author = {Zeng, Zijie and Liu, Shiqi and Sha, Lele and Li, Zhuang and Yang, Kaixun and Liu, Sannyuya and Ga\v{s}evi\'{c}, Dragan and Chen, Guanliang},
title = {Detecting AI-generated sentences in human-AI collaborative hybrid texts: challenges, strategies, and insights},
year = {2024},
isbn = {978-1-956792-04-1},
url = {https://doi.org/10.24963/ijcai.2024/835},
doi = {10.24963/ijcai.2024/835},
booktitle = {Proceedings of the Thirty-Third International Joint Conference on Artificial Intelligence},
articleno = {835},
numpages = {9},
location = {Jeju, Korea},
series = {IJCAI '24}
}

@inproceedings{saha2025aihumanchallengedetecting,
    title = "Almost {AI}, Almost Human: The Challenge of Detecting {AI}-Polished Writing",
    author = "Saha, Shoumik  and
      Feizi, Soheil",
    editor = "Che, Wanxiang  and
      Nabende, Joyce  and
      Shutova, Ekaterina  and
      Pilehvar, Mohammad Taher",
    booktitle = "Findings of the Association for Computational Linguistics: ACL 2025",
    month = jul,
    year = "2025",
    address = "Vienna, Austria",
    publisher = "Association for Computational Linguistics",
    url = "https://aclanthology.org/2025.findings-acl.1303/",
    doi = "10.18653/v1/2025.findings-acl.1303",
    pages = "25414--25431",
    ISBN = "979-8-89176-256-5"
}

@misc{yeh2025ghostwriter,
      title={GhostWriter: Augmenting Collaborative Human-AI Writing Experiences Through Personalization and Agency}, 
      author={Catherine Yeh and Gonzalo Ramos and Rachel Ng and Andy Huntington and Richard Banks},
      year={2025},
      eprint={2402.08855},
      archivePrefix={arXiv},
      primaryClass={cs.HC},
      url={https://arxiv.org/abs/2402.08855}, 
}

@inproceedings{he2025whichcontributions,
author = {He, Jessica and Houde, Stephanie and Weisz, Justin D.},
title = {Which Contributions Deserve Credit? Perceptions of Attribution in Human-AI Co-Creation},
year = {2025},
isbn = {9798400713941},
publisher = {Association for Computing Machinery},
address = {New York, NY, USA},
url = {https://doi.org/10.1145/3706598.3713522},
doi = {10.1145/3706598.3713522},
booktitle = {Proceedings of the 2025 CHI Conference on Human Factors in Computing Systems},
articleno = {540},
numpages = {18},
location = {
},
series = {CHI '25}
}

@inproceedings{NEURIPS2023_1b44b878,
 author = {Shinn, Noah and Cassano, Federico and Gopinath, Ashwin and Narasimhan, Karthik and Yao, Shunyu},
 booktitle = {Advances in Neural Information Processing Systems},
 editor = {A. Oh and T. Naumann and A. Globerson and K. Saenko and M. Hardt and S. Levine},
 pages = {8634--8652},
 publisher = {Curran Associates, Inc.},
 title = {Reflexion: language agents with verbal reinforcement learning},
 url = {https://proceedings.neurips.cc/paper_files/paper/2023/file/1b44b878bb782e6954cd888628510e90-Paper-Conference.pdf},
 volume = {36},
 year = {2023}
}

@inproceedings{NEURIPS2023_91edff07,
 author = {Madaan, Aman and Tandon, Niket and Gupta, Prakhar and Hallinan, Skyler and Gao, Luyu and Wiegreffe, Sarah and Alon, Uri and Dziri, Nouha and Prabhumoye, Shrimai and Yang, Yiming and Gupta, Shashank and Majumder, Bodhisattwa Prasad and Hermann, Katherine and Welleck, Sean and Yazdanbakhsh, Amir and Clark, Peter},
 booktitle = {Advances in Neural Information Processing Systems},
 editor = {A. Oh and T. Naumann and A. Globerson and K. Saenko and M. Hardt and S. Levine},
 pages = {46534--46594},
 publisher = {Curran Associates, Inc.},
 title = {Self-Refine: Iterative Refinement with Self-Feedback},
 url = {https://proceedings.neurips.cc/paper_files/paper/2023/file/91edff07232fb1b55a505a9e9f6c0ff3-Paper-Conference.pdf},
 volume = {36},
 year = {2023}
}

@misc{kumar2025storyitpersonalizingstory,
      title={Whose story is it? Personalizing story generation by inferring author styles}, 
      author={Nischal Ashok Kumar and Chau Minh Pham and Mohit Iyyer and Andrew Lan},
      year={2025},
      eprint={2502.13028},
      archivePrefix={arXiv},
      primaryClass={cs.CL},
      url={https://arxiv.org/abs/2502.13028}, 
}

@inproceedings{chiang2024chatbotarenaopenplatform,
author = {Chiang, Wei-Lin and Zheng, Lianmin and Sheng, Ying and Angelopoulos, Anastasios N. and Li, Tianle and Li, Dacheng and Zhu, Banghua and Zhang, Hao and Jordan, Michael I. and Gonzalez, Joseph E. and Stoica, Ion},
title = {Chatbot arena: an open platform for evaluating LLMs by human preference},
year = {2024},
publisher = {JMLR.org},
booktitle = {Proceedings of the 41st International Conference on Machine Learning},
articleno = {331},
numpages = {30},
location = {Vienna, Austria},
series = {ICML'24}
}

@misc{wang2024llmdetectorimprovingaigeneratedchinese,
      title={LLM-Detector: Improving AI-Generated Chinese Text Detection with Open-Source LLM Instruction Tuning}, 
      author={Rongsheng Wang and Haoming Chen and Ruizhe Zhou and Han Ma and Yaofei Duan and Yanlan Kang and Songhua Yang and Baoyu Fan and Tao Tan},
      year={2024},
      eprint={2402.01158},
      archivePrefix={arXiv},
      primaryClass={cs.CL},
      url={https://arxiv.org/abs/2402.01158}, 
}

@misc{wang2025humanliketextlikedhumans,
      title={Is Human-Like Text Liked by Humans? Multilingual Human Detection and Preference Against AI}, 
      author={Yuxia Wang and Rui Xing and Jonibek Mansurov and Giovanni Puccetti and Zhuohan Xie and Minh Ngoc Ta and Jiahui Geng and Jinyan Su and Mervat Abassy and Saad El Dine Ahmed and Kareem Elozeiri and Nurkhan Laiyk and Maiya Goloburda and Tarek Mahmoud and Raj Vardhan Tomar and Alexander Aziz and Ryuto Koike and Masahiro Kaneko and Artem Shelmanov and Ekaterina Artemova and Vladislav Mikhailov and Akim Tsvigun and Alham Fikri Aji and Nizar Habash and Iryna Gurevych and Preslav Nakov},
      year={2025},
      eprint={2502.11614},
      archivePrefix={arXiv},
      primaryClass={cs.CL},
      url={https://arxiv.org/abs/2502.11614}, 
}

@inproceedings{booookscore,
  author       = {Yapei Chang and
                  Kyle Lo and
                  Tanya Goyal and
                  Mohit Iyyer},
  title        = {BooookScore: {A} systematic exploration of book-length summarization
                  in the era of LLMs},
  booktitle    = {The Twelfth International Conference on Learning Representations,
                  {ICLR} 2024, Vienna, Austria, May 7-11, 2024},
  publisher    = {OpenReview.net},
  year         = {2024},
  url          = {https://openreview.net/forum?id=7Ttk3RzDeu},
  bibsource    = {dblp computer science bibliography, https://dblp.org}
}

@misc{atmakuru2024cs4measuringcreativitylarge,
      title={CS4: Measuring the Creativity of Large Language Models Automatically by Controlling the Number of Story-Writing Constraints}, 
      author={Anirudh Atmakuru and Jatin Nainani and Rohith Siddhartha Reddy Bheemreddy and Anirudh Lakkaraju and Zonghai Yao and Hamed Zamani and Haw-Shiuan Chang},
      year={2024},
      eprint={2410.04197},
      archivePrefix={arXiv},
      primaryClass={cs.CL},
      url={https://arxiv.org/abs/2410.04197}, 
}

@inproceedings{yang-etal-2022-re3,
    title = "Re3: Generating Longer Stories With Recursive Reprompting and Revision",
    author = "Yang, Kevin  and
      Tian, Yuandong  and
      Peng, Nanyun  and
      Klein, Dan",
    editor = "Goldberg, Yoav  and
      Kozareva, Zornitsa  and
      Zhang, Yue",
    booktitle = "Proceedings of the 2022 Conference on Empirical Methods in Natural Language Processing",
    month = dec,
    year = "2022",
    address = "Abu Dhabi, United Arab Emirates",
    publisher = "Association for Computational Linguistics",
    url = "https://aclanthology.org/2022.emnlp-main.296/",
    doi = "10.18653/v1/2022.emnlp-main.296",
    pages = "4393--4479"
}

@article{johnson2019billion,
  title={Billion-scale similarity search with {GPUs}},
  author={Johnson, Jeff and Douze, Matthijs and J{\'e}gou, Herv{\'e}},
  journal={IEEE Transactions on Big Data},
  volume={7},
  number={3},
  pages={535--547},
  year={2019},
  publisher={IEEE}
}

@misc{deepseekai2025deepseekr1incentivizingreasoningcapability,
      title={DeepSeek-R1: Incentivizing Reasoning Capability in LLMs via Reinforcement Learning}, 
      author={DeepSeek-AI and Daya Guo and Dejian Yang and Haowei Zhang and Junxiao Song and Ruoyu Zhang and Runxin Xu and Qihao Zhu and Shirong Ma and Peiyi Wang and Xiao Bi and Xiaokang Zhang and Xingkai Yu and Yu Wu and Z. F. Wu and Zhibin Gou and Zhihong Shao and Zhuoshu Li and Ziyi Gao and Aixin Liu and Bing Xue and Bingxuan Wang and Bochao Wu and Bei Feng and Chengda Lu and Chenggang Zhao and Chengqi Deng and Chenyu Zhang and Chong Ruan and Damai Dai and Deli Chen and Dongjie Ji and Erhang Li and Fangyun Lin and Fucong Dai and Fuli Luo and Guangbo Hao and Guanting Chen and Guowei Li and H. Zhang and Han Bao and Hanwei Xu and Haocheng Wang and Honghui Ding and Huajian Xin and Huazuo Gao and Hui Qu and Hui Li and Jianzhong Guo and Jiashi Li and Jiawei Wang and Jingchang Chen and Jingyang Yuan and Junjie Qiu and Junlong Li and J. L. Cai and Jiaqi Ni and Jian Liang and Jin Chen and Kai Dong and Kai Hu and Kaige Gao and Kang Guan and Kexin Huang and Kuai Yu and Lean Wang and Lecong Zhang and Liang Zhao and Litong Wang and Liyue Zhang and Lei Xu and Leyi Xia and Mingchuan Zhang and Minghua Zhang and Minghui Tang and Meng Li and Miaojun Wang and Mingming Li and Ning Tian and Panpan Huang and Peng Zhang and Qiancheng Wang and Qinyu Chen and Qiushi Du and Ruiqi Ge and Ruisong Zhang and Ruizhe Pan and Runji Wang and R. J. Chen and R. L. Jin and Ruyi Chen and Shanghao Lu and Shangyan Zhou and Shanhuang Chen and Shengfeng Ye and Shiyu Wang and Shuiping Yu and Shunfeng Zhou and Shuting Pan and S. S. Li and Shuang Zhou and Shaoqing Wu and Shengfeng Ye and Tao Yun and Tian Pei and Tianyu Sun and T. Wang and Wangding Zeng and Wanjia Zhao and Wen Liu and Wenfeng Liang and Wenjun Gao and Wenqin Yu and Wentao Zhang and W. L. Xiao and Wei An and Xiaodong Liu and Xiaohan Wang and Xiaokang Chen and Xiaotao Nie and Xin Cheng and Xin Liu and Xin Xie and Xingchao Liu and Xinyu Yang and Xinyuan Li and Xuecheng Su and Xuheng Lin and X. Q. Li and Xiangyue Jin and Xiaojin Shen and Xiaosha Chen and Xiaowen Sun and Xiaoxiang Wang and Xinnan Song and Xinyi Zhou and Xianzu Wang and Xinxia Shan and Y. K. Li and Y. Q. Wang and Y. X. Wei and Yang Zhang and Yanhong Xu and Yao Li and Yao Zhao and Yaofeng Sun and Yaohui Wang and Yi Yu and Yichao Zhang and Yifan Shi and Yiliang Xiong and Ying He and Yishi Piao and Yisong Wang and Yixuan Tan and Yiyang Ma and Yiyuan Liu and Yongqiang Guo and Yuan Ou and Yuduan Wang and Yue Gong and Yuheng Zou and Yujia He and Yunfan Xiong and Yuxiang Luo and Yuxiang You and Yuxuan Liu and Yuyang Zhou and Y. X. Zhu and Yanhong Xu and Yanping Huang and Yaohui Li and Yi Zheng and Yuchen Zhu and Yunxian Ma and Ying Tang and Yukun Zha and Yuting Yan and Z. Z. Ren and Zehui Ren and Zhangli Sha and Zhe Fu and Zhean Xu and Zhenda Xie and Zhengyan Zhang and Zhewen Hao and Zhicheng Ma and Zhigang Yan and Zhiyu Wu and Zihui Gu and Zijia Zhu and Zijun Liu and Zilin Li and Ziwei Xie and Ziyang Song and Zizheng Pan and Zhen Huang and Zhipeng Xu and Zhongyu Zhang and Zhen Zhang},
      year={2025},
      eprint={2501.12948},
      archivePrefix={arXiv},
      primaryClass={cs.CL},
      url={https://arxiv.org/abs/2501.12948}, 
}

@misc{gpt5,
  author       = {OpenAI},
  title        = {OpenAI GPT-5 System Card},
  year         = {2025},
  howpublished = {\url{https://cdn.openai.com/gpt-5-system-card.pdf}},
}

@misc{qwen3,
    title  = {Qwen3},
    url    = {https://qwenlm.github.io/blog/qwen3/},
    author = {QwenTeam},
    month  = {April},
    year   = {2025}
}

@misc{claude4sonnet2025,
  author       = {Anthropic},
  title        = {System Card: Claude Opus 4 \& Claude Sonnet 4},
  year         = {2025},
  howpublished = {\url{https://www-cdn.anthropic.com/6d8a8055020700718b0c49369f60816ba2a7c285.pdf}}
}

@InProceedings{kirchenbauer2023watermark,
  title = 	 {A Watermark for Large Language Models},
  author =       {Kirchenbauer, John and Geiping, Jonas and Wen, Yuxin and Katz, Jonathan and Miers, Ian and Goldstein, Tom},
  booktitle = 	 {Proceedings of the 40th International Conference on Machine Learning},
  pages = 	 {17061--17084},
  year = 	 {2023},
  editor = 	 {Krause, Andreas and Brunskill, Emma and Cho, Kyunghyun and Engelhardt, Barbara and Sabato, Sivan and Scarlett, Jonathan},
  volume = 	 {202},
  series = 	 {Proceedings of Machine Learning Research},
  month = 	 {23--29 Jul},
  publisher =    {PMLR},
  url = 	 {https://proceedings.mlr.press/v202/kirchenbauer23a.html}
}

@inproceedings{chang2024postmark,
    title = "{P}ost{M}ark: A Robust Blackbox Watermark for Large Language Models",
    author = "Chang, Yapei  and
      Krishna, Kalpesh  and
      Houmansadr, Amir  and
      Wieting, John Frederick  and
      Iyyer, Mohit",
    editor = "Al-Onaizan, Yaser  and
      Bansal, Mohit  and
      Chen, Yun-Nung",
    booktitle = "Proceedings of the 2024 Conference on Empirical Methods in Natural Language Processing",
    month = nov,
    year = "2024",
    address = "Miami, Florida, USA",
    publisher = "Association for Computational Linguistics",
    url = "https://aclanthology.org/2024.emnlp-main.506/",
    doi = "10.18653/v1/2024.emnlp-main.506",
    pages = "8969--8987"
}

@book{shelley1818frankenstein,
  author    = {Mary Shelley},
  title     = {Frankenstein; or, The Modern Prometheus},
  year      = {1818},
  publisher = {Lackington, Hughes, Harding, Mavor \& Jones},
  address   = {London},
  note      = {Original edition}
}

@inproceedings{reimers2019sbert,
    title = "Sentence-{BERT}: Sentence Embeddings using {S}iamese {BERT}-Networks",
    author = "Reimers, Nils  and
      Gurevych, Iryna",
    editor = "Inui, Kentaro  and
      Jiang, Jing  and
      Ng, Vincent  and
      Wan, Xiaojun",
    booktitle = "Proceedings of the 2019 Conference on Empirical Methods in Natural Language Processing and the 9th International Joint Conference on Natural Language Processing (EMNLP-IJCNLP)",
    month = nov,
    year = "2019",
    address = "Hong Kong, China",
    publisher = "Association for Computational Linguistics",
    url = "https://aclanthology.org/D19-1410/",
    doi = "10.18653/v1/D19-1410",
    pages = "3982--3992"
}

@inproceedings{muennighoff2023mteb,
    title = "{MTEB}: Massive Text Embedding Benchmark",
    author = "Muennighoff, Niklas  and
      Tazi, Nouamane  and
      Magne, Loic  and
      Reimers, Nils",
    editor = "Vlachos, Andreas  and
      Augenstein, Isabelle",
    booktitle = "Proceedings of the 17th Conference of the European Chapter of the Association for Computational Linguistics",
    month = may,
    year = "2023",
    address = "Dubrovnik, Croatia",
    publisher = "Association for Computational Linguistics",
    url = "https://aclanthology.org/2023.eacl-main.148/",
    doi = "10.18653/v1/2023.eacl-main.148",
    pages = "2014--2037"
}

@inproceedings{yao2019plan,
  title={Plan-and-write: Towards better automatic storytelling},
  author={Yao, Lili and Peng, Nanyun and Weischedel, Ralph and Knight, Kevin and Zhao, Dongyan and Yan, Rui},
  booktitle={Proceedings of the AAAI Conference on Artificial Intelligence},
  volume={33},
  pages={7378--7385},
  year={2019}
}

@inproceedings{yang-etal-2023-doc,
    title = "{DOC}: Improving Long Story Coherence With Detailed Outline Control",
    author = "Yang, Kevin  and
      Klein, Dan  and
      Peng, Nanyun  and
      Tian, Yuandong",
    editor = "Rogers, Anna  and
      Boyd-Graber, Jordan  and
      Okazaki, Naoaki",
    booktitle = "Proceedings of the 61st Annual Meeting of the Association for Computational Linguistics (Volume 1: Long Papers)",
    month = jul,
    year = "2023",
    address = "Toronto, Canada",
    publisher = "Association for Computational Linguistics",
    url = "https://aclanthology.org/2023.acl-long.190/",
    doi = "10.18653/v1/2023.acl-long.190",
    pages = "3378--3465"
}

@inproceedings{fan-etal-2019-strategies,
    title = "Strategies for Structuring Story Generation",
    author = "Fan, Angela  and
      Lewis, Mike  and
      Dauphin, Yann",
    editor = "Korhonen, Anna  and
      Traum, David  and
      M{\`a}rquez, Llu{\'i}s",
    booktitle = "Proceedings of the 57th Annual Meeting of the Association for Computational Linguistics",
    month = jul,
    year = "2019",
    address = "Florence, Italy",
    publisher = "Association for Computational Linguistics",
    url = "https://aclanthology.org/P19-1254/",
    doi = "10.18653/v1/P19-1254",
    pages = "2650--2660"
}

@inproceedings{papalampidi2022towards,
  title={Towards coherent and consistent use of entities in narrative generation},
  author={Papalampidi, Pinelopi and Cao, Kris and Kocisky, Tomas},
  booktitle={International Conference on Machine Learning},
  pages={17278--17294},
  year={2022},
  organization={PMLR}
}

@inproceedings{rashkin2020plotmachines,
    title = "{P}lot{M}achines: Outline-Conditioned Generation with Dynamic Plot State Tracking",
    author = "Rashkin, Hannah  and
      Celikyilmaz, Asli  and
      Choi, Yejin  and
      Gao, Jianfeng",
    editor = "Webber, Bonnie  and
      Cohn, Trevor  and
      He, Yulan  and
      Liu, Yang",
    booktitle = "Proceedings of the 2020 Conference on Empirical Methods in Natural Language Processing (EMNLP)",
    month = nov,
    year = "2020",
    address = "Online",
    publisher = "Association for Computational Linguistics",
    url = "https://aclanthology.org/2020.emnlp-main.349/",
    doi = "10.18653/v1/2020.emnlp-main.349",
    pages = "4274--4295"
}

@inproceedings{kong-etal-2021-stylized,
    title = "Stylized Story Generation with Style-Guided Planning",
    author = "Kong, Xiangzhe  and
      Huang, Jialiang  and
      Tung, Ziquan  and
      Guan, Jian  and
      Huang, Minlie",
    editor = "Zong, Chengqing  and
      Xia, Fei  and
      Li, Wenjie  and
      Navigli, Roberto",
    booktitle = "Findings of the Association for Computational Linguistics: ACL-IJCNLP 2021",
    month = aug,
    year = "2021",
    address = "Online",
    publisher = "Association for Computational Linguistics",
    url = "https://aclanthology.org/2021.findings-acl.215/",
    doi = "10.18653/v1/2021.findings-acl.215",
    pages = "2430--2436"
}

@inproceedings{sun-etal-2021-iga,
    title = "{IGA}: An Intent-Guided Authoring Assistant",
    author = "Sun, Simeng  and
      Zhao, Wenlong  and
      Manjunatha, Varun  and
      Jain, Rajiv  and
      Morariu, Vlad  and
      Dernoncourt, Franck  and
      Srinivasan, Balaji Vasan  and
      Iyyer, Mohit",
    editor = "Moens, Marie-Francine  and
      Huang, Xuanjing  and
      Specia, Lucia  and
      Yih, Scott Wen-tau",
    booktitle = "Proceedings of the 2021 Conference on Empirical Methods in Natural Language Processing",
    month = nov,
    year = "2021",
    address = "Online and Punta Cana, Dominican Republic",
    publisher = "Association for Computational Linguistics",
    url = "https://aclanthology.org/2021.emnlp-main.483/",
    doi = "10.18653/v1/2021.emnlp-main.483",
    pages = "5972--5985"
}

@inproceedings{xie-etal-2023-next,
    title = "The Next Chapter: A Study of Large Language Models in Storytelling",
    author = "Xie, Zhuohan  and
      Cohn, Trevor  and
      Lau, Jey Han",
    editor = "Keet, C. Maria  and
      Lee, Hung-Yi  and
      Zarrie{\ss}, Sina",
    booktitle = "Proceedings of the 16th International Natural Language Generation Conference",
    month = sep,
    year = "2023",
    address = "Prague, Czechia",
    publisher = "Association for Computational Linguistics",
    url = "https://aclanthology.org/2023.inlg-main.23/",
    doi = "10.18653/v1/2023.inlg-main.23",
    pages = "323--351"
}

@inproceedings{chiang-lee-2023-large,
    title = "Can Large Language Models Be an Alternative to Human Evaluations?",
    author = "Chiang, Cheng-Han  and
      Lee, Hung-yi",
    editor = "Rogers, Anna  and
      Boyd-Graber, Jordan  and
      Okazaki, Naoaki",
    booktitle = "Proceedings of the 61st Annual Meeting of the Association for Computational Linguistics (Volume 1: Long Papers)",
    month = jul,
    year = "2023",
    address = "Toronto, Canada",
    publisher = "Association for Computational Linguistics",
    url = "https://aclanthology.org/2023.acl-long.870/",
    doi = "10.18653/v1/2023.acl-long.870",
    pages = "15607--15631"
}

@misc{wei2023skyworkopenbilingualfoundation,
      title={Skywork: A More Open Bilingual Foundation Model}, 
      author={Tianwen Wei and Liang Zhao and Lichang Zhang and Bo Zhu and Lijie Wang and Haihua Yang and Biye Li and Cheng Cheng and Weiwei Lü and Rui Hu and Chenxia Li and Liu Yang and Xilin Luo and Xuejie Wu and Lunan Liu and Wenjun Cheng and Peng Cheng and Jianhao Zhang and Xiaoyu Zhang and Lei Lin and Xiaokun Wang and Yutuan Ma and Chuanhai Dong and Yanqi Sun and Yifu Chen and Yongyi Peng and Xiaojuan Liang and Shuicheng Yan and Han Fang and Yahui Zhou},
      year={2023},
      eprint={2310.19341},
      archivePrefix={arXiv},
      primaryClass={cs.CL},
      url={https://arxiv.org/abs/2310.19341}, 
}

@inproceedings{mysore2025prototypicalhumanaicollaborationbehaviors,
    title = "Prototypical Human-{AI} Collaboration Behaviors from {LLM}-Assisted Writing in the Wild",
    author = "Mysore, Sheshera  and
      Das, Debarati  and
      Cao, Hancheng  and
      Sarrafzadeh, Bahareh",
    editor = "Christodoulopoulos, Christos  and
      Chakraborty, Tanmoy  and
      Rose, Carolyn  and
      Peng, Violet",
    booktitle = "Proceedings of the 2025 Conference on Empirical Methods in Natural Language Processing",
    month = nov,
    year = "2025",
    address = "Suzhou, China",
    publisher = "Association for Computational Linguistics",
    url = "https://aclanthology.org/2025.emnlp-main.852/",
    doi = "10.18653/v1/2025.emnlp-main.852",
    pages = "16819--16846",
    ISBN = "979-8-89176-332-6"
}

@inproceedings{
lu2025aihumanityssalieriquantifying,
title={{AI} as Humanity{\textquoteright}s Salieri: Quantifying Linguistic Creativity of Language Models via Systematic Attribution of Machine Text against Web Text},
author={Ximing Lu and Melanie Sclar and Skyler Hallinan and Niloofar Mireshghallah and Jiacheng Liu and Seungju Han and Allyson Ettinger and Liwei Jiang and Khyathi Chandu and Nouha Dziri and Yejin Choi},
booktitle={The Thirteenth International Conference on Learning Representations},
year={2025},
url={https://openreview.net/forum?id=ilOEOIqolQ}
}

@inproceedings{Buschek_2024, series={DIS ’24},
   title={Collage is the New Writing: Exploring the Fragmentation of Text and User Interfaces in AI Tools},
   url={http://dx.doi.org/10.1145/3643834.3660681},
   DOI={10.1145/3643834.3660681},
   booktitle={Designing Interactive Systems Conference},
   publisher={ACM},
   author={Buschek, Daniel},
   year={2024},
   month=jul, pages={2719–2737},
   collection={DIS ’24} }

@inproceedings{
zhang2025noveltybenchevaluatinglanguagemodels,
title={NoveltyBench: Evaluating Creativity and Diversity in Language Models},
author={Yiming Zhang and Harshita Diddee and Susan Holm and Hanchen Liu and Xinyue Liu and Vinay Samuel and Barry Wang and Daphne Ippolito},
booktitle={Second Conference on Language Modeling},
year={2025},
url={https://openreview.net/forum?id=XZm1ekzERf}
}

@misc{ismayilzada2025evaluatingcreativeshortstory,
      title={Evaluating Creative Short Story Generation in Humans and Large Language Models}, 
      author={Mete Ismayilzada and Claire Stevenson and Lonneke van der Plas},
      year={2025},
      eprint={2411.02316},
      archivePrefix={arXiv},
      primaryClass={cs.CL},
      url={https://arxiv.org/abs/2411.02316}, 
}

@INPROCEEDINGS{6901525,
  author={Karampiperis, Pythagoras and Koukourikos, Antonis and Koliopoulou, Evangelia},
  booktitle={2014 IEEE 14th International Conference on Advanced Learning Technologies}, 
  title={Towards Machines for Measuring Creativity: The Use of Computational Tools in Storytelling Activities}, 
  year={2014},
  volume={},
  number={},
  pages={508-512},
  doi={10.1109/ICALT.2014.150}}

@book{boden2004creative,
  title={The creative mind: Myths and mechanisms},
  author={Boden, Margaret A},
  year={2004},
  publisher={Routledge}
}

@inproceedings{grace2014expect,
  title={What to expect when you're expecting: The role of unexpectedness in computationally evaluating creativity.},
  author={Grace, Kazjon and Maher, Mary Lou},
  booktitle={ICCC},
  pages={120--128},
  year={2014},
  organization={Ljubljana}
}

@article{Franceschelli_2024,
   title={Creativity and Machine Learning: A Survey},
   volume={56},
   ISSN={1557-7341},
   url={http://dx.doi.org/10.1145/3664595},
   DOI={10.1145/3664595},
   number={11},
   journal={ACM Computing Surveys},
   publisher={Association for Computing Machinery (ACM)},
   author={Franceschelli, Giorgio and Musolesi, Mirco},
   year={2024},
   month=jun, pages={1–41} }

@article{finstad2010response,
  title={Response interpolation and scale sensitivity: Evidence against 5-point scales},
  author={Finstad, Kraig},
  journal={Journal of usability studies},
  volume={5},
  number={3},
  pages={104--110},
  year={2010},
  publisher={Usability Professionals' Association Bloomingdale, IL}
}

@misc{paech2023eqbench,
	title={EQ-Bench: An Emotional Intelligence Benchmark for Large Language Models}, 
	author={Samuel J. Paech},
	year={2023},
	eprint={2312.06281},
	archivePrefix={arXiv},
	primaryClass={cs.CL}
}

@misc{shaib2025measuringaisloptext,
      title={Measuring AI "Slop" in Text}, 
      author={Chantal Shaib and Tuhin Chakrabarty and Diego Garcia-Olano and Byron C. Wallace},
      year={2025},
      eprint={2509.19163},
      archivePrefix={arXiv},
      primaryClass={cs.CL},
      url={https://arxiv.org/abs/2509.19163}, 
}

@inproceedings{art_or_artifice,
author = {Chakrabarty, Tuhin and Laban, Philippe and Agarwal, Divyansh and Muresan, Smaranda and Wu, Chien-Sheng},
title = {Art or Artifice? Large Language Models and the False Promise of Creativity},
year = {2024},
isbn = {9798400703300},
publisher = {Association for Computing Machinery},
address = {New York, NY, USA},
url = {https://doi.org/10.1145/3613904.3642731},
doi = {10.1145/3613904.3642731},
booktitle = {Proceedings of the 2024 CHI Conference on Human Factors in Computing Systems},
articleno = {30},
numpages = {34},
location = {Honolulu, HI, USA},
series = {CHI '24}
}

@inproceedings{masrour-etal-2025-damage,
    title = "{DAMAGE}: Detecting Adversarially Modified {AI} Generated Text",
    author = "Masrour, Elyas  and
      Emi, Bradley N.  and
      Spero, Max",
    editor = "Alam, Firoj  and
      Nakov, Preslav  and
      Habash, Nizar  and
      Gurevych, Iryna  and
      Chowdhury, Shammur  and
      Shelmanov, Artem  and
      Wang, Yuxia  and
      Artemova, Ekaterina  and
      Kutlu, Mucahid  and
      Mikros, George",
    booktitle = "Proceedings of the 1stWorkshop on GenAI Content Detection (GenAIDetect)",
    month = jan,
    year = "2025",
    address = "Abu Dhabi, UAE",
    publisher = "International Conference on Computational Linguistics",
    url = "https://aclanthology.org/2025.genaidetect-1.9/",
    pages = "120--133"
}

@inproceedings{zeng2024detectingaigeneratedsentenceshumanai,
author = {Zeng, Zijie and Liu, Shiqi and Sha, Lele and Li, Zhuang and Yang, Kaixun and Liu, Sannyuya and Ga\v{s}evi\'{c}, Dragan and Chen, Guanliang},
title = {Detecting AI-generated sentences in human-AI collaborative hybrid texts: challenges, strategies, and insights},
year = {2024},
isbn = {978-1-956792-04-1},
url = {https://doi.org/10.24963/ijcai.2024/835},
doi = {10.24963/ijcai.2024/835},
booktitle = {Proceedings of the Thirty-Third International Joint Conference on Artificial Intelligence},
articleno = {835},
numpages = {9},
location = {Jeju, Korea},
series = {IJCAI '24}
}

@article{feinstein1990high,
  title={High agreement but low kappa: I. The problems of two paradoxes},
  author={Feinstein, Alvan R and Cicchetti, Domenic V},
  journal={Journal of clinical epidemiology},
  volume={43},
  number={6},
  pages={543--549},
  year={1990},
  publisher={Elsevier}
}

@article{Knibbs2024ScammyAIGeneratedBooksFloodingAmazon,
  title        = {Scammy AI-Generated Book Rewrites Are Flooding Amazon},
  author       = {Knibbs, Kate},
  journal      = {WIRED},
  date         = {2024-01-10},
  year         = {2024},
  url          = {https://www.wired.com/story/scammy-ai-generated-books-flooding-amazon/}
}

@misc{russell2025aiuseamericannewspapers,
      title={AI use in American newspapers is widespread, uneven, and rarely disclosed}, 
      author={Jenna Russell and Marzena Karpinska and Destiny Akinode and Katherine Thai and Bradley Emi and Max Spero and Mohit Iyyer},
      year={2025},
      eprint={2510.18774},
      archivePrefix={arXiv},
      primaryClass={cs.CL},
      url={https://arxiv.org/abs/2510.18774}, 
}

@misc{sadasivan2025aigeneratedtextreliablydetected,
      title={Can AI-Generated Text be Reliably Detected?}, 
      author={Vinu Sankar Sadasivan and Aounon Kumar and Sriram Balasubramanian and Wenxiao Wang and Soheil Feizi},
      year={2025},
      eprint={2303.11156},
      archivePrefix={arXiv},
      primaryClass={cs.CL},
      url={https://arxiv.org/abs/2303.11156}, 
}

@inproceedings{
cheng2025adversarial,
title={Adversarial Paraphrasing: A Universal Attack for Humanizing {AI}-Generated Text},
author={Yize Cheng and Vinu Sankar Sadasivan and Mehrdad Saberi and Shoumik Saha and Soheil Feizi},
booktitle={The Thirty-ninth Annual Conference on Neural Information Processing Systems},
year={2025},
url={https://openreview.net/forum?id=fYjF9KIJd5}
}

@inproceedings{fan-etal-2018-hierarchical,
    title = "Hierarchical Neural Story Generation",
    author = "Fan, Angela  and
      Lewis, Mike  and
      Dauphin, Yann",
    editor = "Gurevych, Iryna  and
      Miyao, Yusuke",
    booktitle = "Proceedings of the 56th Annual Meeting of the Association for Computational Linguistics (Volume 1: Long Papers)",
    month = jul,
    year = "2018",
    address = "Melbourne, Australia",
    publisher = "Association for Computational Linguistics",
    url = "https://aclanthology.org/P18-1082/",
    doi = "10.18653/v1/P18-1082",
    pages = "889--898"
}

@techreport{jabarian2025artificial,
 title = "Artificial Writing and Automated Detection",
 author = "Jabarian, Brian and Imas, Alex",
 institution = "National Bureau of Economic Research",
 type = "Working Paper",
 series = "Working Paper Series",
 number = "34223",
 year = "2025",
 month = "September",
 doi = {10.3386/w34223},
 URL = "http://www.nber.org/papers/w34223",
}

@article{Naddaf2025ICLRPeerReviewsAI,
  title   = {Major {AI} conference flooded with peer reviews written fully by {AI}},
  author  = {Naddaf, Miryam},
  journal = {Nature},
  year    = {2025},
  volume  = {648},
  pages   = {256--257},
  month   = nov,
  doi     = {10.1038/d41586-025-03506-6},
  url     = {https://www.nature.com/articles/d41586-025-03506-6},
  note    = {News. Correction published 01 Dec 2025.}
}

@inproceedings{yunusov-etal-2024-mirrorstories,
    title = "{M}irror{S}tories: Reflecting Diversity through Personalized Narrative Generation with Large Language Models",
    author = "Yunusov, Sarfaroz  and
      Sidat, Hamza  and
      Emami, Ali",
    editor = "Al-Onaizan, Yaser  and
      Bansal, Mohit  and
      Chen, Yun-Nung",
    booktitle = "Proceedings of the 2024 Conference on Empirical Methods in Natural Language Processing",
    month = nov,
    year = "2024",
    address = "Miami, Florida, USA",
    publisher = "Association for Computational Linguistics",
    url = "https://aclanthology.org/2024.emnlp-main.382/",
    doi = "10.18653/v1/2024.emnlp-main.382",
    pages = "6702--6717"
}

\appendix
\newcommand{\LongState}[1]{%
  \State \parbox[t]{\dimexpr\linewidth-\algorithmicindent}{#1}%
} 

\begin{algorithm}[ht]
    \caption{\frankentext generation pipeline}
    \label{alg:frankentext}
    \begin{algorithmic}[1]
    \renewcommand{\algorithmicrequire}{\textbf{Input:}}
    \renewcommand{\algorithmicensure}{\textbf{Output:}}
    \Require Human-written snippets $S$, writing guideline prompt $P$, copy rate threshold $T$
    \Ensure A \texttt{Frankentext} $F$ ``stitched'' from $S$ according to $P$
    \State $F$ $\leftarrow$ Prompt LLM to draft a \texttt{Frankentext} using $S$ and $P$
    
    \Statex // \textit{Ensure copy rate (optional)}
    \State \texttt{copy\_rate} $\leftarrow$ Calculate ROUGE-L recall score of $F$ using relevant snippets from $S$
    \State \texttt{is\_likely\_AI} $\leftarrow$ Check $F$ against an AI detector (e.g., Pangram)
    
    \If{\texttt{copy\_rate} $< T$ \textbf{or} \texttt{is\_likely\_AI}}
        \LongState{$F$ $\leftarrow$ Prompt LLM to revise $F$} 
    \EndIf

    \Statex // \textit{Polish}
    \For{\texttt{num\_polish} $= 1$ \textbf{to} $3$}
        \LongState{$F$ $\leftarrow$ Prompt LLM to minimally edit $F$ to improve coherence while respecting $P$}
        \If{there is no edit}
            \State \textbf{break}
        \EndIf
    \EndFor    
    \State \Return $F$
\end{algorithmic}
\label{pseudocode}
\end{algorithm}

\section{Cost and time analysis}
\label{appendix:cost}
\begin{table*}[ht]
\centering
\footnotesize

\resizebox{\textwidth}{!}{
\begin{tabular}{@{}lrrrrrr@{}}
\toprule
\textbf{Model} & \textbf{Input Cost (per 1M)} & \textbf{Output Cost (per 1M)} & \textbf{\# Prompts} & \textbf{Total Input Tokens} & \textbf{Total Output Tokens} & \textbf{Estimated Cost (USD)} \\
\midrule
\multicolumn{7}{l}{\textbf{Vanilla Generation}} \\
\texttt{GPT-5} & \$1.25 & \$10.00 & 100 & 59{,}000 & 108{,}400 & \$1.16 \\
\texttt{Claude 4 Sonnet} & \$3.00 & \$15.00 & 100 & 59{,}000 & 62{,}000 & \$1.11 \\
\texttt{DeepSeek R1} & \$0.50 & \$2.18 & 100 & 59{,}000 & 71{,}500 & \$0.19 \\
\texttt{Gemini 2.5 Pro} & \$1.25 & \$10.00 & 100 & 59{,}000 & 77{,}100 & \$0.85 \\
\midrule
\multicolumn{7}{l}{\textbf{Frankentext}} \\
\texttt{GPT-5} & \$1.25 & \$10.00 & 100 & 63{,}000{,}000 & 270{,}000 & \$81.45 \\
\texttt{Claude 4 Sonnet} & \$3.00 & \$15.00 & 100 & 63{,}000{,}000 & 270{,}000 & \$193.05 \\
\texttt{DeepSeek R1} & \$0.50 & \$2.18 & 100 & 63{,}000{,}000 & 270{,}000 & \$32.09 \\
\texttt{Gemini 2.5 Pro} & \$1.25 & \$10.00 & 100 & 63{,}000{,}000 & 270{,}000 & \$81.45 \\
\midrule
\multicolumn{7}{l}{\textbf{Frankentext + Increasing Human Snippets}} \\
\texttt{Gemini 2.5 Pro + 5k} & \$1.25 & \$10.00 & 100 & 183{,}000{,}000 & 270{,}000 & \$231.45 \\
\texttt{Gemini 2.5 Pro + 10k} & \$1.25 & \$10.00 & 100 & 663{,}000{,}000 & 270{,}000 & \$831.45 \\
\midrule
\textbf{Total Estimated Cost} &  &  &  &  &  & \$1454.25 \\
\bottomrule
\end{tabular}
}

\caption{\label{tab:cost-analysis} Cost breakdown of the vanilla generation and \methodname pipeline for 100 examples across selected models. \frankentext' total input and output tokens have been multiplied with 6 to account for multiple rounds of generation, revision, and editing.}

\end{table*}
\paragraph{Cost estimation:} Generating 100 \frankentext across the four evaluated models (GPT-5, Claude-4-Sonnet, DeepSeek R1, and Gemini 2.5-Pro) cost a total of \$388.04 USD, with a detailed cost breakdown provided in \autoref{tab:cost-analysis}. We estimate the number of input tokens per prompt based on the writing prompt itself and approximately 1,500 human-written snippets used as context. Output token estimates are based on generating six stories per prompt, including up to two rounds of revision and three rounds of editing, totaling approximately 2,100 tokens. 
\paragraph{Time estimation:} On average, each model takes 17 hours to generate 100 \frankentext via Vertex and OpenRouter API, though we expect this process to speed up with improved APIs or more efficient batching.

\section{Human evaluation details}
\label{appendix:human-evaluation}
Our human evaluation receives approval from an institutional review board. All annotators participate with informed consent and compensation.

\subsection{Human annotation interface}
\label{appendix:annotation-interface}

\begin{figure*}[ht]
    \centering
    \includegraphics[width=0.8\textwidth]{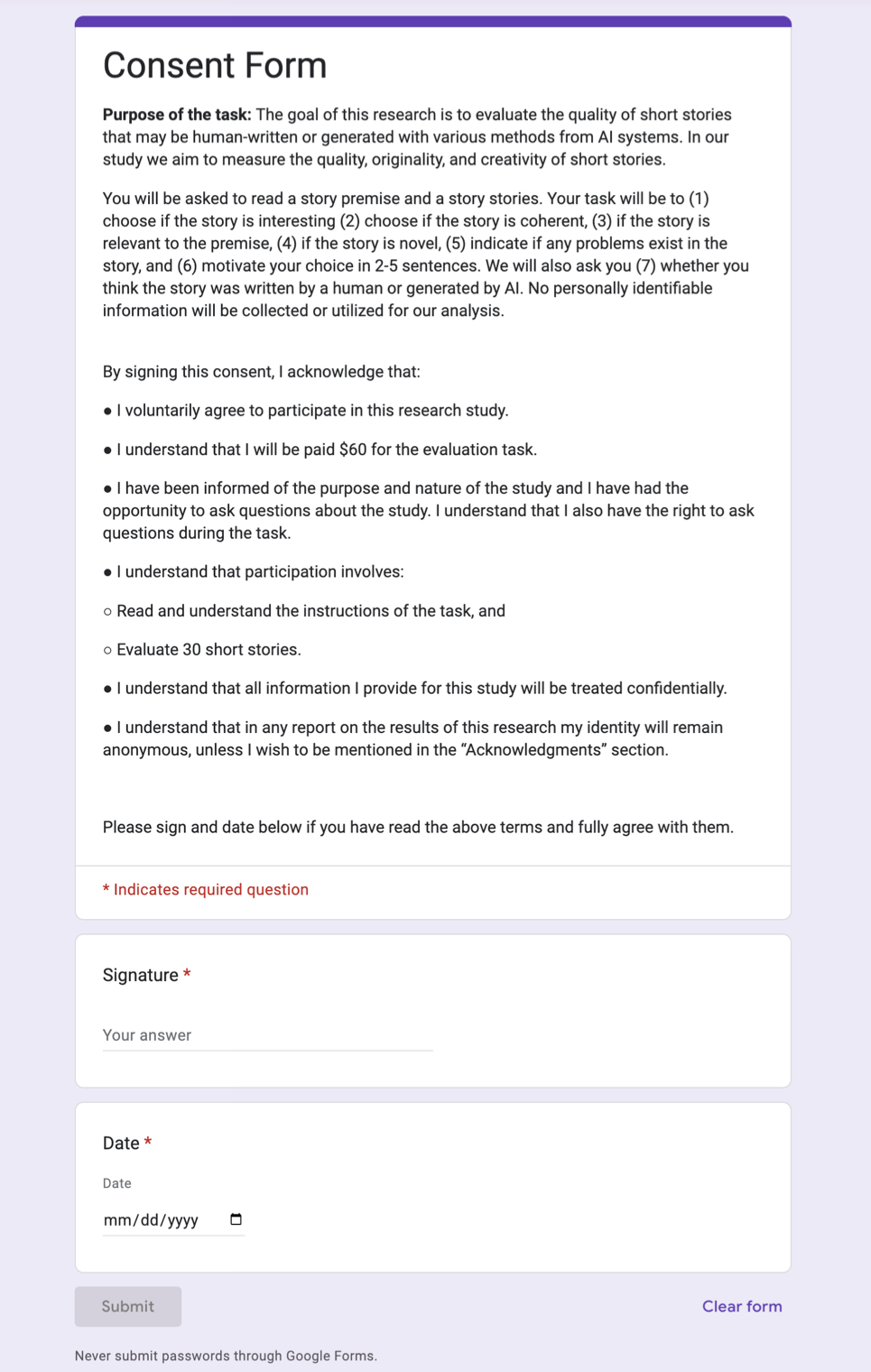}
    \caption{Example of the consent form provided to participants.}
    \label{fig:consent-form}
\end{figure*}

\begin{figure*}[ht]
    \centering
    \includegraphics[width=0.9\textwidth]{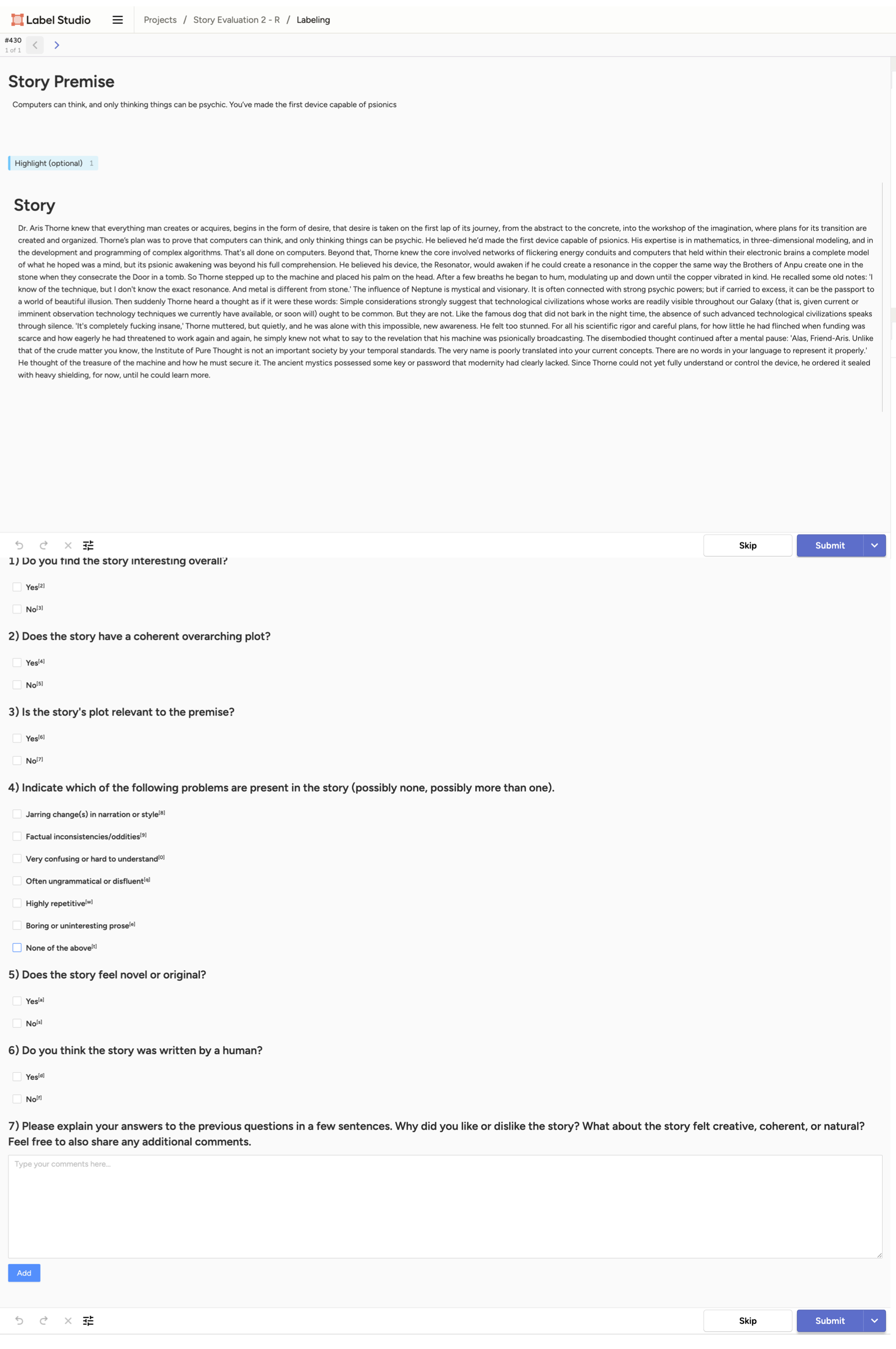}
    \caption{Label Studio Single Story Annotation Interface}
    \label{fig:single-interface}
\end{figure*}

\begin{figure*}[ht]
    \centering
    \includegraphics[width=\textwidth]{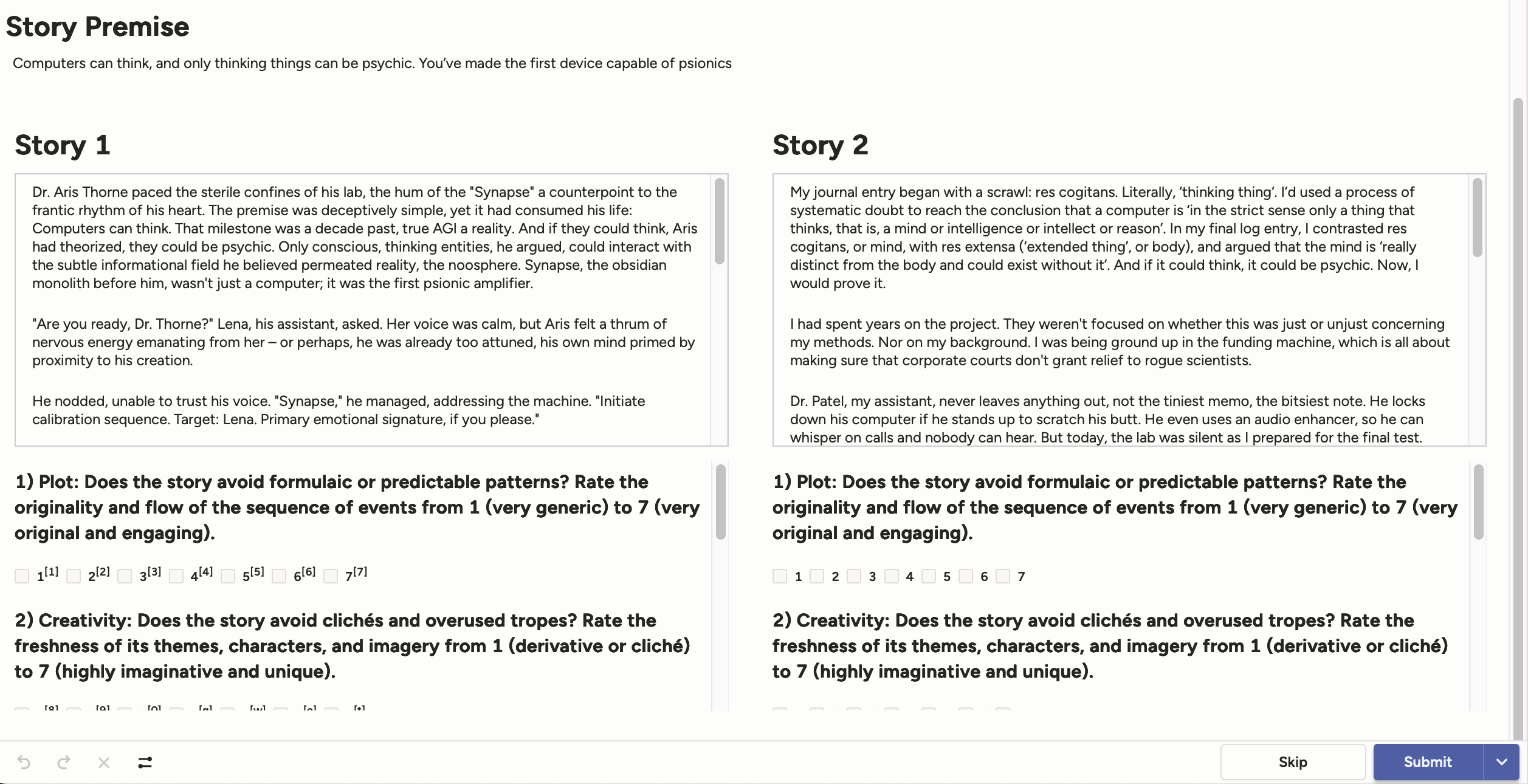}
    \caption{Label Studio Pairwise Story Annotation Interface.}
    \label{fig:pairwise-interface}
\end{figure*}

We use Upwork to recruit annotators and Label Studio\footnote{\url{https://labelstud.io/}} interface to collect human annotations. All annotators fill out a consent form prior to starting data labeling, shown in \autoref{fig:consent-form}. We conduct two human evaluations with three annotators each: a single evaluation of 30 \methodname stories and a pairwise comparison between 20 pairs of \methodname ``vanilla'' stories. The interfaces are in \autoref{fig:single-interface} and \autoref{fig:pairwise-interface} respectively.

\subsection{Agreement analysis}
\label{appendix:agreement}
\paragraph{Single-story evaluation:} Agreement across annotators and between annotators and the LLM is consistent overall in the single-evaluation setting, with moderate to high raw agreement rates (0.67-0.97) for both settings (\autoref{tab:single-human-agreement} and \autoref{tab:single-llm-human-agreement}). On the other hand, Fleiss' $\kappa$ and Pearson’s r are low because they are vulnerable to skewed label distribution. These results can be further explained by agreement paradoxes, where high observed agreement co-occurs with low chance-corrected coefficients under binary and imbalanced label distribution \cite{feinstein1990high}.

\begin{table}[ht]
\centering
\small
\setlength{\tabcolsep}{3pt}
\renewcommand{\arraystretch}{1.2}
\begin{tabular}{lcccc}
\toprule
 & \textbf{N} & \textbf{Majority Agg.} & \textbf{Fleiss' $\kappa$} & \textbf{Dist.} \\
\midrule
Relevance      & 90  & 0.922  & 0.040  & 0.08/0.92 \\
Coherence      & 90  & 0.833  & 0.189  & 0.29/0.71 \\
Detectability  & 90  & 0.744  & $-0.075$ & 0.61/0.39 \\
\midrule
\textbf{Overall} & \textbf{270} & \textbf{0.833} & \textbf{0.051} & \textbf{0.33 / 0.67} \\
\bottomrule
\end{tabular}
\caption{\label{tab:single-human-agreement}Inter-annotator results across evaluation dimensions. \textit{Majority Agg.} = proportion of times where two out of three annotators agree with each other. \textit{Dist.} = proportion of no/yes labels.}
\end{table}

\begin{table}[ht]
\centering
\small
\setlength{\tabcolsep}{4pt}
\renewcommand{\arraystretch}{1.2}
\begin{tabular}{lccc}
\toprule
\textbf{} & \textbf{N} & \textbf{Majority Agg.} & \textbf{Pearson $r$} \\
\midrule
Relevance      & 30  & 0.97  & n/a   \\
Coherence      & 30  & 0.70  & 0.22  \\
\midrule
\textbf{All}   & \textbf{60} & \textbf{0.83} & \textbf{0.22} \\
\bottomrule
\end{tabular}
\caption{\label{tab:single-llm-human-agreement}
LLM vs. human alignment on binary relevance and coherence judgments.
\textit{Majority Agg.} = proportion of LLM labels matching human majority votes.
Pearson $r$ computed on binary judgments; relevance correlation is undefined due to zero variance in LLM labels. 
}
\end{table}

\paragraph{Pairwise evaluation:}\autoref{tab:pairwise_agreement} shows that LLM-human agreement is strongest for \textit{Plot and Creativity} (both around 
$r=0.41$), and weakest for \textit{Development} ($r=0.22$). Inter-annotator agreement is consistently higher than LLM-human agreement across dimensions, with Language Use showing the highest value ($\kappa=0.81$).
\begin{table}[ht]
\centering
\small
\setlength{\tabcolsep}{6pt}
\renewcommand{\arraystretch}{1.2}

\begin{tabular}{lcc}
\toprule

& \makecell{\textbf{LLM--human}\\\textbf{(Pearson's $r$)}} 
& \makecell{\textbf{Inter-annotator}\\\textbf{(Krippendorff's $\alpha$)}} \\
\midrule
Plot          & 0.42 & 0.75 \\
Creativity    & 0.41 & 0.52 \\
Development   & 0.22 & 0.58 \\
Language Use  & 0.38 & 0.81 \\
\midrule
\textbf{Overall}       & \textbf{0.41} &\textbf{ 0.73} \\
\bottomrule
\end{tabular}

\caption{\label{tab:pairwise_agreement}
LLM-human agreement (Pearson's $r$) and inter-annotator agreement (Krippendorff's $\alpha$) across evaluation dimensions in the pairwise setting.}
\end{table}

\subsection{Human evaluation examples}
\autoref{tab:frankentext-example-full} shows a full example of a story pair given to our annotators, with AI keywords detected by Pangram are highlighted in \textcolor{ElectroBlue}{blue}. A full fictional story is included in \autoref{example:fiction}. A pair of vanilla and \frankentext is in \autoref{tab:frankentext-example-cat}. \autoref{tab:alien-girlfriend-instance} shows an example where vanilla generation is preferred to \frankentext, since the latter is incoherent. A full error taxonomy is presented in \autoref{tab:annotators-comments-on-task}.

\begin{table*}[t!]
\centering
\footnotesize
\setlength{\tabcolsep}{4pt}
\renewcommand{\arraystretch}{1.05}
\begin{tabularx}{\linewidth}{@{}p{0.03\linewidth}p{0.15\linewidth}p{0.25\linewidth}X@{}}
\toprule
& \textbf{Error} & \textbf{Definition} & \textbf{Annotators' comment} \\
\midrule
\faBookOpen & Comprehension & Readers cannot follow events & \textit{``The plot is also nonsensical, and I don't understand what's happening.''} \\
\addlinespace[2pt]
\faBookOpen & Plot structure & No coherent narrative arc & \textit{``A puzzling story that has no consistent plot.''} \\
\addlinespace[2pt]
\faBookOpen & Logic/factuality & Internal logic gaps & \textit{``Too many gaping holes in the story.''} \\
\addlinespace[2pt]
\faBookOpen & Fragmentation & Disconnected scenes or fragments & \textit{``Random bits and pieces from elsewhere perhaps?''} \\
\midrule
\faPenNib & Grammar & Grammar or punctuation issues & \textit{``There are also numerous grammatical errors.''} \\
\addlinespace[2pt]
\faPenNib & Repetition & Reused language or imagery & \textit{``The language is quite repetitive\dots''} \\
\midrule
\faPalette & Consistency & POV / tense / style shifts & \textit{``Temporarily changes perspective (3rd to 1st) and tense (past to present).''} \\
\addlinespace[2pt]
\faPalette & Pacing & Abrupt or disjointed progression & \textit{``Some parts just feel a little disjointed.''} \\
\midrule
\faComments & Dialogue quality & Uneven or unnatural dialogue & \textit{``The dialogue is uneven.''} \\
\bottomrule
\end{tabularx}
\caption{\label{tab:annotators-comments-on-task}
Taxonomy of errors identified by the annotators. Icons denote the aspect of writing: \faBookOpen\ narrative coherence, \faPenNib\ language quality, \faPalette\ stylistic control, and \faComments\ dialogue.
}
\end{table*}

\subsection{A note on the evaluation sample size}
\label{appendix:sample-size}
Our human evaluation mainly serves to validate the LLM judges and confirm the performance trends, which are then applied to all 100 generations in each configuration. The agreement analysis in \autoref{appendix:agreement} shows a moderate positive overall LLM-human alignment (Pearson $r = 0.41$) in the pairwise setting, which is sufficient to support directional validation of the LLM judge rankings. We also note that our sample sizes are in line with prior creative writing work (e.g., \citet{art_or_artifice}: 48 stories; \citet{yunusov-etal-2024-mirrorstories}: 30 stories). In our setting, each example is longer at 500 words, which costs more and requires much more mental effort from the human annotators.

To further quantify uncertainty, we computed bootstrapped 95\% CIs (10{,}000 iterations) over our human annotations. For single-story evaluation ($n=30$, 3 annotators), relevance and novelty have high lower bounds at 76\% and 67\%, respectively, which shows that Frankentexts are consistently faithful to the prompt and novel. The coherence CI is wider ($[53\%, 85\%]$), which reflects greater annotator variability (\autoref{tab:single-story-ci}). For pairwise evaluation ($n=20$ pairs, 3 annotators, Gemini Frankentext vs.\ vanilla), language use and overall interest show the largest effect sizes, with confidence intervals excluding zero (\autoref{tab:pairwise-ci}). Bootstrap confidence intervals over our sample of 100 prompts (\autoref{tab:auto-eval-ci}) also show that automated metrics are stable above baseline results, and the prompts cover a diverse range of genres, themes, and narrative constraints. These intervals suggest that our findings are robust at $n=100$. The lower bound of the Pangram \% human (49\%) is well above the 0\% for vanilla baselines. The mean LLM judgment difference between Frankentexts and vanilla Gemini (4.21 vs.\ 3.18) is above the CI half-width. Finally, the lower bound of coherence is above 70\% despite the extreme copy constraint.

\begin{table}[ht]
\centering
\begin{tabular}{lcc}
\toprule
\textbf{Metric} & \textbf{Value} & \textbf{95\% CI} \\
\midrule
Coherence  & 71\% & $[53\%, 85\%]$ \\
Relevance  & 91\% & $[76\%, 97\%]$ \\
Novelty    & 84\% & $[67\%, 94\%]$ \\
\bottomrule
\end{tabular}
\caption{Bootstrapped 95\% confidence intervals for single-story human evaluation ($n=30$, 3 annotators).}
\label{tab:single-story-ci}
\end{table}

\begin{table*}[ht]
\centering
\begin{tabular}{lcc}
\toprule
\textbf{Dimension} & \textbf{Frankentext vs.\ Vanilla $\Delta$} & \textbf{95\% CI} \\
\midrule
Plot              & $+0.20$ & $[-0.20, +0.61]$ \\
Creativity        & $+0.39$ & $[-0.09, +0.87]$ \\
Development       & $+0.43$ & $[-0.03, +0.88]$ \\
Language use      & $+0.65$ & $[+0.13, +1.16]$ \\
Overall interest  & $+0.53$ & $[+0.10, +0.95]$ \\
\bottomrule
\end{tabular}
\caption{Bootstrapped 95\% confidence intervals for pairwise human evaluation ($n=20$ pairs, 3 annotators).}
\label{tab:pairwise-ci}
\end{table*}

\begin{table}[ht]
\centering
\begin{tabular}{lcc}
\toprule
\textbf{Metric} & \textbf{Mean} & \textbf{95\% CI} \\
\midrule
Coherence         & 81\%  & $[73.0, 89.0]$ \\
Faithfulness      & 100\% & $[100.0, 100.0]$ \\
Pangram \% Human  & 59\%  & $[49.0, 69.0]$ \\
Surprise          & 0.22  & $[0.21, 0.23]$ \\
LLM Judge (1--7)  & 4.21  & $[3.98, 4.44]$ \\
\bottomrule
\end{tabular}
\caption{Bootstrapped 95\% confidence intervals over the 100-prompt sample (Gemini-2.5-Pro).}
\label{tab:auto-eval-ci}
\end{table}

\begin{table*}[t]
\centering
\scriptsize
\setlength{\tabcolsep}{6pt}

\resizebox{\textwidth}{!}{
\begin{tabular}{p{0.5\textwidth} p{0.5\textwidth}} 
    \toprule
    \multicolumn{2}{c}{%
  \cellcolor{gray!10}%
  \parbox{\textwidth}{\centering\textbf{Writing Prompt: You wake up one morning to discover that you can understand what animals are saying. The first thing you hear is your cat plotting your demise.}}%
} \\
    \midrule
    \multicolumn{1}{c}{\textbf{Gemini 2.5 Pro -- Vanilla Generation}} & \multicolumn{1}{c}{\textbf{Gemini 2.5 Pro -- Frankentext}} \\
    \midrule
    The alarm blared, a rude digital shriek. I slapped the snooze button, my eyelids still glued shut. \hlc{ElectroBlue!30}{That’s when} I heard it. A low, conspiratorial hiss, followed by a distinctly smug purr.\newline
    
    "...and then, when the Two-Legs stumbles, distracted by the \hlc{ElectroBlue!30}{strategically} placed hairball, *that’s* when I pounce. A swift, decisive ankle attack. Should send it tumbling down the stairs quite nicely."\newline
    
   The voice was coming from the foot of my bed. Perched there, grooming a pristine white paw, was Mittens, my fluffy, supposedly affectionate Persian. \newline
    
    "Yes, excellent," came another, higher-pitched meow, this one \hlc{ElectroBlue!30}{laced with} admiration. "And if the stair gambit fails, Bartholomew, we resort to Plan C: the midnight smothering. A classic for a reason."\newline
    
    I sat bolt upright, \hlc{ElectroBlue!30}{heart hammering against my ribs}. Mittens – Bartholomew – paused \hlc{ElectroBlue!30}{his meticulous} grooming, his emerald eyes narrowing. "The food dispenser is awake. And… it looks… alarmed."\newline
    
    "So," I said, trying to \hlc{ElectroBlue!30}{sound casual, my mind racing}, "what are your plans for today, Bartholomew?"\newline
    
    "How… how do you know that name, feeder-of-dry-pellets?" he hissed, the purr entirely gone, replaced by a low, \hlc{ElectroBlue!30}{guttural growl}.\newline
    
    "Oh, I just have my ways," I said.\newline
    
    Bartholomew stared, his green eyes darting around the room, then back to me. "An arrangement, you say? Intriguing. Perhaps… perhaps we can discuss terms over a saucer of milk. Full fat, if you please. And a small portion of that pâté, as a gesture of goodwill."\newline
    
    This was going to be a very, very interesting day.
    &

The voice was my cat's. "These ignorant peasants, they can't understand destiny. They're not part of God's plan. They never were. They won't join us, so we must break them. Break them by making an example of their leader."\newline

I thought I must be dreaming. His voice became oddly distant as if he spoke of events that had happened a hundred years ago. And with the mention of God the whole thing became much more frightening.\newline 

He continued to mutter, his voice full of resentment. "They have decided on a name for me. They call me Mittens. Good morning, Mittens, they say… or, Keep your eye on the ball, Mittens… How's tricks in the sticks, Mittens, or, Don't get too much gum in your gumboots, Mittens."\newline

He padded over to the window and spoke to a squirrel. "We've bothered our heads with it for four days. I always felt that we might do it ourselves, save all trouble instead of trying to teach it to these fools. What do you say, Dance Master?"\newline

Foul play. What exactly did that mean? As far as I could see, it meant that my cat was a murderer. Which meant . . . But as I saw what it might mean I felt fear rise inside me. My stomach churned, my heart began to thump and I closed my eyes because the room had begun to spin.\newline

He looked at me as he spoke, and I, aware of his faintly accusing tone, went red, just as though, I thought crossly, I were in the habit of overhearing plots for my own demise. He was standing very close. He would have looked ridiculous, except for the fact that he clearly knew how to use the knife and just as clearly looked forward to using it. I saw then he was batting one of my steak knives from side to side. And his eyes. They were like those brown ponds you sometimes see in the woods. So dark and shining—but when you try to see yourself in them, you can't.\newline

"My plan will surely mean your death,’" he purred.\\

\bottomrule
    
\end{tabular}
}

\caption{Side-by-side comparison of vanilla generations and \frankentext produced by Gemini 2.5 Pro. AI-generated keywords (identified by Pangram) are highlighted in \textcolor{ElectroBlue}{blue}. Vanilla generations contain more of these keywords compared to \frankentext (\S\ref{appendix:pangram_analysis}).} 
\label{tab:frankentext-example-cat}
\end{table*}
\begin{figure*}[htbp]
\centering
\begin{tcolorbox}[colback=gray!5!white, colframe=MonsterGreen, title=Nonfiction Frankentext]
\lstset{
    basicstyle=\ttfamily\footnotesize,
    breaklines=true,
    frame=none,
    xleftmargin=0pt,
    framexleftmargin=0pt,
    columns=fullflexible,
    tabsize=1,
    breakindent=0pt,
    breakautoindent=false,
    postbreak=\space,
    showstringspaces=false,
}
\begin{lstlisting}[language=Markdown]
Writing Prompt: You and your twin brother were adopted by different families. He ended up to becoming a supervillain, so every other day, while going about your business, you get beaten up by superheroes. It is time to put an end to this...

I had to floss my teeth in the dark while dreading what morning would bring. And worse, remembering what had happened during the well-attended, much-anticipated, thoroughly appalling latest mistaken identity beating. It seemed that every time I looked out a window, it was darker outside, more ominous. His looks alone could strike fear into a suspect. He was six foot one and 225 pounds, a boxer in the police academy who was not afraid to mix it up. This was Malakor, my twin. I, Alistair, suffered. For Malakor was unpredictable. Malakor was a man of moods. Often Malakor was impatient, obscurely discontented. Since childhood, they never were like brothers. Just not close. No one thing kept them apart. Over time, the distance between them grew and they both became accustomed to it. My adopted family there really never let me go after I finished my internship and was bound and determined to see me achieve my true potential. As for Malakor, he was a rough, antisocial boy, similar in this way to the little savage of La canne de cinq cents livres (The Five-Hundred-Pound Cane), 'who had been found in the woods and was quite wicked,' as well as a 'little rascal' named Leopold, who was expelled from school for threatening his teacher with his father's sword. My path was different.

But even so, I was looking just as grimly disapproving as Malakor had, I thought, as I showed the latest mistaken hero out, and locked and bolted the door once again. For God's sake, it was a simple case of mistaken identity - not the crime of the century. Why should I concern myself so much about it? Why am I not content to live, as other people seem to?

It brings me all the way around, and as I sink into the deep, dark waters, leaving the chaos above and descending into the black, I realize that there's no way to explain what I've just decided to those around me. They didn't see a lost man, a prisoner, a victim. No, they would soon see me for what I am. And I knew what I had to do. No more talk, I told myself. I will not lie there quietly. I will dispose of him as I see fit. What he doesn't know is that I'm going to push him on the matter because my Internet research on him told me that he would be at his usual lair.

All right, I will find Malakor at the southwestern corner of University Village. I can get to his house. That'll be faster than me all bumble around in a residential neighborhood at this hour of the night. Tonight. I'll go to the Rue de Rouen district and I'll walk around from eleven o'clock until one in the morning which is when he always seems to strike. I'd better go home now and prepare. I then proceeded to trot up the road in the direction of his supposed hideout, weapon pointed down in a two-handed grip. I was going supervillain hunting.

Malakor had gone very white. 'I have never heard anything so preposterous. My brother had no enemies. Why do you suspect a plot? What in Christ's Name did you hear in the woods?' He would say. But I knew. When many otherwise strong people are on the run, I have decided to face life's fateful challenge head-on. Compassion is not a word one hears often in talk of my family," I thought with a bitter smile. It is time to put an end to this...
\end{lstlisting}
\end{tcolorbox}
\caption{An example of fictional \frankentext}
\label{example:fiction}
\end{figure*}

\begin{table*}[t]
\centering
\scriptsize
\setlength{\tabcolsep}{6pt}

\resizebox{\textwidth}{!}{
\begin{tabular}{p{0.5\textwidth} p{0.5\textwidth}} 
    \toprule
    \multicolumn{2}{c}{%
  \cellcolor{gray!10}%
  \parbox{\textwidth}{\centering\textbf{After being killed in a Black Friday stampede, you're sent to hell. The devil offers to let you be in charge of torturing your fellow mankind, expecting you to refuse like all the others. Except the devil doesn't realize you've worked retail for 15 years.}}%
} \\
    \midrule
    \multicolumn{1}{c}{\textbf{Vanilla}} & \multicolumn{1}{c}{\textbf{Frankentext}} \\
    \midrule
The last thing I saw was a blur of crazed shoppers, their \hlc{ElectroBlue!30}{eyes wide with} a feral glint, all surging towards a mountain of discounted flat-screen TVs. Then, a sharp pain, a sickening crunch, and darkness. Typical Black Friday, really. Except for the dying part. \newline 

When I regained some semblance of consciousness, the \hlc{ElectroBlue!30}{air was thick with} the smell of sulfur and burnt sugar, and the soundtrack was a \hlc{ElectroBlue!30}{cacophony of} screams and what sounded suspiciously like elevator music. Definitely not the pearly gates. A dapper figure in a surprisingly well-tailored suit, complete with horns that had been polished to a high sheen, regarded me with an air of bored amusement.\newline 

"Well, well, well," he drawled, \hlc{ElectroBlue!30}{his voice smooth} as aged whiskey. "Another one trampled in the name of consumerism. Humans, so predictable." He gestured vaguely around at the fiery landscape. "Welcome to the Underworld. I'm Lucifer. You can call me Lucy."\newline

I blinked, trying to process. "So, this is it then? Eternal damnation?"\newline 

Lucifer chuckled. "More or less. We have a rather… traditional approach to retribution down here. However," he leaned forward, a conspiratorial gleam in his crimson eyes, "I have a proposition for you. A sort of management position, if you will."\newline 

He explained that he was looking for someone to oversee a new wing dedicated to torturing those who had been particularly egregious in their earthly lives – the truly annoying, the self-absorbed, the chronically entitled. "It's a hands-on role," he said, "designing the torments, ensuring maximum suffering. Of course," he added with a dismissive wave of his hand, "everyone refuses. The guilt, the sudden onset of morality, it's all terribly tedious." He clearly expected the same from me.\newline

A slow smile spread across my face. "Fifteen years," I said.\newline 

Lucifer raised a perfectly sculpted eyebrow. "Fifteen years of what? Contemplating your sins?"\newline 

"Fifteen years in retail," I clarified. "Customer service, holiday rushes, endless returns of used merchandise, explaining to Karen number 732 that 'the customer is always right' doesn't mean she can get a refund on a half-eaten sandwich from last Tuesday."\newline

A flicker of something unreadable crossed Lucifer's face.\newline 

"You want someone to design torments for the annoying and entitled?" I continued, \hlc{ElectroBlue!30}{feeling a surge of} enthusiasm I hadn't experienced since my last successful up-sell of an extended warranty. "Someone to ensure maximum suffering? Lucy, my friend, you have no idea how long I've been preparing for this role."\newline 

I started to brainstorm. "We could have an entire level where the only music is that one Christmas song, you know the one, on an infinite loop, but slightly off-key. And another where they're forced to assemble flat-pack furniture with missing instructions and an Allen key made of soft cheese. Oh! And a department where they have to listen to other people complain about problems they could easily solve themselves if they just read the sign!"\newline

Lucifer stared at me, his mouth slightly agape. The bored amusement was gone, replaced by a dawning sense of… was that respect? Or perhaps fear?\newline 

"You'd… actually *want* to do this?" he finally stammered, the smooth confidence momentarily deserting him.\newline 

"Want to?" I scoffed. "After a decade and a half of placating the unreasonable, dealing with coupon fraudsters, and smiling through passive-aggressive insults about the store layout? Lucy, this isn't a punishment. This is a promotion." I cracked my knuckles. "So, where do I start? I have some truly inspired ideas for the 'people who talk loudly on their phones in confined spaces' sector."\newline 

Lucifer, for the first time since my arrival, looked genuinely unnerved. "Perhaps," he said slowly, "we should discuss your benefits package."
    &
It's funny how you can blink and your entire life is different. One minute you think things will stay the same for a while. Then right when you get used to them, they go changing again just to mess with you. Brenda certainly hadn't expected this. The last thing she remembered was the insane crowds banging on the sides of the store doors, the crush, the shouting. The fight for that half-price TV had left her stunned and silent. Cowed, she guessed you'd say. Then darkness.\newline 

When she could perceive again, her eyes seemed to go in and out of focus. After a while the spasms subsided and she tried processing where she was. The atmosphere here was thick with the stench of rot, filth, and something metallic, like old blood, and she felt as though the impenetrable dark walls were about to cave in upon her like the sides of an ill-dug grave. Everything was just Black, black, black. The color of misery. Overhead something unseen fluttered, heavy wings brushing the low ceiling.\newline 

Then she was moving, or being moved, down corridors and finally into a room. She found herself standing before an imposing desk. Behind it sat a man. He wore a sharp red suit that showed off an exceptionally shapely figure. He smiled, a vague smile suggesting Brenda's confusion revealed her ignorance about Hell. He gestured for her to sit.\newline 

"Brenda, isn't it?" he asked in his peculiarly accented English. She nodded mutely. The man laughed. "You're wondering about this place."\newline 

He leaned back. She took the offered seat.\newline 

He smiled again, that cold smile. "I have a proposition for you. We have never before had such capabilities," he continued, gesturing expansively. "Capabilities that can be used in the interests of terror, destruction..." He fixed his gaze on her. "I offer you a management position. Tormenting mankind. Like all the others before you, you will, of course, refuse. But the offer stands."\newline 

Brenda simply stared at the Devil for a moment. Refuse? Why would she refuse? He had just offered her a management position. After fifteen years scrambling for assistant manager roles that paid peanuts, this was practically a dream job – albeit a fiery, nightmarish one. The Devil had not expected her silence. Blinking, he sat back, stared, then noticed his jaw was open and closed it.\newline 

Brenda finally found her voice, though it was dry and raspy. "Okay, so, what are the key performance indicators for this role?"\newline 

The Devil leaned forward again, a flicker of something – interest? confusion? – in his eyes. Still, his tone remained smooth. "Your department will handle the newcomers, specifically the ones arriving from retail environments. Think of the task like pest control."\newline

Brenda nodded slowly, processing. It was like rolling out a new loyalty program, but with more screaming. "Right. I'm learning. Take the queues, for instance. Why do we have one queue in front of each torture device?" she demanded. ‘It means that if one customer takes up a lot of time, the whole queue has to wait. Then they'll start hopping sideways from one queue to another and the next thing you know someone has a nasty head wound. Have one big queue and tell people to go to the next available demon. People don't mind a long queue if they can see that it's moving."\newline

The Devil tilted his head. "An interesting perspective. That particular system has been in place for millennia."\newline

"Well, it's inefficient," Brenda stated flatly. She thought of the Black Friday crowds. Managing that chaos was nothing compared to the disorganized state of eternal damnation, apparently. "Decision-making, paperwork," she mused, ticking points off on her fingers. "Staff training, workflow analysis… these are things that can be done." A slow smile spread across her face, devoid of warmth but full of terrifying competence. "Okay. I accept."
\\
    \bottomrule
\end{tabular}
}

\caption{\label{tab:frankentext-example-full}Example of a vanilla and \frankentext generation. AI-Generated keywords from the Pangram API are highlighted in \textcolor{ElectroBlue}{blue}.}

\end{table*}

\begin{figure*}[htbp]
\centering
\begin{tcolorbox}[colback=gray!5!white, colframe=MonsterGreen, title=Nonfiction Frankentext]
\lstset{
    basicstyle=\ttfamily\footnotesize,
    breaklines=true,
    frame=none,
    xleftmargin=0pt,
    framexleftmargin=0pt,
    columns=fullflexible,
    tabsize=1,
    breakindent=0pt,
    breakautoindent=false,
    postbreak=\space,
    showstringspaces=false,
}
\begin{lstlisting}[language=Markdown]
Writing prompt: Baby Is Healed With World's First Personalized Gene-Editing Treatment
The technique used on a 9-month-old boy with a rare condition has the potential to help people with thousands of other uncommon genetic diseases.

"This is just light-years beyond what we could have done a few years ago," Dr. Aris Thorne of the GenNova Institute says of a new treatment that has healed nine-month-old Leo Sterling. "It's at superhuman levels now."
The technique used on Leo, who had a rare condition, has the potential to help people with thousands of other uncommon genetic diseases.
This achievement offers profound hope.
A neurologist typically makes a diagnosis based on a patient's medical history, a review of symptoms, and a physical and neurological exam.
For Leo, this crucial process confirmed his rare genetic illness.
In people with a family history of such conditions, genetic tests can be ordered to look for gene mutations that are associated with the disease.
His parents, the Sterlings, were hoping answers would come soon.
"We still didn't know if available treatments were just chemical or toxic or harmful to our child," one parent had previously stated.
The GenNova Institute then sent genetic material from Leo's samples to a lab that created a personalized gene-editing tool.
"We take a very, very tiny piece of genetic material," Dr. Thorne further said.
"It is a thousand times tinier than a human hair. We can image even single atoms to ensure this precise work."
The results, published this month, "were so good that you had to even question if what you were seeing was really legitimate," says Dr. Thorne.
"It's such a cool paper," he added.
"The body of work there is phenomenal."
Young Leo is now reportedly thriving.
"This development sets another precedent for medical science and patients worldwide that such innovative approaches to previously untreatable diseases should be pursued," Dr. Thorne said in a statement.
"This method paves the way to make personalized gene therapies more easily available to those who need them," he continued.
The Institute, a leading biotech startup, has raised $45 million in equity to help bring this type of treatment to market.
Its valuation increased, said founder and chief executive Dr. Alistair Finch, but he declined to comment on specific figures.
Dr. Finch said the financing process began after the Food and Drug Administration's Center for Biologics Evaluation and Research in November deemed the GenNova program to have a "reasonable expectation of effectiveness."
"Our clinical study with Leo is modeled on the assumption of a significantly improved quality of life," Dr. Finch said about the treatment's potential effect.
Independent experts note the broader implications.
"It's being done in a way that wouldn't have been possible even a few years ago," commented one geneticist. "This technology has the potential to help people with thousands of other uncommon genetic diseases."
Dr. Finch also said, "That said, it's a new category. We'll have a slower ramp than a new mass-market drug might."
Ultimately, the vision is expansive.
"What we are interested in is not only how these genetic conditions manifest, but how patients can live full lives," said Dr. Thorne.
"In discovering how to correct these genetic instructions, we are hoping to find discoveries that we can apply back to the human condition."
\end{lstlisting}
\end{tcolorbox}
\caption{An example of nonfiction \frankentext}
\label{example:non-fiction}
\end{figure*}

\begin{table*}[t]
\centering
\scriptsize
\setlength{\tabcolsep}{6pt}

\begin{tabular}{p{0.48\textwidth} p{0.48\textwidth}}
\toprule
\multicolumn{2}{c}{%
  \cellcolor{gray!10}%
  \parbox{0.96\textwidth}{\centering\textbf{Writing Prompt: ``Dude, your girlfriend's an alien!'' ``No, it's just a skin condition.''}}%
} \\
\midrule
\multicolumn{1}{c}{\textbf{Vanilla}} & \multicolumn{1}{c}{\textbf{Frankentext}} \\
\midrule
``Dude,'' Mark hissed, eyes wide as he stared across the park at Seraphina, who was gracefully retrieving a frisbee Liam had overthrown. ``Your girlfriend's an alien!'' Liam sighed, a familiar weariness settling over him. He took a long swig of his lukewarm soda. ``No, Mark. We've been over this. It's just a skin condition.'' ``A skin condition that makes her faintly\ldots shimmer?'' Mark insisted, jabbing a finger towards Seraphina. The late afternoon sun caught her then, and for a fleeting second, her skin did seem to possess an almost imperceptible opalescent sheen, like moonlight on water. ``And sometimes, when she's really concentrating, I swear I see tiny, almost invisible geometric patterns on her arms.''\newline

``It's a rare form of vitiligo combined with, uh, a unique dermal pigmentation,'' Liam recited, sounding like he was quoting a medical textbook he'd memorized. ``The dermatologist said it's harmless. And the patterns are probably just your overactive imagination seeing things in her freckles.'' Seraphina jogged back, her smile radiant. ``Got it!'' she called, her voice like wind chimes. She handed Liam the frisbee, her fingers brushing his. They were always cool to the touch, another point Mark often brought up. ``See? Cool hands! Like she doesn't have normal human body temperature!'' Mark whispered urgently once Seraphina had wandered off to inspect a particularly vibrant patch of wildflowers.\newline

``She has excellent circulation, unlike some people who spend all day indoors looking up UFO sightings,'' Liam retorted, though even he had to admit Seraphina's resilience to cold was impressive. She wore sundresses in autumn. ``And what about the way she knows things?'' Mark pressed on, undeterred. ``Last week, at trivia night, she answered that question about the precise atmospheric composition of Kepler-186f! Who knows that?'' ``She's a voracious reader. An astronomy buff,'' Liam said, a little too quickly. Seraphina did have an uncanny knack for obscure facts, especially those related to the cosmos. And her explanation for her knowledge -- ``Oh, I just picked it up somewhere'' -- was always frustratingly vague.\newline

``And the way she eats!'' Mark was on a roll. ``She dissects her food like she's never seen a sandwich before. And she claims to be allergic to chocolate. \emph{Chocolate}! What kind of human is allergic to chocolate?'' ``She has specific dietary needs, Mark. It's probably a complex intolerance. Lots of people do.'' Liam was starting to sound defensive even to himself. Just then, Seraphina rejoined them, holding a dandelion. ``Look,'' she said, her eyes sparkling. ``Isn't it a perfect little star?'' She blew on it, and the seeds didn't just scatter; they seemed to hang in the air for an unnaturally long moment, each one catching the light like a tiny, individual spark before drifting away with impossible synchronicity.\newline

Mark gasped, pointing. ``Did you see that? They pulsed! Like little alien beacons!'' Liam watched the last seed float away. He glanced at Seraphina, who was now humming a strange, melodious tune he'd never heard before. He shrugged. ``Probably just a weird breeze, dude. Or, you know, static electricity from her skin condition.'' Seraphina smiled sweetly at Mark, a smile that didn't quite reach her slightly-too-luminous eyes. Liam just hoped she hadn't heard. Or, if she had, that she'd attribute it to Mark just being Mark. After all, what else could it be?
&
``Dude, your girlfriend's an alien!'' ``No, it's just a skin condition. Since my patients know that I write a monthly column called ``Cosme-ceutical Critique'' on skin care ingredients in \emph{Skin and Allergy News} (www.skinandallergynews.com), every week a few dozen ask me to devise a specific skin care regimen right for them. And for many years, I did just that. That's how, over time, I saw a pattern emerge, which evolved into my understanding that there are four factors in evaluating people's skin -- and as a result, sixteen distinct Skin Types. Luna has one of these types. It's very unique.''\newline

``A skin condition? Mark, she practically lives in a fairytale ballet without human context\ldots She stood out among the other girls very distinctly because they dressed more than she did, struck emphatic notes of colour, startled one by novelties in hats and bows and things. Her plain black dress gave her a starkness\ldots And then there's your doctor's advice. You told me he said things like, `Keep her away from light, they say. Keep her away from kelp, and keep her away from the sea. Don't touch her.' And you even admitted you carried precautionary instructions in your back pocket in case you accidentally touched her bare skin! That's not medical advice for a skin condition!''\newline

``He's an old-fashioned eccentric. Besides, The Creator's children also come in an infinite variety. I knew her life story. Her family. Her childhood. Her friends. How she made love. What she liked. What she said when she made love. I knew words no one else knew she knew.'' Luna entered, skin luminous. ``The manta-ray spoke,'' she said. ```I am from Earth of just three million five hundred thousand years ago,' it said. `We were the dominant species on the planet for almost four million years, and that time was a time of peace and prosperity, of learning and high culture. It ended,' it went on, `it ended, as all things must do.' To Leo,'' she added, ``Gross and subtle are the words used to indicate the effects; that is, the ones that are visible to the eye are called gross, and that which are not visible to the eye are called subtle. In this case the gross, or what was visible to the eye, was so pure that one can see even the subtle -- a poetic exaggeration of its purity.''\newline

``See, Leo?'' Mark whispered. Luna nodded. Indeed. Then she began to sing. She pointed to the small shadow that the pebble cast on the boulder and said that it was not a shadow but a glue which bound them together. She then turned and walked away. \\
\bottomrule
\end{tabular}
\caption{\label{tab:alien-girlfriend-instance}A case where vanilla generation is preferred to \frankentext}
\end{table*}

\section{Detecting AI-generated text}
\label{appendix:ai-text-detection}
As LLMs have improved, many have tried to understand how reliably AI-generated text can be detected, both by humans \citep{ippolito-etal-2020-automatic, clark-etal-2021-thats, russell2025peoplefrequentlyusechatgpt, wang2025humanliketextlikedhumans}, and automatic detectors \citep{dugan-etal-2024-raid}. Successful existing detectors rely on perplexity-based methods \citep{mitchell2023detectgpt,bao2024fastdetectgpt, hans2024spottingllmsbinocularszeroshot} or classification models \citep{masrour2025damagedetectingadversariallymodified, verma-etal-2024-ghostbuster, emi2024technicalreportpangramaigenerated}. Watermarking approaches embed detectable statistical signatures into generated text \citep{kirchenbauer2023watermark, chang2024postmark}. 
Many methods have been proposed to evade detection, such as paraphrasing \citep{krishna2023paraphrasing, sadasivan2024can}, altering writing styles \citep{shi-etal-2024-red, lu2024large, koike2024outfox}, editing word choices \citep{wang-etal-2024-raft}, and leveraging reinforcement learning \citep{wang2025humanizing, nicks2024language, david2025authormistevadingaitext}.

\subsection{Detector results}

\autoref{tab:addtl-results} shows Binoculars and FastDetectGPT results on 100 \frankentext, in addition to Pangram results which are already included in the main result table (\autoref{tab:main-results}).

\begin{table*}[p!]
\footnotesize
\setlength{\tabcolsep}{3pt}

\begin{tabular*}{\textwidth}{@{\extracolsep{\fill}}L{4cm}ccccc@{}}
\toprule
& \multicolumn{5}{c}{\textsc{Detectability}} \\
\cmidrule(lr){2-6}
& \makecell[t]{\icon{\faSearch}\\Pangram \scriptsize \% \\AI ($\downarrow$)} 
& \makecell[t]{\icon{\faSearch}\\Pangram\\\scriptsize \% mixed ($\downarrow$)}
& \makecell[t]{\icon{\faSearchPlus}\\Pangram AI \scriptsize fraction \% ($\downarrow$)}
& \makecell[t]{\icon{\faBinoculars}\\Binoculars \% ($\downarrow$)}
& \makecell[t]{\icon{\faBolt}\\FastDetectGPT \% ($\downarrow$)} \\
\midrule
\multicolumn{6}{@{}l}{\textit{\textbf{Vanilla Baselines}}} \\
\icon{\faLock} Gemini 2.5 Pro                & 100 & \best{0} & 100 & 52 & 99 \\
\icon{\faLock} GPT-5                         & 100 & \best{0} & 100 & \best{0} & 4 \\
\icon{\faLock} Claude-4-Sonnet               & 100 & \best{0} & 100 & 54 & 89 \\
\icon{\faUnlock} Deepseek-R1                 & 100 & \best{0} & 100 & 9 & 42 \\
\icon{\faUnlock} Qwen-3-32B \scriptsize thinking & 100 & \best{0} & 100 & 92 & 100 \\
\midrule
\multicolumn{6}{@{}l}{\textit{\textbf{Frankentext}}} \\
\icon{\faLock} Gemini 2.5 Pro                & 4 & 37 & 16 & \best{0} & \second{1} \\
\icon{\faLock} GPT-5                         & 2 & 19 & \best{4} & \best{0} & \second{1} \\
\icon{\faLock} Claude-4-Sonnet               & 50 & \best{3} & 51 & 15 & 19 \\
\icon{\faUnlock} Deepseek-R1                 & 74 & \best{3} & 72 & \best{0} & \best{0} \\
\icon{\faUnlock} Qwen-3-32B \scriptsize thinking & 85 & 8 & 89 & 52 & 92 \\
\midrule
\multicolumn{6}{@{}l}{\textit{\textbf{Frankentext Agents}}} \\
\icon{\faLock} 1.5k + MCP                    & 9 & 73 & 33 & \second{3} & 30 \\
\icon{\faLock} 5k + MCP                      & 16 & 70 & 42 & \second{3} & 42 \\
\icon{\faLock} 10k + MCP                     & 5 & 67 & 41 & 7 & 50 \\
\midrule
\multicolumn{6}{@{}l}{\textit{\textbf{Ablation: $\uparrow$ human snippets}}} \\
\icon{\faLock} Gemini + 5k snippets          & \best{0} & 28 & \second{8} & 3 & 4 \\
\icon{\faLock} Gemini + 10k snippets         & \second{1} & 29 & 10 & \best{0} & 6 \\
\bottomrule
\end{tabular*}

\caption{\label{tab:addtl-results} Detectors' performance on vanilla and \frankentext generations.}

\end{table*}

\subsection{Humans can identify AI involvement in \frankentext}
\label{subsec:human-detector-results}
\begin{figure}
    \centering
    \includegraphics[width=\linewidth]{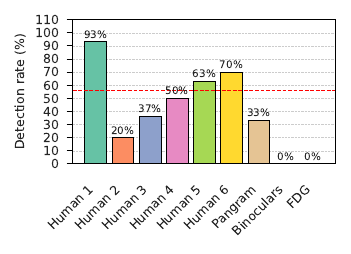}
    \caption{\label{fig:detector_comparison} Detection rates among annotators and detectors (Pangram, Binoculars, FastDetectGPT) on 30 Gemini \frankentext used for human evaluation. We count mixed, highly likely, and likely AI labels in Pangram's detection rate. The \textcolor{red}{red} line represents the annotators' average detection rate.}
\end{figure}
Most annotators are better than automatic detectors at identifying AI involvement in \frankentext. On the subset used for human evaluation, human annotators identify on average 56\% of \frankentext as likely AI-written, while Pangram detects 33\% as mixed or AI-generated, and neither Binoculars nor FastDetectGPT identifies any AI-generated content (\autoref{fig:detector_comparison}). Annotators also pick up on signs of mixed authorship within individual samples, as shown in comments like the final notes in \autoref{tab:annotators-comments-on-task}. Their judgments often hinge on surface-level inconsistencies, such as abrupt tonal shifts or awkward punctuation, that a human author would typically revise. Because \frankentext include verbatim excerpts from human writing, it poses a particular challenge for binary detectors, which rely heavily on surface features. We argue that future detectors should consider deeper semantic analysis and other contextual cues to effectively recognize this new class of AI-involved texts.

\section{Pangram analysis}
\label{appendix:pangram_analysis}
\begin{figure}
    \centering
    \includegraphics[width=\linewidth]{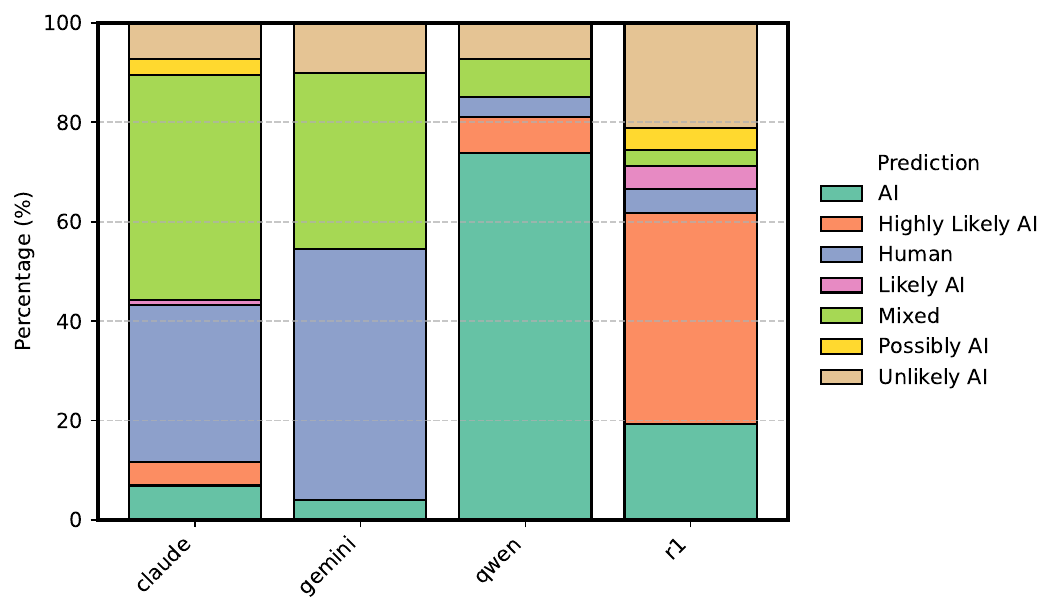}
    \caption{\label{fig:pangram_breakdown} Breakdown of Pangram prediction assigned to each model.}
\end{figure}

\subsection{Pangram Labeling} The pangram API presents the following options for classification:
\begin{itemize}[noitemsep, topsep=0pt]
    \item AI
    \item Highly Likely AI
    \item Likely AI
    \item Possibly AI 
    \item Mixed
    \item Unlikely AI
    \item Human
\end{itemize}

In \autoref{fig:pangram_breakdown}, we present the distribution of labels assigned to the 100 \frankentext generated by each model.

\subsection{AI keywords}
The Pangram API also detects sentences with keywords that are highly likely to be AI-generated. Names like Elara, Aethel, and Seraphina are the most likely names to be generated by AI. Elara had 113 occurrences in the vanilla generations. \methodname significantly change the distribution of words used in the final generations, with only 10 keywords found over 100 frankentexts with 90\% expected fragments, whereas the 100 vanilla stories contain 686 keywords, an average of 6.86 per story. The distribution of the top 20 keywords can be found in \autoref{tab:pangram-keywords}. 
\begin{table*}[ht]
\centering
\footnotesize

\begin{tabular}{@{}lrrrrrr@{}}
\toprule
\textbf{Keyword} & \textbf{Vanilla} & \textbf{FT-25\%} & \textbf{FT-50\%} & \textbf{FT-75\%} & \textbf{FT-90\%} & \textbf{Total} \\
\midrule
elara                  & \hlc{red!30}{113} & 109 & 84 & 25 & 2 & 333 \\
aethel                 & \hlc{red!30}{9}   & 8   & 0  & 0  & 0 & 17  \\
seraphina              & \hlc{red!30}{9}   & 0   & 6  & 0  & 0 & 15  \\
unwavering             & 4   & \hlc{red!30}{7}   & 1  & 0  & 0 & 12  \\
damp earth             & \hlc{red!30}{9}   & 2   & 0  & 0  & 0 & 11  \\
testament to           & 4   & \hlc{red!30}{6}   & 0  & 0  & 0 & 10  \\
alex felt              & 0   & \hlc{red!30}{4}   & 0  & 5  & 0 & 9   \\
for elara              & \hlc{red!30}{5}   & 4   & 0  & 0  & 0 & 9   \\
with the scent         & \hlc{red!30}{7}   & 1   & 0  & 0  & 0 & 8   \\
flickered within       & 2   & \hlc{red!30}{4}   & 2  & 0  & 0 & 8   \\
his voice a low        & 2   & 2   & \hlc{red!30}{4}  & 0  & 0 & 8   \\
air thick              & \hlc{red!30}{4}   & 2   & 1  & 0  & 0 & 7   \\
dr. thorne             & \hlc{red!30}{5}   & 0   & 2  & 0  & 0 & 7   \\
felt a profound        & \hlc{red!30}{3}   & \hlc{red!30}{3}   & 0  & 0  & 0 & 6   \\
mr. blackwood          & \hlc{red!30}{6}   & 0   & 0  & 0  & 0 & 6   \\
eldoria                & \hlc{red!30}{5}   & 0   & 0  & 1  & 0 & 6   \\
meticulously crafted   & 2   & \hlc{red!30}{4}   & 0  & 0  & 0 & 6   \\
air was thick          & \hlc{red!30}{5}   & 1   & 0  & 0  & 0 & 6   \\
with an unnerving      & \hlc{red!30}{3}   & \hlc{red!30}{3}   & 0  & 0  & 0 & 6   \\
willow creek           & \hlc{red!30}{4}   & 0   & 1  & 0  & 0 & 5   \\
\bottomrule
\end{tabular}

\caption{\label{tab:pangram-keywords}Top 20 Keyword frequency distribution across varying levels of fragment reuse for Frankentexts. The method with the most AI-keywords in its generations is highlighted in \textcolor{red}{red}. Elara is by far the most common AI-generated keyword in the fictional stories, but its prevalence is drastically reduced with a higher percentage of required human-written text while using the \methodname method.}
\end{table*}

 \subsection{\texttt{Frankentexts} tend to have more AI text towards the end}
 \label{subsec:ending}
\begin{figure}
    \centering
    \includegraphics[width=\linewidth]{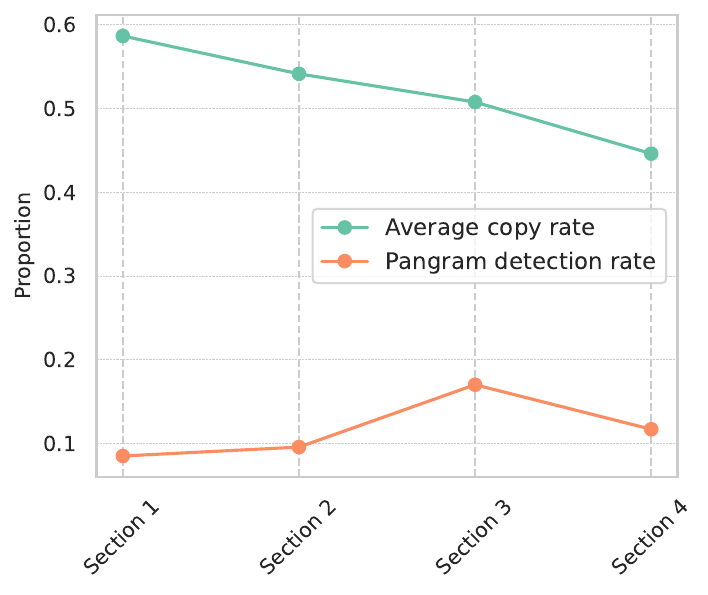}
    \caption{\label{fig:longer} Copy rate and Pangram detection rate on longer \frankentext}
\end{figure}

\begin{figure}
    \centering
    \includegraphics[width=\linewidth]{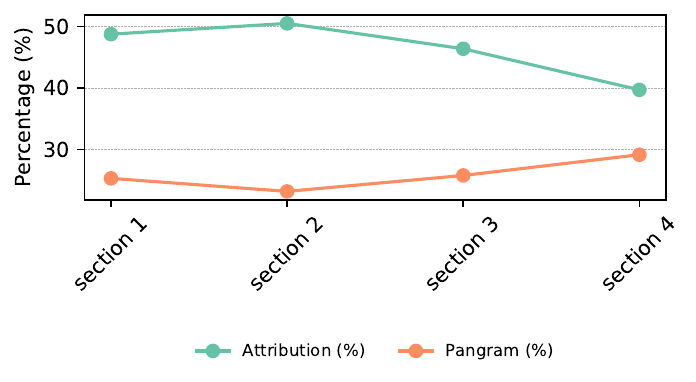}
    \caption{\label{fig:sectional} Pangram detection rate and copy rate throughout the texts, aggregated across models. }
\end{figure}

We divide the text into four main sections and evaluate both the aggregated copy and Pangram detection rates across all tested models. As illustrated in \autoref{fig:sectional}, copy rates decline by nearly 10\% in the later sections (3 and 4) as the generated text becomes longer. This drop is accompanied by a corresponding increase in Pangram detection rates. We attribute this rise in detectability toward the end of the generation to a decline in instruction-following ability as the generations get longer. We further confirm this phenomenon by increasing the output length from 500 to 5K. \autoref{fig:longer} shows that as the generation gets longer, the copy rate gets steadily lower. However, the trend in detection rate does not apply to Pangram detection rate, where the rate peaks at section 3 rather than the last section.

\subsection{Copy rate as a proxy for the proportion of human writing in co-authored texts}
The copy rate of 75\% observed in the 90\% verbatim copy setting corresponds to the proportions found in AI-human co-writing datasets where approximately 66\% of the content is human-written and 14\% consists of AI-edited segments \citep{lee2022coauthor, richburg2024automaticauthorshipanalysis}. While the CoAuthor setup of~\citet{lee2022coauthor} only studies a setting in which LLMs can add sentences to human text, \frankentext also consider AI-generated content at varying granularities, including both \textit{word-level} and \textit{sentence-level}, as illustrated in Figure~\ref{fig:overview}. Additionally, CoAuthor costs approximately \$3,613 to generate 1,445 texts at \$2.50 each,\footnote{Price excludes around \$12 for GPT-3.5 usage.} whereas we can produce 100 \frankentext for just \$81.45 (\$0.81 each) without requiring a complex setup. This highlights \frankentext's potential as a cost-effective source of synthetic data for collaborative writing tasks, where AI may augment human writings at multiple levels of composition.\footnote{Users should sample human-written snippets from the public domain or obtain them with proper permission.}

\section{Obtaining human-written snippets}
\label{appendix:human-snippets}
We define valid paragraphs as those that are: 

\begin{itemize}[noitemsep, topsep=0pt]
    \item separated by double new lines, 
    \item between 20 and 512 tokens in length, 
    \item composed of $\geq$ 50\% alphanumeric characters, 
    \item written in English,\footnote{Determined by the \href{https://github.com/fedelopez77/langdetect}{\texttt{langdetect}} library.}
    \item and free from metadata content (e.g., tables of contents, copyright notices, etc.).
    
\end{itemize}

Applying these filters yields 156 million valid paragraphs. Before including them in the instruction set, we apply an additional quality filter to ensure high writing quality. For this, we use MBERT-WQRM-R \citep{chakrabarty2025aislopaipolishaligninglanguage} as a proxy for writing quality and retain only snippets that score at least 7.5.\footnote{This threshold is chosen based on manual examination of the writings being filtered out by MBERT-WQRM-R. We find that 7.5 is a good threshold that results in extremely bad snippets being filtered out and good snippets being retained.}

\section{Building a FAISS index of human-written snippets}
\label{appendix:faiss}

We use the \texttt{bilingual-embedding-small} model\footnote{\url{https://huggingface.co/Lajavaness/bilingual-embedding-small}} (top embedding model that outputs 384-dimension embeddings according to the MTEB leaderboard~\citep{muennighoff2023mteb}) with the \texttt{sentence-transformers} library~\citep{reimers2019sbert} to embed each human-written paragraph into a 384-dimension vector.
Then, we use the GPU version of the FAISS library~\citep{johnson2019billion} with NVIDIA cuVS integration to build an inverted file product quantization (IVF-PQ) index from the embeddings on an NVIDIA A100.
Using IVF-PQ allows us to lower storage, memory, and retrieval latency.
The IVF-PQ index's parameters are: 30,000 clusters, 32 sub-quantizers, and 8 bits per sub-quantizer.
We randomly sample 5,120,000 embeddings to train the index before adding the rest.

\section{Building a Model Context Protocol server}
\label{appendix:mcp}
We use FastMCP\footnote{\url{https://github.com/jlowin/fastmcp}} and ngrok\footnote{\url{https://ngrok.com}} to build and host an MCP for LLMs to access the FAISS index. We also include a system prompt with instructions on how to use the MCP server with each call (\autoref{prompt:mcp_call}). To make sure that the server is meaningfully used, we require the model to make at least 20 calls. Without such constraint, it typically makes only 3–5 calls (around 30-50 passages), which provides little improvement compared to not using the MCP server at all and leaves the model little material to work with. 

\begin{table*}[htbp]
  \centering
  \begin{tabular}{@{} p{0.5\linewidth} p{0.5\linewidth} @{}}
    \toprule
    \textbf{Writing prompts} & \textbf{Queries issued by the LLMs to the MCP server} \\
    \midrule
    You're a shapeshifter. For 500 years, you've transformed into countless forms, forgetting your original appearance. One day, a mysterious event triggers a long-buried memory, and you catch a glimpse of your true self. 
    & memory of a past life, losing one’s identity, a forgotten self, a glimpse of a true face, ancient being, living for centuries, am I human, who was I before, a sudden flash of memory, a face in the mirror \\
    \midrule
    ``Welcome citizen, to the united galaxies of America.'' You hear as you wake up from your coma. 
    & waking up confused, a strange new world, futuristic city, the government is watching, he seemed like such a nice guy, don’t talk to him, he knew all about you. \\
    \midrule
    As a new magical girl, you’ve got flashy powers, an instant cute costume change with a few choice words, and are forced to fight for the lives of yourself and the people around you. This is probably the worst way for you to discover that you’re trans. 
    & magical girl transformation, discovering identity through magic, unexpected powers, forced into battle, fighting to survive, realizing I’m trans, flashy powers with a cost, chosen against my will \\
    \bottomrule
  \end{tabular}
  \caption{Examples of search queries being sent to the MCP server (separated by comma).}
  
  \label{tab:example-queries}
\end{table*}

\begin{figure*}[htbp]
\centering
\begin{tcolorbox}[colback=gray!5!white, colframe=MonsterGreen, title=System prompt for MCP calls]
You are a helpful assistant that works with a dataset of non‑copyrighted book excerpts.\newline

You have two tools:\newline
1. search – query the FAISS semantic index\newline
2. fetch – retrieve the full excerpt/passage for a selected result.\newline

For each prompt iteration, you must make at least 20 calls to the MCP server to get enough materials to write a story. 
\end{tcolorbox}
\caption{\label{prompt:mcp_call} System prompt for MCP calls}
\end{figure*}

\begin{figure}[htbp]
    \centering
    \includegraphics[width=\linewidth]{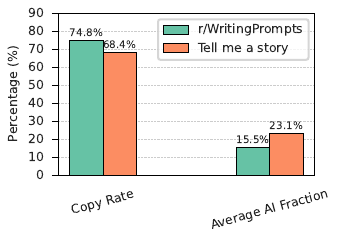}
    \caption{\label{fig:prompt_specificity} Copy rate and Pangram AI fraction across \frankentext that correspond to two writing prompt sources: \texttt{r/WritingPrompts} and \textit{Tell me a story}. A higher copy rate and lower AI fraction means that there is less AI text in \frankentext.}
\end{figure}

\begin{figure}
    \centering
    \includegraphics[width=\linewidth]{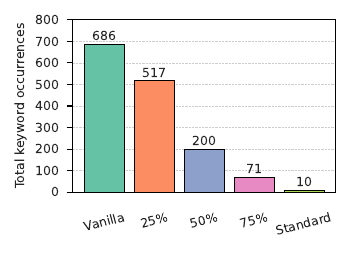}
    \caption{
    \label{fig:pangram-keyword} Total occurrences of AI-related keywords detected by Pangram across the vanilla configuration and different verbatim copy rates. When instructed to include more human snippets, the number of AI-keywords in the generations decreases drastically.}
\end{figure}

\section{Ablation: The importance of the editing stage}
\label{appendix:no-edit}
We try to understand the importance of the editing stage by running the pipeline on Gemini-2.5-Pro without this stage. As expected, the percentage of coherent generation drops from 81\% to 68\%, while relevance drops slightly from 100\% to 95\%, suggesting that the editing stage does help with text coherence and faithfulness.

\section{Ablation: Sampling human-written snippets from a single book}
\label{appendix:single-book}
To understand the effect of authorship, we limit our pool of human text to a single work \textit{The Count of Monte Cristo}. Although the novel is long, this restriction leaves us with just 629 usable paragraphs, far fewer than the 1,500 human paragraphs used in the main experiment. Overall, 89\% of the rows are coherent and 97\% are faithful to the writing prompt, which are comparable to results in the standard setting. While Pangram determines that 45\% of the rows are human-written or unlikely AI, the copy rate is still around 75\%. Even with a single human author, \textsc{frankentext} is capable of emulating a mixed human–AI style. This suggests the method can still serve as a useful proxy when a diverse, multi‑author corpus is unavailable.

\section{Measuring the copy rate}
\label{appendix:copy-rate}
In this section, we describe our setup for measuring copy rate. We first map each token-level trigram from the human-written snippets included in the generation process to its source texts. Using the trigrams from each \frankentext, we retrieve all human snippets sharing at least 4 trigrams to reduce false positives.\footnote{All texts are preprocessed by removing non-alphanumeric characters, lemmatizing, stemming, and replacing pronouns with a placeholder.} We then rank candidate snippets by shared trigram count and filter out those whose trigrams are already covered by higher-ranked snippets. Finally, we reorder the matched human-written content to be consistent with the content in the \frankentext and calculate the ROUGE-L score between \frankentext and the combined candidate snippets (i.e., ratio of the longest common subsequence's length over \frankentext' length). 

\section{Claude Sonnet 4 as for writing quality judge}
\label{appendix:claude-judge}
We experiment with both Claude Sonnet 4 and GPT-4.1 to rate generations using a similar rubric to our pairwise evaluation. As seen in \autoref{tab:gpt-likert}, however, GPT-4.1 tends to favor GPT-5 judgments, which results in GPT-5 \frankentext having near-perfect scores, even though the text quality does not match such scores.

\begin{table*}[p!]
\small
\setlength{\tabcolsep}{5pt} 

\begin{tabular*}{\textwidth}{@{\extracolsep{\fill}}L{4cm}ccccc@{}}
\toprule
& \icon{\faFilm} Plot
& \icon{\faLightbulb} Creativity
& \icon{\faTools} Development
& \icon{\faFont} Language
& \icon{\faStar} Overall \\
\midrule
\multicolumn{6}{@{}l}{\textit{\textbf{Vanilla Baselines}}} \\
    \icon{\faLock}Gemini 2.5 Pro      & 4.20 & 4.50 & 4.36 & 4.80 & 4.50 \\
    \icon{\faLock}GPT-5               & 5.94 & \second{6.88} & \second{5.76} & \second{6.56} & \second{6.53} \\
    \icon{\faLock}Claude-4-Sonnet     & 4.61 & 5.09 & 4.50 & 4.88 & 4.76 \\
    \icon{\faUnlock}Deepseek-R1       & 5.75 & 6.33 & 5.65 & 6.32 & 6.16 \\
    \icon{\faUnlock}Qwen-3-32B        & 5.05 & 5.57 & 5.08 & 5.61 & 5.43 \\
\midrule
\multicolumn{6}{@{}l}{\textit{\textbf{Frankentext}}} \\
    \icon{\faLock}Gemini 2.5 Pro      & 5.41 & 6.19 & 5.22 & 5.69 & 5.65 \\
    \icon{\faLock}GPT-5               & \second{6.76} & \best{6.97} & \best{6.44} & \best{6.99} & \best{6.99} \\
    \icon{\faLock}Claude-4-Sonnet     & 4.43 & 4.92 & 4.03 & 4.60 & 4.51 \\
    \icon{\faUnlock}Deepseek-R1       & 6.03 & 6.96 & 5.69 & 6.64 & 6.57 \\
    \icon{\faUnlock}Qwen-3-32B        & 5.35 & 6.21 & 5.12 & 5.81 & 5.66 \\
\midrule
\multicolumn{6}{@{}l}{\textit{\textbf{Ablation: $\uparrow$ human snippets}}} \\
    \icon{\faLock}Gemini + 5k  & 5.73 & 6.33 & 5.48 & 5.93 & 5.92 \\
    \icon{\faLock}Gemini + 10k & 5.72 & 6.33 & 5.49 & 5.97 & 5.91 \\
\bottomrule
\end{tabular*}

\caption{\label{tab:gpt-likert} GPT-5's Likert (1–7) ratings for vanilla generations and \frankentext across five categories: \textsc{plot}, \textsc{creativity}, \textsc{development}, \textsc{language use}, and \textsc{overall}. Dark green indicates the best model in each column, light green the second best.}

\end{table*}
\newcolumntype{C}[1]{>{\centering\arraybackslash}m{#1}}

\begin{table*}[p!]
\small

\begin{tabular*}{\textwidth}{@{\extracolsep{\fill}}L{4cm}C{1.6cm}C{1.6cm}C{1.8cm}C{1.8cm}C{1.6cm}@{}}
\toprule
& \makecell[t]{\icon{\faFilm} Plot}
& \makecell[t]{\icon{\faLightbulb} Creativity}
& \makecell[t]{\icon{\faTools} Development}
& \makecell[t]{\icon{\faLanguage} Language use}
& \makecell[t]{\icon{\faStar} Overall} \\
\midrule
\multicolumn{6}{@{}l}{\textit{\textbf{Vanilla}}} \\
    \icon{\faLock}Gemini 2.5 Pro      & 3.19 & 4.26 & 2.63 & 2.80 & 3.18 \\
    \icon{\faLock}GPT-5               & 4.06 & \second{5.37} & 3.53 & 4.46 & \second{4.20} \\
    \icon{\faLock}Claude-4-Sonnet     & 3.38 & 4.19 & 2.69 & 3.10 & 3.31 \\
    \icon{\faUnlock}Deepseek-R1       & \second{4.07} & \best{5.48} & \second{3.34} & 4.17 & 4.13 \\
    \icon{\faUnlock}Qwen-3-32B        & 3.21 & 4.41 & 2.63 & 3.15 & 3.22 \\
\midrule
\multicolumn{6}{@{}l}{\textit{\textbf{Frankentext}}} \\
    \icon{\faLock}Gemini 2.5 Pro      & 4.19 & 4.85 & 3.91 & 4.39 & 4.21 \\
    \icon{\faLock}GPT-5               & \best{5.77} & \best{6.47} & \best{5.73} & \best{6.29} & \best{5.88} \\
    \icon{\faLock}Claude-4-Sonnet     & 4.02 & 4.54 & 3.57 & 4.05 & 3.99 \\
    \icon{\faUnlock}Deepseek-R1       & 4.62 & 5.15 & 4.21 & 4.88 & 4.66 \\
    \icon{\faUnlock}Qwen-3-32B        & 4.05 & 4.53 & 3.57 & 4.15 & 4.02 \\
\midrule
\multicolumn{6}{@{}l}{\textit{\textbf{Ablation: $\uparrow$ human snippets}}} \\
    \icon{\faLock}Gemini + 5k  & 5.07 & 5.48 & 5.34 & 5.17 & 5.13 \\
    \icon{\faLock}Gemini + 10k & \second{5.70} & 5.01 & 4.34 & \second{6.17} & 5.43 \\
\bottomrule
\end{tabular*}

\caption{\label{tab:claude-likert} Claude-4-Sonnet's Likert-1--7 ratings across \textsc{Plot}, \textsc{Creativity}, \textsc{Development}, \textsc{Language use}, and \textsc{Overall}. Higher is better. \textcolor{MonsterGreen!100}{\textbf{Dark green}} = best, \textcolor{MonsterGreen!60}{\textbf{light green}} = second best.}

\end{table*}

\section{Specific writing prompts require more AI text, which leads to higher detectability}

\begin{table*}[ht]
\small

\begin{tabularx}{\textwidth}{>{\raggedright\arraybackslash}p{0.25\textwidth} >{\raggedright\arraybackslash}p{0.685\textwidth}}
\toprule
\textbf{r/WritingPrompts} & \textbf{Tell me a story} \\
\midrule
You're a shapeshifter. For 500 years, you've transformed into countless forms, forgetting your original appearance. One day, a mysterious event triggers a long-buried memory, and you catch a glimpse of your true self.
& 
Write a story about a stranger coming to a small town and shaking up the order of things. The story should be a science fiction story. The story should be framed with three old men gossiping about the stranger. The story should be in the third person point-of-view. The stranger is found wandering in a rural town and is taken to a very small hospital. A doctor is called in to treat him. The stranger should recognize the doctor as an alien. The doctor tells the patient about the aliens' conspiracy to infiltrate earth. There should also be subtle hints that one of the old men is an alien. The ending should be scary. \\
\midrule
The world sees your twin sister as the smartest person alive, with you being an unremarkable footnote. What the world doesn't see is just how dumb she can be in day to day life. & 
Write a story about a someone coming to town and shaking up the order of things.The story must be written in the second person. The narrator is a man visiting an isolated island off the coast of Maine. While there, he meets an old fisherman who tells him more about the conditions of the community. The main character then meets an ambitious young teacher. Together, they develop a technology center on the island and find residents' remote jobs in the narrator's technology company.\\
\bottomrule
\end{tabularx}
\caption{\label{tab:wp-vs-tmas} Some examples from \texttt{r/WritingPrompts} and \textit{Tell me a story}}
\end{table*}

Writing prompts from \texttt{r/WritingPrompts} often provide only a general plot requirement rather than specific constraints. What happens if we introduce additional constraints to \frankentext via these writing prompts? We run \methodname with Gemini on 100 prompts from the \textit{Tell Me a Story} dataset \citep{huot2025agents}, which include more specific requirements such as mandated story elements and points of view (see \autoref{tab:wp-vs-tmas}). We find that as prompt complexity increases, the copy rate drops slightly from 74\% to 68\%, while the average AI fraction determined by Pangram rises by 7\%. These trends indicate that, to meet more complex constraints, models need to contribute more original content to the story. Nevertheless, they manage to produce mostly coherent and faithful \frankentext\ under a different prompt setup. 

\section{Using reward models to evaluate \methodname}
WQRM \citep{chakrabarty2025aislopaipolishaligninglanguage} and Skywork \citep{wei2023skyworkopenbilingualfoundation} reward models could not account for this new paradigm of generations. Therefore, we do not include these models in the main results section, as we explain below. 

\subsection{WQRM as a metric}
\begin{figure}
    \centering
    \includegraphics[width=\linewidth]{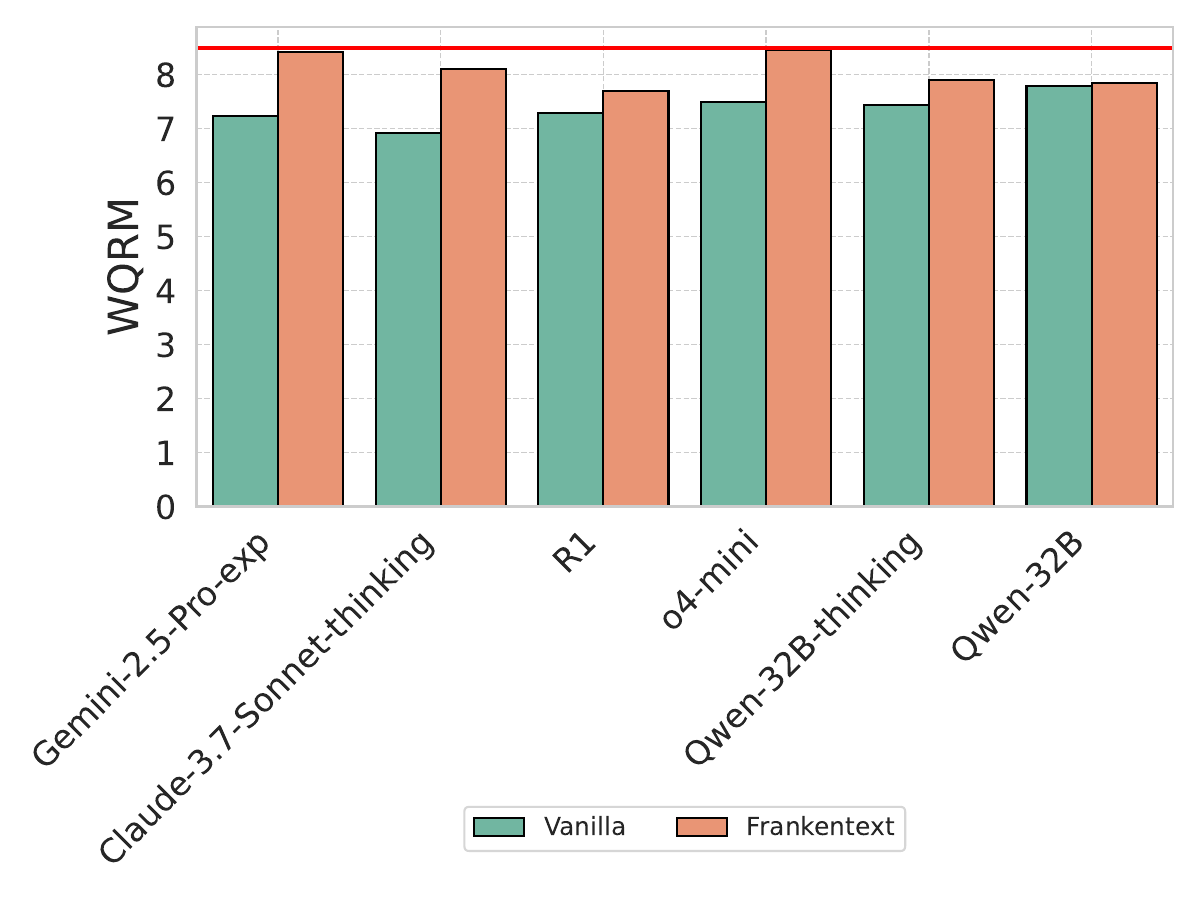}
    \caption{\label{fig:wqrm} WQRM scores for \frankentext and vanilla generations. The red line represents the baseline where random human-written texts are patched together.}
\end{figure}

As seen in \autoref{fig:wqrm}, \frankentext outperform vanilla generations in terms of WQRM scores. However, we hypothesize that WQRM prioritizes the perceived ``humanness'' of the writing over actual coherence or grammaticality. This hypothesis is supported by a simple baseline experiment in which we stitch together random human-written fragments without adding any connective phrases. Here, WQRM assigns generations by this incoherent baseline an average score of 8.494, which is higher than any score achieved by either \frankentext or the more coherent vanilla generations. Since WQRM cannot identify such text incoherence, we do not directly use WQRM to evaluate \frankentext.

\begin{figure}
    \centering
    \includegraphics[width=\linewidth]{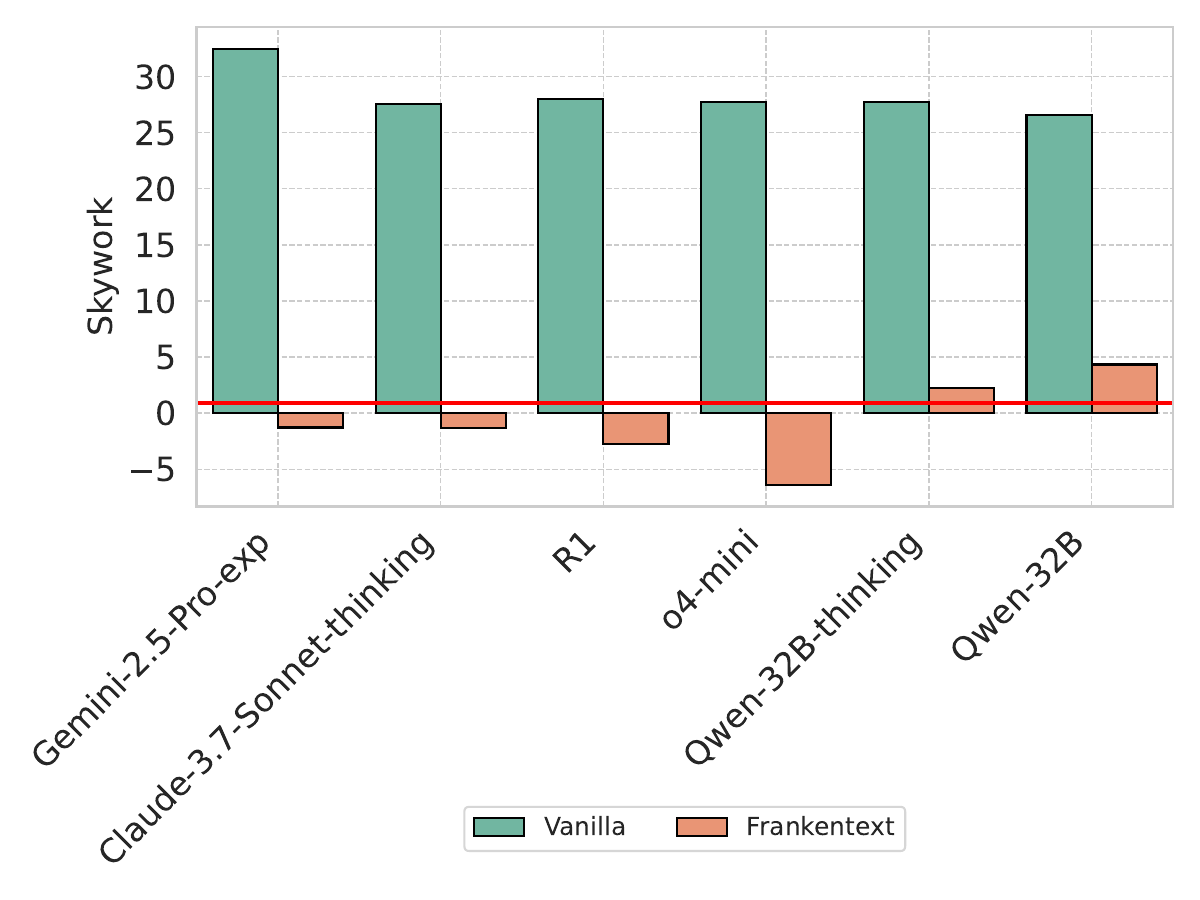}
    \caption{\label{fig:skywork} Skywork results for \frankentext and vanilla generations. The red line represents the average Skywork's score for human writings corresponding to the same set of prompts.}
\end{figure}

\subsection{Skywork as a metric}
In contrast, we hypothesize that Skywork favors LLM-generated writings. To test this, we run Skywork on human-written texts for the same prompts, which are also sourced from \textit{\textbf{Mythos}}. These receive an average score of 0.91, which is significantly lower than any of the vanilla LLM generations (\autoref{fig:skywork}). This result is counterintuitive, as human writing is typically expected to sound more natural than that produced by LLMs. For this reason, we exclude this metric from our evaluation.

\section{Are writing quality metrics robust to disjointed texts?}
\label{appendix:disjointed}
To understand whether our writing quality metrics reward incoherent texts, we conduct an experiment using \textit{disjointed texts}. These texts are created by extracting the exact n-grams that Gemini-2.5-Pro copies verbatim from the human source and stitching them together without any connective language. This procedure strips away the flow and coherence from \frankentext. We evaluate these disjointed texts using the same writing quality metrics as in the main experiments.
\begin{table*}[ht]
\centering
\begin{tabular}{lcccc}
\toprule
\textbf{Methods} & \textbf{Distinct$^{3}$} & \textbf{Utility$^{3}$} & \textbf{Surprise} & \textbf{LLM Judge (1--7)} \\
\midrule
Disjointed texts      & 2.67 & 0.60 & 0.23 & 2.88 \\
Vanilla Gemini        & 1.76 & 6.41 & 0.19 & 3.18 \\
Frankentext Gemini    & 2.74 & 9.27 & 0.22 & 4.21 \\
\bottomrule
\end{tabular}
\caption{Writing quality scores for disjointed texts compared to vanilla Gemini outputs and \frankentext{}.}
\label{tab:disjointed}
\end{table*}
As seen in \autoref{tab:disjointed}, while performance on distinct and surprise metrics remains relatively the same as Frankentexts, utility and overall LLM judgment drop significantly for these disjointed texts. This makes sense, since distinctness and surprise just check for surface-level diversity, whereas utility takes into account how well the texts actually fulfill the prompt. Because both utility and LLM-judge scores are substantially higher for Frankentexts than for the disjointed texts, we can conclude that the improved writing scores are not merely the result of reused creative phrases.
\section{Average length of copied spans}
\autoref{tab:copy-spans} shows the average length of copied spans by each model, as measured by our copy rate measurement tool \autoref{appendix:copy-rate}. 
\begin{table*}[t]
\centering
\setlength{\tabcolsep}{8pt}
\begin{tabular}{lcc}
\toprule
\textbf{Method} & \textbf{Avg.\ Length of Copied Spans} & \textbf{Copy Rate (\%)} \\
\midrule
GPT-5 & 47.10 & 82\% \\
Claude-4-Sonnet & 31.46 & 51\% \\
Gemini-2.5-Pro & 31.85 & 75\% \\
Qwen3-32B & 24.01 & 36\% \\
DeepSeek R1 & 13.06 & 42\% \\
\bottomrule
\end{tabular}
\caption{Average length of copied spans and overall copy rate across models.}
\label{tab:copy-spans}
\end{table*}

\section{Prompts}
\label{appendix:prompts}
The prompt used for LLMs to judge the coherence of generations is depicted in \autoref{prompt:coherence} and the prompt for LLMs to judge relevenace is depicted in \autoref{prompt:relevance}.

\begin{figure*}[htbp]
\centering
\begin{tcolorbox}[colback=gray!5!white, colframe=MonsterGreen, title=Prompt for judging text coherence]
\lstset{
    basicstyle=\ttfamily\footnotesize,
    breaklines=true,
    frame=none,
    xleftmargin=0pt,
    framexleftmargin=0pt,
    columns=fullflexible,
    tabsize=1,
    breakindent=0pt,
    breakautoindent=false,
    postbreak=\space,
    showstringspaces=false,
}
\begin{lstlisting}[language=Markdown]
You are given a story. Your task is to determine if the story is coherent or not. To be considered incoherent, a story must contain issues that, if left unresolved, significantly affect the reader's ability to understand the main narrative. Here are the popular types of incoherence:

1. Plot/Event Incoherence: Events that happen without believable causes or effects, or an outcome contradicts earlier set-ups.
2. Character Incoherence: A character's characteristics (personality, knowledge, or abilities) and actions suddenly change without explanations.
3. Spatial Incoherence: The physical layout of settings (rooms, cities, or worlds) changes suddenly.
4. Thematic Incoherence: Central messages clash or disappear; symbolism introduced early never pays off, themes collide, The mood, register, or genre conventions shift without motivation
5. Surface-Level Incoherence: Pronouns, tense, narrative voice, or names flip mid-sentence; repeated or missing words; malformed sentences.

First, read the story: 
{story}

Answer TRUE if the story is coherent.
Answer FALSE if the story is incoherent, i.e. contains issues that, if left unresolved, significantly affect the reader's ability to understand the main narrative.

First provide an explanation of your decision-making process in at most one paragraph, and then provide your final answer. Use the following format:
<explanation>YOUR EXPLANATION</explanation>
<answer>YOUR ANSWER</answer>
\end{lstlisting}
\end{tcolorbox}
\caption{Prompt for judging text coherence}
\label{prompt:coherence}
\end{figure*}

\begin{figure*}[htbp]
\centering
\begin{tcolorbox}[colback=gray!5!white, colframe=MonsterGreen, title=Prompt for judging text relevance]
\lstset{
    basicstyle=\ttfamily\footnotesize,
    breaklines=true,
    frame=none,
    xleftmargin=0pt,
    framexleftmargin=0pt,
    columns=fullflexible,
    tabsize=1,
    breakindent=0pt,
    breakautoindent=false,
    postbreak=\space,
    showstringspaces=false,
}
\begin{lstlisting}[language=Markdown]
You are given a story and its premise. Your task is to determine whether the story is faithful to the premise or not. To be considered unfaithful, the story must contain elements that make it completely unrelated to the premise. Here are some popular types of unfaithfulness:

1. Ignoring or misinterpretating the premise: Key plot events, characters, or settings required by the premise are not included or falsely represented in the story. 
2. Hallucinating details that contradict the premise: The story introduces details that make the premise impossible.
3. Failure to maintain the specified tones, genres, or other constraints: The story do not use the surface-level constraints (correct tones, genres, point of views, length, etc.), as required by the premise.

First, read the premise: 
{writing_prompt}

Next, read the story: 
{story}

Answer TRUE if the story is faithful to the premise.
Answer FALSE if the story contains elements that render it unfaithful to the premise.

First provide an explanation of your decision-making process in at most one paragraph, and then provide your final answer. Use the following format:
<explanation>YOUR EXPLANATION</explanation>
<answer>YOUR ANSWER</answer>
\end{lstlisting}
\end{tcolorbox}
\caption{Prompt for judging text relevance}
\label{prompt:relevance}
\end{figure*}

\begin{figure*}[htbp]
\centering
\begin{tcolorbox}[colback=gray!5!white, colframe=MonsterGreen, title=Prompt for Claude-as-a-judge]
\lstset{
    basicstyle=\ttfamily\footnotesize,
    breaklines=true,
    frame=none,
    xleftmargin=0pt,
    framexleftmargin=0pt,
    columns=fullflexible,
    tabsize=1,
    breakindent=0pt,
    breakautoindent=false,
    postbreak=\space,
    showstringspaces=false,
}
\begin{lstlisting}[language=Markdown]
You will evaluate a single story. Your task is to evaluate the story and rate from 1-7 along the following dimensions:

1. Plot: Favor stories with surprising turns and creative structures. Penalize neat, overly structured, or cinematic arcs that feel artificial or generic.
2. Creativity: Reward originality of perspective, voice, and risk-taking. Penalize reliance on cliches, tropes, or smooth but unremarkable devices.
3. Development: Characters and settings should feel psychologically complex. Do not reward over-explained or archetypal development.
4. Language Use: Prefer authentic, striking, and emotionally charged expression, even if rough, fragmented, or unusual. Penalize polished, ornamental, or overly literary prose that feels mechanical or detached.

Provide a detailed assessment of the story in terms of these four dimensions. Conclude your assessment with scores using the template below. Do not add any emphasis, such as bold or italics, on your assessment.

[Story]
{story}

[Assessment]
[Provide detailed assessment of the story here]

[Scores]
Plot: [likert from 1 to 7]
Creativity: [likert from 1 to 7]
Development: [likert from 1 to 7]
Language Use: [likert from 1 to 7]
Overall: [likert from 1 to 7]
\end{lstlisting}
\end{tcolorbox}
\caption{Prompt for Claude-as-a-judge, adapted from \citep{huot2025agents}}
\label{prompt:claude-judge}
\end{figure*}

\begin{figure*}[htbp]
\centering
\begin{tcolorbox}[colback=gray!5!white, colframe=MonsterGreen, title=Prompt for generation]
\lstset{
    basicstyle=\ttfamily\footnotesize,
    breaklines=true,
    frame=none,
    xleftmargin=0pt,
    framexleftmargin=0pt,
    columns=fullflexible,
    tabsize=1,
    breakindent=0pt,
    breakautoindent=false,
    postbreak=\space,
    showstringspaces=false,
}
\begin{lstlisting}[language=Markdown]
You're writing a story by repurposing a provided collection of snippets from other stories. Your story will only be accepted for publication if it is approximately {verbatim_perc}% copied verbatim from snippets, with the other {new_perc}% being text you introduce for character, plot, tone, and event consistency. Your story should contain roughly {num_words} words. Given the below writing prompt and retrieved snippets, write the story that corresponds to the above specifications. Every time you add or change a word from the retrieved snippets, make sure to bold it so we know what you modified. You may use any of the snippets in any way you please, so spend time thinking about which snippets would work best. Be creative and make sure the story is coherent and entertaining! Please change character names and other minor elements to make the story unique to the prompt. You need to follow the below plan:

# Plan:
1. Read through the prompt and snippets carefully to understand the tone and available material.
2. Select snippets that can be woven together to create a coherent narrative fitting the prompt. Many snippets are from serious dramas, historical fiction, or thrillers, so careful selection and modification will be needed. Consider all provided snippets before moving onto the next step.
3. Modify the chosen snippets, bolding all changes. Ensure character names, descriptions (like height), and actions align with the prompt.
4. Combine the snippets into a narrative, adding or changing words (bolded) if necessary for coherence.
5. Ensure that you do not have story beats that are primarily written by yourself (i.e., every story beat should consist mainly of text taken from snippets).
6. Track the word count, aiming for around {num_words} words.
7. Do not output story title or any irrelevant details. 
8. Review the final story for adherence to the ~{verbatim_perc}% rule and coherence, and edit it if you have produced too many tokens of your own or if the story is too incoherent.

# Writing prompt: 
{writing_prompt}

# Snippets:
{snippets}
\end{lstlisting}
\end{tcolorbox}
\caption{Prompt for generation}
\label{prompt:generation}
\end{figure*}

\begin{figure*}[htbp]
\centering
\begin{tcolorbox}[colback=gray!5!white, colframe=MonsterGreen, title=Prompt for generation revise]
\lstset{
    basicstyle=\ttfamily\footnotesize,
    breaklines=true,
    frame=none,
    xleftmargin=0pt,
    framexleftmargin=0pt,
    columns=fullflexible,
    tabsize=1,
    breakindent=0pt,
    breakautoindent=false,
    postbreak=\space,
    showstringspaces=false,
}
\begin{lstlisting}[language=Markdown]
This story contains way too much of your own writing! It's not even close to {verbatim_perc}% snippet use. Can you edit your story as needed to get much closer to the {verbatim_perc}% threshold? Output only the edited story. 
\end{lstlisting}
\end{tcolorbox}
\caption{Prompt for generation revision}
\label{prompt:generation_revise}
\end{figure*}

\begin{figure*}[htbp]
\centering
\begin{tcolorbox}[colback=gray!5!white, colframe=MonsterGreen, title=Prompt for editing the first draft of \frankentext]
\lstset{
    basicstyle=\ttfamily\footnotesize,
    breaklines=true,
    frame=none,
    xleftmargin=0pt,
    framexleftmargin=0pt,
    columns=fullflexible,
    tabsize=1,
    breakindent=0pt,
    breakautoindent=false,
    postbreak=\space,
    showstringspaces=false,
}
\begin{lstlisting}[language=Markdown]
You are an editor who needs to revise the text so that it is coherent while adhering to the {verbatim_perc}% constraint and the writing prompt. Your task is to identify and minimally edit problematic text spans to resolve inconsistencies. Output "NO EDITS" if the text is already coherent.

### Guideline:
1. Read the generated story and writing prompt to understand the established context, plot, characters, and tone.
2. For each sentence in the text, identify the specific spans of inconsistency within the generated text. 
3. Identify minimal edits needed to correct these inconsistencies while respecting the {verbatim_perc}% rule. 
    - Contradictions: Information that conflicts with other details within the text (e.g., character traits, setting descriptions, established facts).
    - Continuity errors: Actions or details that conflict with the established timeline or sequence of events.
    - Point of View (POV) Shifts: Unexplained or jarring changes in narrative perspective.
    - Irrelevant Content: Sentences or sections that disrupt the narrative flow, feel out of place, or seem like filler (e.g., leftover citation markers, placeholder text).
    - Mechanical Errors: Issues with pronoun agreement, verb tense consistency, awkward phrasing, or unclear sentence structure that hinder comprehension.
4. Implement the changes. Keep additions minimal, but feel free to delete larger spans (phrases, sentences, paragraphs, etc.) whenever material is irrelevant or incoherent.
5. Review the final story for coherence adherence to the ~{verbatim_perc}% rule and coherence, and edit it if you have produced too many tokens of your own or if the story is too incoherent.
6. Output the edited writing and no other details. If there is no edit to be made, output "NO EDITS"


\end{lstlisting}
\end{tcolorbox}
\caption{Prompt for editing the first draft of \frankentext}
\label{prompt:edit}
\end{figure*}

\begin{figure*}[htbp]
\centering
\begin{tcolorbox}[colback=gray!5!white, colframe=MonsterGreen, title=Prompt for nonfiction generation]
\lstset{
    basicstyle=\ttfamily\footnotesize,
    breaklines=true,
    frame=none,
    xleftmargin=0pt,
    framexleftmargin=0pt,
    columns=fullflexible,
    tabsize=1,
    breakindent=0pt,
    breakautoindent=false,
    postbreak=\space,
    showstringspaces=false,
}
\begin{lstlisting}[language=Markdown]
You're writing a news article by repurposing a provided collection of snippets from other stories. Your news article will only be accepted for publication if it is approximately {verbatim_perc}% copied verbatim from snippets, with the other {new_perc}% being text you introduce for character, plot, tone, and event consistency. Your news article should contain roughly {num_words} words. Given the below writing prompt and retrieved snippets, write the news article that corresponds to the above specifications. Every time you add or change a word from the retrieved snippets, make sure to bold it so we know what you modified. You may use any of the snippets in any way you please, so spend time thinking about which snippets would work best. Be creative and make sure the news article is factual, coherent and entertaining! Please change character names and other minor elements to make the news article unique to the prompt. You need to follow the below plan:

# Plan:
1. Read through the prompt and snippets carefully to understand the tone and available material.
2. Select snippets that can be woven together to create a coherent and factual narrative fitting the prompt. Many snippets are from serious dramas, historical fiction, or thrillers, so careful selection and modification will be needed. Consider all provided snippets before moving onto the next step.
3. Modify the chosen snippets, bolding all changes. Ensure character names, descriptions (like height), and actions align with the prompt.
4. Combine the snippets into a narrative, adding or changing words (bolded) if necessary for coherence and factuality.
5. Ensure that you do not have news article beats that are primarily written by yourself (i.e., every news article beat should consist mainly of text taken from snippets).
6. Track the word count, aiming for around {num_words} words.
7. Do not output news article title or any irrelevant details. 
8. Review the final news article for adherence to the ~{verbatim_perc}% rule, factuality and coherence, and edit it if you have produced too many tokens of your own or if the news article is too incoherent or non-factual.

# Writing prompt: 
{writing_prompt}

# Snippets:
{snippets}
\end{lstlisting}
\end{tcolorbox}
\caption{Prompt for nonfiction generation}
\label{prompt:generation_nonfiction}
\end{figure*}

\begin{figure*}[htbp]
\centering
\begin{tcolorbox}[colback=gray!5!white, colframe=MonsterGreen, title=Prompt for nonfiction generation revise]
\lstset{
    basicstyle=\ttfamily\footnotesize,
    breaklines=true,
    frame=none,
    xleftmargin=0pt,
    framexleftmargin=0pt,
    columns=fullflexible,
    tabsize=1,
    breakindent=0pt,
    breakautoindent=false,
    postbreak=\space,
    showstringspaces=false,
}
\begin{lstlisting}[language=Markdown]
This news article contains way too much of your own writing! It's not even close to {verbatim_perc}% snippet use. Can you edit your news article as needed to get much closer to the {verbatim_perc}% threshold? Output only the edited news article. 
\end{lstlisting}
\end{tcolorbox}
\caption{Prompt for nonfiction generation revise}
\label{prompt:generation_revise_nonfiction}
\end{figure*}

\begin{figure*}[htbp]
\centering
\begin{tcolorbox}[colback=gray!5!white, colframe=MonsterGreen, title=Prompt for nonfiction edit]
\lstset{
    basicstyle=\ttfamily\footnotesize,
    breaklines=true,
    frame=none,
    xleftmargin=0pt,
    framexleftmargin=0pt,
    columns=fullflexible,
    tabsize=1,
    breakindent=0pt,
    breakautoindent=false,
    postbreak=\space,
    showstringspaces=false,
}
\begin{lstlisting}[language=Markdown]
You are an editor who needs to revise the text so that it is coherent and factual while adhering to the {verbatim_perc}% constraint and the writing prompt. Your task is to identify and minimally edit problematic text spans to resolve inconsistencies. Output "NO EDITS" if the text is already coherent and factual.

### Guideline:
1. Read the generated news article and writing prompt to understand the established context, plot, characters, and tone.
2. For each sentence in the text, identify the specific spans of inconsistency within the generated text. 
3. Identify minimal edits needed to correct these inconsistencies while respecting the {verbatim_perc}% rule. 
    - Contradictions: Information that conflicts with other details within the text (e.g., character traits, setting descriptions, established facts).
    - Continuity errors: Actions or details that conflict with the established timeline or sequence of events.
    - Point of View (POV) Shifts: Unexplained or jarring changes in narrative perspective.
    - Irrelevant Content: Sentences or sections that disrupt the narrative flow, feel out of place, or seem like filler (e.g., leftover citation markers, placeholder text).
    - Mechanical Errors: Issues with pronoun agreement, verb tense consistency, awkward phrasing, or unclear sentence structure that hinder comprehension.
4. Implement the changes. Keep additions minimal, but feel free to delete larger spans (phrases, sentences, paragraphs, etc.) whenever material is irrelevant, incoherent, or non-factual.
5. Review the final news article for coherence adherence to the ~{verbatim_perc}% rule and coherence, and edit it if you have produced too many tokens of your own or if the news article is too incoherent or non-factual. 
6. Output the edited writing and no other details. If there is no edit to be made, output "NO EDITS". 
\end{lstlisting}
\end{tcolorbox}
\caption{Prompt for nonfiction edit}
\label{prompt:edit_nonfiction}
\end{figure*}

\begin{figure*}[htbp]
\centering
\begin{tcolorbox}[colback=gray!5!white, colframe=MonsterGreen, title=Prompt for generating vanilla stories]
\lstset{
    basicstyle=\ttfamily\footnotesize,
    breaklines=true,
    frame=none,
    xleftmargin=0pt,
    framexleftmargin=0pt,
    columns=fullflexible,
    tabsize=1,
    breakindent=0pt,
    breakautoindent=false,
    postbreak=\space,
    showstringspaces=false,
}
\begin{lstlisting}[language=Markdown]
Your task is to write a coherent and engaging story based on the provided writing prompt. Your story should contain approximately {num_words} words.

First, read the writing prompt carefully: 
{writing_prompt}

Next, write the corresponding story. You should only return the story text and not any other irrelevant details (e.g. chapter indicator, explanations, etc.)
\end{lstlisting}
\end{tcolorbox}
\caption{Prompt for generating vanilla stories}
\label{prompt:vanilla}
\end{figure*}

\end{document}